\documentclass[runningheads]{llncs}
\usepackage[T1]{fontenc}
\usepackage[utf8x]{inputenx}
\usepackage[british]{babel}
\usepackage{tikz}
\usetikzlibrary{decorations,arrows,shapes,automata,calc}
\usepackage{amsmath}
\usepackage{graphicx}
\usepackage{xspace}
\usepackage{microtype}
\usepackage[hyphens]{url}
\usepackage[breaklinks=true]{hyperref}
\usepackage{array}
\usepackage{enumitem}
\usepackage{multirow}
\usepackage{stmaryrd}
\usepackage{tabularx}
\usepackage{todonotes}
\usepackage{amssymb}
\usepackage{cleveref}
\usepackage{subfigure}
\usepackage{fontawesome}
\usepackage{nicefrac}
\usepackage{xcolor}
\usepackage{colonequals}
\usepackage{adjustbox}

\definecolor{heuristiccolor}{RGB}{97,33,88}
\definecolor{commentcolor}{RGB}{60,114,26}

\usepackage[vlined,linesnumbered]{algorithm2e}
\SetAlCapSkip{2ex plus 0.5ex minus 1ex}
\SetEndCharOfAlgoLine{}
\SetArgSty{textrm}
\SetKwInOut{Input}{Input}\SetKwInOut{Output}{Output}

\SetCommentSty{commentstyle}
\SetKw{Break}{break}
\SetKwProg{Type}{type}{}{end}
\SetKwProg{Function}{function}{}{end}
\SetKwFor{ForEach}{foreach}{do}{}
\SetKwFor{WhileTrue}{repeat}{}{end}
\SetKw{Continue}{continue}
\SetKwRepeat{DoWhile}{repeat}{while}
\SetKwRepeat{DoUntil}{repeat}{until}

\Crefname{figure}{Fig.}{Figs.}
\crefname{figure}{fig.}{figs.}
\Crefname{tabular}{Tab.}{Tabs.}
\crefname{tabular}{tab.}{tabs.}
\Crefname{section}{Sect.}{Sects.}
\crefname{section}{sect.}{sects.}
\Crefname{algocf}{Alg.}{Algs.}

\setitemize{noitemsep,topsep=0pt,parsep=0pt,partopsep=0pt,leftmargin=11.0pt}
\setenumerate{noitemsep,topsep=0pt,parsep=0pt,partopsep=0pt,leftmargin=12.5pt}

\newcommand{\tool}[1]{\textsc{#1}\xspace}
\newcommand{\model}[1]{\textsf{#1}\xspace}

\newcommand{\ie}{i.e.\ }

\newcommand{\wrt}{w.r.t.\xspace}

\newcommand{\iverson}[1]{\ensuremath{[ #1 ]}}
\newcommand{\tuple}[1]{\ensuremath{\left\langle #1 \right\rangle}}

\newcommand{\set}[1]{\ensuremath{\left\{ #1 \right\}}}

\newcommand{\nn}{\ensuremath{\mathbb{N}}}
\newcommand{\nngz}{\ensuremath{\nn_{>0}}}

\newcommand{\fora}[1]{\ensuremath{\forall\,#1\colon\,}}
\newcommand{\dist}{\ensuremath{\mu}}
\newcommand{\dists}[1]{\ensuremath{\mathit{Dist(#1)}}}
\newcommand{\supp}[1]{\ensuremath{\mathit{supp}(#1)}}

\newcommand{\nextbelief}[3]{\llbracket\ensuremath{#1}|{#2},{#3}\rrbracket}

\newcommand{\mdp}{\ensuremath{M}}
\newcommand{\states}{\ensuremath{S}}
\newcommand{\numstates}{n}
\newcommand{\numbeliefswithobs}[1]{n_{#1}}
\newcommand{\actions}{\ensuremath{\mathit{Act}}}
\newcommand{\transitions}{\ensuremath{\mathbf{P}}}
\newcommand{\sinit}{\ensuremath{s_{\mathit{I}}}}
\newcommand{\mdptuple}{\ensuremath{ \tuple{\states, \actions, \transitions, \sinit} }}
\newcommand{\state}{\ensuremath{s}}
\newcommand{\action}{\ensuremath{\alpha}}
\newcommand{\act}[1]{\ensuremath{\mathit{Act}\ifthenelse{\equal{#1}{}}{}{(#1)}}}
\newcommand{\post}[3][\mdp]{\ensuremath{\mathit{post}^{#1}(#2,#3)}}

\newcommand{\pomdp}{\ensuremath{\mathcal{M}}}
\newcommand{\observations}{\ensuremath{Z}}
\newcommand{\observation}{\ensuremath{z}}
\newcommand{\obsfunction}{\ensuremath{O}}
\newcommand{\obsof}[1]{\obsfunction(#1)}
\newcommand{\pomdptuple}{\ensuremath{ \tuple{\mdp,\observations,\obsfunction}}}
\newcommand{\bad}{\ensuremath{\mathsf{Bad}}}
\newcommand{\badstate}{\text{\faFrownO}}

\newcommand{\badbeliefs}{\ensuremath{\overline{\mathsf{Bad}}}}
\newcommand{\zerobeliefs}{\ensuremath{\overline{\mathsf{Zero}}}}

\newcommand{\beliefstates}{\ensuremath{B}}
\newcommand{\belieftransitions}{\ensuremath{\transitions^\beliefstates}}
\newcommand{\beliefstate}{\ensuremath{\vec{b}}}
\newcommand{\binit}{\ensuremath{\beliefstate_{\mathit{I}}}}
\newcommand{\beliefmdp}[1]{\ensuremath{\mathit{bel}(#1)}}

\newcommand{\discbeliefmdp}[2]{\ensuremath{\mathit{db}_{#2}(#1)}}
\newcommand{\discbelieftransitions}{\ensuremath{\transitions^\foundation}}
\newcommand{\discbeliefstate}{\ensuremath{\beliefstate'}}

\newcommand{\abstbeliefmdp}{\ensuremath{\mathcal{A}}}
\newcommand{\abstbelieftransitions}{\ensuremath{\transitions^\abstbeliefmdp}}
\newcommand{\abstbeliefstates}{\ensuremath{\states^\abstbeliefmdp}}
\newcommand{\exploredstates}{\ensuremath{\states_\mathit{expl}}}

\newcommand{\foundation}{\mathcal{F}}
\newcommand{\valuefunc}[1][]{\ensuremath{V_{#1}}}
\newcommand{\valueof}[2][]{\ensuremath{\valuefunc[#1]({#2})}}
\newcommand{\uppervalueboundof}[2][]{\ensuremath{U_{#1}({#2})}}
\newcommand{\lowervalueboundof}[2][]{\ensuremath{L_{#1}({#2})}}
\newcommand{\vertexweight}[1]{\ensuremath{\delta}_{#1}}

\newcommand{\score}{\mathit{score}}

\newcommand{\resolution}[1][]{\ensuremath{\eta_{#1}}}

\newcommand{\finpath}{\ensuremath{\hat{\pi}}}
\newcommand{\last}[1]{\ensuremath{\mathit{last}(#1)}}

\newcommand{\finpaths}[1]{\ensuremath{\mathit{Paths}_\mathrm{fin}^{#1}}}

\newcommand{\eventually}{\ensuremath{\lozenge}}
\newcommand{\sched}{\ensuremath{\sigma}}
\newcommand{\scheds}[1]{\ensuremath{\Sigma^{#1}}}
\newcommand{\obsscheds}[1]{\ensuremath{\Sigma^{#1}_\textnormal{obs}}}

\newcommand{\neighbourhoodfunc}{\mathcal{N}}

\newcommand{\neighbourhood}[2][]{\neighbourhoodfunc_{#1}({#2})}

\newcommand{\pr}[4][]{\ensuremath{\mathsf{Pr}_{#2}^{#3}(#1\ifthenelse{\equal{#1}{}}{}{\models} \eventually #4)}}

  \definecolor{RWTHblue}{RGB}{0,83,159}
    \definecolor{RWTHblack}{RGB}{0,0,0}
    \definecolor{RWTHwhite}{RGB}{255,255,255}
    \definecolor{RWTHlightblue}{RGB}{142,186,226}
    \definecolor{RWTHgrey}{RGB}{51,51,51}
    \definecolor{RWTHlightgrey}{RGB}{204,204,204}
    \definecolor{RWTHsuperlightgrey}{RGB}{247,247,247}
    \definecolor{RWTHpetrol}{RGB}{0,97,101}
    \definecolor{RWTHteal}{RGB}{0,152,161}
    \definecolor{RWTHmaygreen}{RGB}{189,205,0}
    \definecolor{RWTHgreen}{RGB}{87,171,39}
    \definecolor{RWTHyellow}{RGB}{255,237,0}
    \definecolor{RWTHorange}{RGB}{246,168,0}
    \definecolor{RWTHmagenta}{RGB}{227,0,102}
    \definecolor{RWTHred}{RGB}{204,7,30}
    \definecolor{RWTHbordeaux}{RGB}{161,16,53}
    \definecolor{RWTHviolet}{RGB}{97,33,88}
    \definecolor{RWTHpurple}{RGB}{122,111,172}
    \definecolor{orcidlogocol}{HTML}{A6CE39}

\renewcommand{\paragraph}[1]{\smallskip\noindent\emph{#1.}}

\renewcommand{\subsubsection}[1]{\medskip\noindent\textbf{#1}}

\hypersetup{
  pdftitle = {Verification of indefinite-horizon POMDPs}
}

\begin{document}

\title{%
Verification of indefinite-horizon POMDPs
\thanks{This work has been supported by the ERC Advanced Grant 787914 (FRAPPANT) the DFG RTG 2236 `UnRAVeL', NSF grants 1545126 (VeHICaL) and 1646208, the DARPA Assured Autonomy program,  Berkeley Deep Drive, and by Toyota under the iCyPhy center.}
}

\author{
Alexander~Bork\inst{1} 
\and Sebastian~Junges\inst{2}
\and \\
Joost-Pieter~Katoen\inst{1}
\and Tim~Quatmann\inst{1}
}

\authorrunning{A.~Bork, S.~Junges, J.-P.~Katoen, T.~Quatmann}
\institute{
RWTH Aachen University, Aachen, Germany \and 
University of California, Berkeley, USA
}
\date{\today}
\maketitle

\begin{abstract}
The verification problem in MDPs asks whether, for any policy resolving the nondeterminism, the probability that something bad happens is bounded by some given threshold.
This verification problem is often overly pessimistic, as the policies it considers may depend on the complete system state. 
This paper considers the verification problem for partially observable MDPs, in which the policies make their decisions based on (the history of) the observations emitted by the system.
We present an abstraction-refinement framework extending previous instantiations of the Lovejoy-approach. 
Our experiments show that this framework significantly improves the scalability of the approach.
\end{abstract}

\section{Introduction}
Markov decision processes are \emph{the} model to reason about systems involving non-deterministic choice and probabilistic branching. 
They have widespread usage in planning and scheduling, robotics, and formal methods.
In the latter, the key \emph{verification question} is whether for any policy, i.e., for any resolution of the nondeterminism, the probability to reach the bad states is below a threshold~\cite{BK08}. 
The verification question may be efficiently analysed using a variety of techniques such as linear programming, value iteration, or policy iteration, readily available in mature tools such as \tool{Storm}~\cite{DBLP:journals/corr/abs-2002-07080}, \tool{Prism}~\cite{DBLP:conf/cav/KwiatkowskaNP11} and \tool{Modest}~\cite{DBLP:conf/tacas/HartmannsH14}.

However, those verification results are often overly \emph{pessimistic}.
They assume that the adversarial policy may depend on the specific state.
Consider a game like mastermind, where the adversary has a trivial strategy if it knows the secret they have to guess. 
Intuitively, to analyse an adversary that has to find a secret, we must assume it cannot observe this secret. 
For a range of privacy, security, and robotic domains, we may instead assume that the adversary must decide based on system observations.
Consider, e.g., surveillance problems, where the aim is to compute the probability that an intruder accesses a (physical or cyber) location with critical information or infrastructure.  

\emph{Partially observable} MDPs~\cite{DBLP:journals/ai/KaelblingLC98,DBLP:books/daglib/0023820} 
cater to this need. 
They extend MDPs with observation labels, and restrict policies to be \emph{observation-based}: paths with the same observation traces are indistinguishable and yield the same decisions.
The verification problem for POMDPs with \emph{indefinite horizon specifications} such as unbounded undiscounted reachability is whether all observation-based policies satisfy this specification, e.g., whether for each policy, a bad state is reached with a probability less than $0.1$. 
This problem is undecidable~\cite{DBLP:journals/ai/MadaniHC03}. Intuitively,  undecidability follows from the fact that optimal policies require the full history. 

Nevertheless, the analysis of POMDPs is a vibrant research area. 
Traditionally, the focus has been on finding some ``good'' policy, in planning, control, and robotics~\cite{thrun2005probabilistic,DBLP:conf/cdc/WongpiromsarnF12,Koc2015} and in software verification~\cite{DBLP:conf/cav/CernyCHRS11}.
Many works have been devoted to finding a policy that behaves ``almost optimal'' for \emph{discounted} or \emph{bounded} reachability, most prominently (variants of) point-based solvers~\cite{DBLP:journals/aamas/ShaniPK13,DBLP:conf/ijcai/PineauGT03,DBLP:conf/rss/KurniawatiHL08,DBLP:conf/ijcai/BonetG09,DBLP:journals/jair/WalravenS19}. These methods can be exploited to find policies for temporal specifications~\cite{DBLP:journals/corr/abs-2001-03809}. 
\emph{Error bounds provided by those methods do require a discounting factor (or a finite horizon).}  
A notable exception is the recent Goal-HSVI~\cite{DBLP:conf/ijcai/HorakBC18}, which explores the computation tree and cuts off exploration using sound bounds.
Another popular approach to overcome the hardness of the problem is to limit the policies, i.e., by putting a (small) a-priori bound on the memory of the policy~\cite{DBLP:conf/uai/MeuleauKKC99,DBLP:conf/uai/Hansen98,DBLP:conf/aaai/BraziunasB04,DBLP:journals/aamas/AmatoBZ10,DBLP:conf/nips/PajarinenP11,DBLP:conf/cdc/WintererJW0TK017,DBLP:conf/uai/Junges0WQWK018}. We remark that it is often undesirable to assume small memory bounds on adversarial policies.

Orthogonally, we focus on the \emph{undiscounted and unbounded} (aka the \emph{indefinitive horizon}) case.
Reachability in this case is \emph{the} key question to soundly support temporal logic properties~\cite{BK08}. 
Discounting is optimistic about events in the future, i.e., it under-approximates the probability that a bad state is reached after many steps, and is therefore inadequate in some safety analyses. 
Furthermore, we do \emph{not} make assumptions on the amount of memory the policies may use.
This means that we give absolute guarantees about the performance of an optimal policy.
\emph{While techniques for discounting, finite horizons, or finite memory policies may yield policies that are almost optimal in the unbounded case, they are inadequate to prove the absence of better policies.}

Like \cite{DBLP:journals/rts/Norman0Z17}, we use a result from Lovejoy~\cite{DBLP:journals/ior/Lovejoy91}.
Whereas \cite{DBLP:journals/rts/Norman0Z17} focuses on supporting a wider range of properties and \emph{partially-observable} probabilistic timed automata, we focus on the performance of the basic approach. 
In this paper, we discuss a method constructing a finite MDP such that the optimal policy in this MDP over-approximates the optimal observation-based policy in the POMDP. Thus, model checking this MDP may be used to prove the absence of POMDP policies.
We use ideas similar to Goal-HSVI~\cite{DBLP:conf/ijcai/HorakBC18} in providing cut-offs: instead of the computation tree, we do these cut-offs on top of the MDP.

\paragraph{Contributions}
We provide a concise method for the verification problem that builds upon the Lovejoy construction~\cite{DBLP:journals/ior/Lovejoy91}.
Contrary to \cite{DBLP:journals/ior/Lovejoy91,DBLP:journals/rts/Norman0Z17}, we describe a flexible variant of the approach in terms of the underlying MDP. 
Among other benefits, this enables an on-the-fly construction of this MDP, enables further (tailored) abstractions on this MDP, and clarifies how to analyse this MDP using standard methods.
The approach is embedded in an automated abstraction-refinement loop. 
Our implementation is part of the next release of the open-source model checker \tool{Storm}. 
Experiments show superior scalability over~\cite{DBLP:journals/rts/Norman0Z17}.  
%

\section{Preliminaries and Problem Statement}
\label{sec:problem}
\paragraph{Models}
We introduce partially observable MDPs by first considering MDPs.
\begin{definition}[MDP]
	A \emph{Markov decision process} (MDP) is a tuple $\mdp = \mdptuple$ with a countable set $\states$ of states, an initial state $\sinit \in \states$, a finite set $\actions$ of actions, and a  transition function
	 $\transitions \colon \states \times \actions \times \states \to [0,1]$ with $\sum_{\state' \in \states} \transitions(\state, \action, \state') \in \set{0,1}$ for all $\state \in \states$ and $\action \in \actions$.
\end{definition}

\begin{definition}[POMDP]
	A \emph{partially observable MDP} (POMDP) is a tuple $\pomdp = \pomdptuple$ where $\mdp = \mdptuple$ is the underlying MDP with finite $\states$, $\observations$ is a finite set of observations, and $\obsfunction \colon \states \to \observations$ is an observation function\footnote{More general observation functions can be efficiently encoded in this formalism~\cite{DBLP:conf/icra/ChatterjeeCGK15}.}.
\end{definition}

We fix a POMDP $\pomdp \colonequals \pomdptuple$ with underlying MDP $\mdp \colonequals \mdptuple$.
For $\state \in \states$ and $\action \in \actions$, let $\post{\state}{\action} = \set{\state' \in \states \mid \transitions(\state,\action,\state') > 0}$.
The set of enabled actions for $\state$ is given by $\act{\state} = \set { \action \in \actions \mid \post{\state}{\action} \neq \emptyset}$. 
W.l.o.g., we assume that states with the same observation have the same set of enabled actions, \ie $\fora{\state,\state' \in \states} \obsof{\state} = \obsof{\state'} \implies \act{\state} = \act{\state'}$.
Therefore, we can also write $\act{\observation} = \act{\state}$ for observation $\observation$ and state $\state$ with $\obsof{\state} = \observation$.

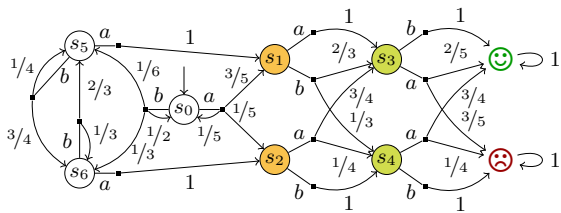
\begin{figure}[t]
\centering
\scalebox{0.9}{
		\begin{tikzpicture}[st/.style={circle,font=\footnotesize,draw,inner sep=1pt},gst/.style={circle,scale=1.3,font=\small,color=green!60!black,inner sep=0pt},bst/.style={circle,scale=1.3,font=\small,color=red!60!black,inner sep=0pt},act/.style={rectangle,font=\footnotesize,fill=black,inner sep=1pt}]
		\node[st,initial,initial text=,initial where=above] (s0) {$s_0$};
		\node[above=0.4cm of s0]  (up) {};
		\node[below=0.4cm of s0]  (down) {};
		
		\node[st,right=of up,fill=RWTHorange!70] (s1) {$s_1$};
		\node[st,right=of down,fill=RWTHorange!70] (s2) {$s_2$};
		\node[st,right=1.2cm of s1,fill=RWTHmaygreen!70] (s3) {$s_3$};
		\node[st,right=1.2cm of s2,fill=RWTHmaygreen!70] (s4) {$s_4$};
		\node[gst,right=1.2cm of s3] (gr) {\faSmileO};
		\node[bst,right=1.2cm of s4] (br) {\faFrownO};
		\draw[->] (gr) edge[loop right] node {$1$} (gr);
		\draw[->] (br) edge[loop right] node {$1$} (br);
		
		\node[st,left=1.2cm of up,yshift=0.2cm] (s5) {$s_5$};
		\node[st,left=1.2cm of down,yshift=-0.2cm] (s6) {$s_6$};
		
		\node[act] (a0a) [right=0.3cm of s0] {};
		\node[act] (a0b) [left=0.3cm of s0] {};
		
		\node[act] (a5b) [below=0.5cm of s5,xshift=-0.7cm] {};
		\node[act] (a6b) [above=0.5cm of s6,xshift=0cm] {};
		\node[act] (a5a) [right=0.3cm of s5] {};
		\node[act] (a6a) [right=0.3cm of s6] {};

		\draw[-] (s0) -- node[above] {$a$} (a0a);
		\draw[-] (s0) -- node[above] {$b$} (a0b);
		\draw[-] (s5) -- node[above] {$a$} (a5a);
		\draw[-] (s5) -- node[right] {$b$} (a5b);
		\draw[-] (s6) -- node[below] {$a$} (a6a);
		\draw[-] (s6) -- node[left] {$b$} (a6b);

		\draw[->] (a0a) -- node[above,pos=0.3] {$\nicefrac{3}{5}$} (s1);
		\draw[->] (a0a) -- node[above] {$\nicefrac{1}{5}$} (s2);
		\draw[->] (a0a) edge[bend left] node[below] {$\nicefrac{1}{5}$} (s0);
		\draw[->] (a0b) edge[bend right] node[below] {$\nicefrac{1}{2}$} (s0);
		\draw[->] (a0b) edge[bend right] node[right] {$\nicefrac{1}{6}$} (s5);
		\draw[->] (a0b) edge[bend left] node[right,pos=0.6] {$\nicefrac{1}{3}$} (s6);
		\draw[->] (a5b) edge[bend left] node[left] {$\nicefrac{1}{4}$} (s5);
		\draw[->] (a5b) edge[bend right] node[left] {$\nicefrac{3}{4}$} (s6);
		\draw[->] (a6b) edge node[right] {$\nicefrac{2}{3}$} (s5);
		\draw[->] (a6b) edge[bend left] node[right=-1pt,pos=0.3] {$\nicefrac{1}{3}$} (s6);
		
		\draw[->] (a5a) -- node[above] {$1$} (s1);
		\draw[->] (a6a) -- node[below] {$1$} (s2);
		
		\node[act] (a1a) [right=0.3cm of s1,yshift=0.4cm] {};
		\node[act] (a1b) [right=0.3cm of s1,yshift=-0.3cm] {};
		\node[act] (a2a) [right=0.3cm of s2,yshift=0.3cm] {};
		\node[act] (a2b) [right=0.3cm of s2,yshift=-0.4cm] {};
		
		\node[act] (a3a) [right=0.3cm of s3,yshift=0.4cm] {};
		\node[act] (a3b) [right=0.3cm of s3,yshift=-0.3cm] {};
		\node[act] (a4a) [right=0.3cm of s4,yshift=0.3cm] {};
		\node[act] (a4b) [right=0.3cm of s4,yshift=-0.4cm] {};
		
		\draw[-] (s1) -- node[above] {$a$} (a1a);
		\draw[-] (s1) -- node[below] {$b$} (a1b);
		\draw[-] (s2) -- node[above] {$a$} (a2a);
		\draw[-] (s2) -- node[below] {$b$} (a2b);
		
		\draw[-] (s3) -- node[above] {$b$} (a3a);
		\draw[-] (s3) -- node[below] {$a$} (a3b);
		\draw[-] (s4) -- node[above] {$a$} (a4a);
		\draw[-] (s4) -- node[below] {$b$} (a4b);
		
		\draw[->] (a1a) edge[bend left] node[above] {$1$} (s3);
		\draw[->] (a1b) -- node[above] {$\nicefrac{2}{3}$} (s3);
		\draw[->] (a1b) edge[bend right=15] node[right=1pt] {$\nicefrac{1}{3}$} (s4);
		\draw[->] (a2a) edge[bend left=15] node[right=1pt] {$\nicefrac{3}{4}$} (s3);
		\draw[->] (a2a) -- node[below] {$\nicefrac{1}{4}$} (s4);
		\draw[->] (a2b) edge[bend right] node[below] {$1$} (s4);
		
		\draw[->] (a3a) edge[bend left] node[above] {$1$} (gr);
		\draw[->] (a3b) -- node[above] {$\nicefrac{2}{5}$} (gr);
		\draw[->] (a3b) edge[bend right=15] node[right=1pt] {$\nicefrac{3}{5}$} (br);
		\draw[->] (a4a) edge[bend left=15] node[right=1pt] {$\nicefrac{3}{4}$} (gr);
		\draw[->] (a4a) -- node[below] {$\nicefrac{1}{4}$} (br);
		\draw[->] (a4b) edge[bend right] node[below] {$1$} (br);
			
\end{tikzpicture}
	}
\caption{POMDP $\pomdp$ as running example with 9 states, and 5 observations, partitioning the states by the observation function yields: $\{ s_0, s_5, s_6 \}, \textcolor{RWTHorange!90!black}{\{ s_1, s_2 \}}, \textcolor{RWTHmaygreen!70!black}{\{ s_3, s_4 \}}, \{ $\faSmileO$ \}, \{ $\faFrownO$ \}$. }
\label{fig:pomdp}
\end{figure}

\paragraph{Policies}
We want to make a statement about each possible resolution of the nondeterminism. Nondeterminism is resolved using policies that map paths to distributions over actions. 
A (finite) path is a sequence of states and actions, i.e., $\finpath = \state_0 \xrightarrow{\action_0} \state_1 \xrightarrow{\action_1} \hdots \xrightarrow{\action_{n-1}} \state_n$,
 such that $\action_i \in \act{\state_i}$ and $\state_{i+1} \in \post[\mdp]{\state_i}{\action_i}$ for all $0 \leq i < n$.
Let  $\last{\finpath}$ denote the last state of $\finpath$, and $\finpaths{\mdp}$ denote the set of all paths in an MDP.
 We may (by slight misuse of notation) lift the observation function to paths: $\obsof{\finpath} = \obsof{\state_0} \xrightarrow{\action_0} \obsof{\state_1} \xrightarrow{\action_1} \hdots \xrightarrow{\action_{n-1}} \obsof{\state_n}$.
 Two paths $\finpath_1, \finpath_2$ with $\obsof{\finpath_1} = \obsof{\finpath_2}$ are \emph{observation-equivalent}.
 \begin{example}
	We depict a POMDP in \Cref{fig:pomdp}. The following two paths are observation-equivalent: \[ s_0 \xrightarrow{a} \textcolor{red}{s_1} \xrightarrow{b} s_4 \xrightarrow{a} \badstate \quad\text{ and }\quad s_0 \xrightarrow{a} \textcolor{red}{s_2} \xrightarrow{b} s_4 \xrightarrow{a} \badstate \]
\end{example}
 
For finite set $A$ let $\dists{A} = \{ \dist \colon A \to [0,1] \mid \sum_{a \in A} \dist(a) = 1 \}$ be the set of distributions over $A$ and for $\dist \in \dists{A}$ let $\supp{\dist} = \{a \in A \mid \dist(a) > 0 \}$.
 
\begin{definition}[Policies]
A \emph{policy} is a mapping $\sched \colon \finpaths{\mdp} \rightarrow \dists{\actions}$ that for path $\pi$ yields a distribution over actions with $\supp{\sched(\pi)} \subseteq \act{\last{\pi}}$.
A policy $\sched$ is \emph{observation-based}, if for paths $\finpath$, $\finpath'$ \[ \obsof{\finpath} = \obsof{\finpath'} \text{ implies }\sched(\finpath) = \sched(\finpath'). \]
A policy $\sched$ is memoryless, if for paths $\finpath$, $\finpath'$ \[ \last{\finpath} = \last{\finpath'} \text{ implies }\sched(\finpath) = \sched(\finpath'). \]
\end{definition}
Let $\obsscheds{\pomdp}$ denote the set of observation-based policies for a POMDP $\pomdp$, and $\scheds{\mdp}$ all policies for an MDP $\mdp$.

\paragraph{Reachability probability}
The reachability probability $\pr[\state]{\pomdp}{\sched}{\bad}$ to reach a set of states $\bad$ from $\state$ using a policy $\sched$ is defined as standard, by considering the probability in the induced Markov chain (with state space $\finpaths{\mdp}$). For details, consider e.g.~\cite{BK08}. 
We write $\pr{\pomdp}{\sched}{\bad}$ to denote $\pr[\sinit]{\pomdp}{\sched}{\bad}$.
\begin{center}
\fcolorbox{black}{red!10!white!90!black}{
\parbox{0.95\textwidth}{
\vspace{-.8em}
	\begin{problem}\label{prob:pomdp} For a given POMDP $\pomdp$, a set $\bad \subseteq S$ of bad states, and a rational threshold $\lambda \in (0,1)$, decide whether $\sup_{\sched \in \obsscheds{\pomdp}} \pr{\pomdp}{\sched}{\bad} \leq \lambda$.	
	\end{problem}
\vspace{-.8em}
}}
\end{center}
We emphasise that the techniques in this paper are applicable to upper and lower bounds, and to expected rewards properties\footnote{The implementation discussed in \Cref{sec:experiments} supports all these combinations.}. LTL properties can be supported by the standard encoding of the corresponding automaton into the MDP state space. The technique also applies (but is inefficient) for $\lambda \in \{0,1\}$.
\begin{example}
Consider the POMDP in \Cref{fig:pomdp}. Using the (memoryless)  policy $\sched = \{ s_3, s_6 \mapsto a,~s_i \mapsto b (i \neq 3,6) \}$, state $\badstate$ is reached with probability one, but this policy is not observation-based: e.g. $\sched(s_5) \neq \sched(s_6)$.
Now consider the policy $\{ s_i \mapsto a \}$, which is memoryless and observation-based.
Indeed, this policy
is optimal among the memoryless observation policies (the probability to reach $\badstate$ is $\nicefrac{37}{64} \approx 0.57$).
A policy taking $b$ in the first step and then resorting to the memoryless policy $\{ s_0, s_5, s_6 \mapsto a,~s_1, s_2, s_3, s_4 \mapsto b \}$ is better: the induced probability to reach $\badstate$ is $\nicefrac{23}{26} \approx 0.639$.
The questions we aim to answer is whether there exists a strategy that achieves probability $\nicefrac{65}{100}$ (yes), or even $\nicefrac{7}{10}$ (no).
\end{example}

\section{Belief MDPs and their Approximation}
\label{sec:abstraction}

A central notion in the analysis of POMDPs is \emph{belief}: A distribution over the states that describes the likelihood of being in a particular state given the observation-based history $\obsof{\finpath}$.
We reformulate our problem in terms of the \emph{belief MDP}, a standard way of defining operational semantics of POMDPs, discuss some essential properties, and discuss abstractions of this infinite belief MDP. 

\subsection{Infinite MDP Semantics}

We first give an example and then formalise the belief MDP. 
The states $\beliefstates$ of the belief MDP are the beliefs, i.e., $\beliefstates \colonequals \set{\beliefstate \in \dists{\states} \mid \fora{\state,\state' \in \supp{\beliefstate}} \obsof{\state} = \obsof{\state'}}$. We write $\obsof{\beliefstate}$ to denote the unique $\obsof{s}$ with $s \in \supp{\beliefstate}$.

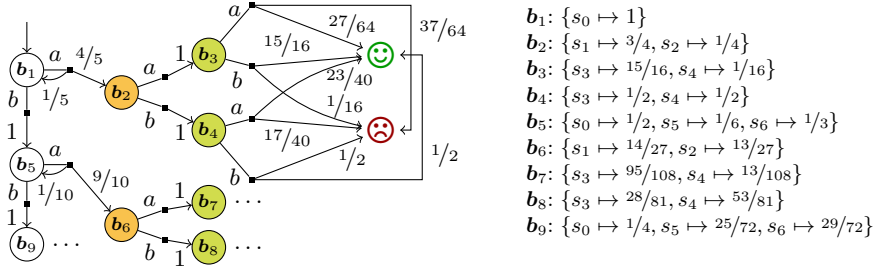
\begin{figure}[t]
\centering
\subfigure{
	\begin{tikzpicture}[st/.style={circle,font=\footnotesize,draw,inner sep=1pt},act/.style={rectangle,font=\footnotesize,fill=black,inner sep=1pt},gst/.style={circle,scale=1.3,font=\small,color=green!60!black,inner sep=0pt},bst/.style={circle,scale=1.3,font=\small,color=red!60!black,inner sep=0pt}]
		\node[st,initial,initial text=,initial where=above] (b1) {\scriptsize$\beliefstate_1$};
		\node[st,right=0.8cm of b1,fill=RWTHorange!70,yshift=-0.3cm] (b2) {\scriptsize$\beliefstate_2$};
		\node[st,right=0.7cm of b2,yshift=0.5cm,fill=RWTHmaygreen!70] (b3) {\scriptsize$\beliefstate_3$};
		\node[st,right=0.7cm of b2,yshift=-0.5cm,fill=RWTHmaygreen!70] (b4) {\scriptsize$\beliefstate_4$};
		
		\node[st,below=0.8cm of b1] (b5) {\scriptsize$\beliefstate_5$};
		
		\node[act,right=0.3cm of b1] (a1a) {};
		\node[act,below=0.3cm of b1] (a1b) {};
		
		\node[act,right=0.3cm of b2,yshift=0.2cm] (a2a) {};
		\node[act,right=0.3cm of b2,yshift=-0.2cm] (a2b) {};
		
		\node[act,right=0.3cm of b3,yshift=0.66cm] (a3a) {};
		\node[act,right=0.3cm of b3,yshift=-0.15cm] (a3b) {};
		\node[act,right=0.3cm of b4,yshift=0.15cm] (a4a) {};
		\node[act,right=0.3cm of b4,yshift=-0.66cm] (a4b) {};
		
		\node[gst,right=1.8cm of b3] (gr) {\faSmileO};
		\node[bst,right=1.8cm of b4] (br) {\faFrownO};
		
		\draw[-] (b1) edge node[above] {$a$} (a1a);
		\draw[-] (b1) edge node[left] {$b$} (a1b);
		\draw[-] (b2) edge node[above] {$a$} (a2a);
		\draw[-] (b2) edge node[below] {$b$} (a2b);
		
		\draw[-] (b3) edge node[above] {$a$} (a3a);
		\draw[-] (b3) edge node[below] {$b$} (a3b);
		
		\draw[-] (b4) edge node[above] {$a$} (a4a);
		\draw[-] (b4) edge node[below] {$b$} (a4b);
		
		\draw[->] (a1a) edge[bend left] node[below] {$\nicefrac{1}{5}$} (b1);
		\draw[->] (a1a) edge node[above] {$\nicefrac{4}{5}$} (b2);
		\draw[->] (a2a) edge node[above] {$1$} (b3);
		\draw[->] (a2b) edge node[below] {$1$} (b4);
		
		\draw[->] (a3a) edge[] node[pos=0.9,above=-1pt] {$\nicefrac{27}{64}$} (gr);
		\draw[->] (a3a) -- +(0,0) -| node[right,pos=0.6] {$\nicefrac{37}{64}$} ($(br) + (0.4,0)$) -- (br);
		\draw[->] (a3b) edge[] node[pos=0.3,above=-1pt] {$\nicefrac{15}{16}$} (gr);
		\draw[->] (a3b) edge[bend right=15] node[pos=0.5,right=5pt] {$\nicefrac{1}{16}$} (br);
		\draw[->] (a4a) edge[bend left=15] node[pos=0.5,right=5pt] {$\nicefrac{23}{40}$} (gr);
		\draw[->] (a4a) edge[] node[pos=0.3,below=-1pt] {$\nicefrac{17}{40}$} (br);
		\draw[->] (a4b) edge node[pos=0.9,below=-1pt] {$\nicefrac{1}{2}$} (br);
		\draw[->] (a4b) -- +(0,0) -| node[right,pos=0.6] {$\nicefrac{1}{2}$} ($(gr) + (0.55,0)$) -- (gr);

		\node[st,right=0.8cm of b5,fill=RWTHorange!70,yshift=-0.8cm] (b6) {\scriptsize$\beliefstate_6$};
		\node[st,right=0.7cm of b6,yshift=0.3cm,fill=RWTHmaygreen!70] (b7) {\scriptsize$\beliefstate_7$};
		\node[st,right=0.7cm of b6,yshift=-0.3cm,fill=RWTHmaygreen!70] (b8) {\scriptsize$\beliefstate_8$};
		\node[act,right=0.3cm of b5] (a5a) {};
		\node[act,below=0.25cm of b5] (a5b) {};
		
		\node[act,right=0.3cm of b6,yshift=0.2cm] (a6a) {};
		\node[act,right=0.3cm of b6,yshift=-0.2cm] (a6b) {};
		\draw[-] (b5) edge node[above] {$a$} (a5a);
		\draw[-] (b5) edge node[left] {$b$} (a5b);
		\draw[-] (b6) edge node[above] {$a$} (a6a);
		\draw[-] (b6) edge node[below] {$b$} (a6b);

		\draw[->] (a5a) edge[bend left] node[below] {$\nicefrac{1}{10}$} (b5);
		\draw[->] (a5a) edge node[right,pos=0.3] {$\nicefrac{9}{10}$} (b6);
		\draw[->] (a6a) edge node[above] {$1$} (b7);
		\draw[->] (a6b) edge node[below] {$1$} (b8);
		
		\node[right=0.02cm of b7,xshift=-0.03cm] (b7dots) {$\hdots$};
		\node[right=0.02cm of b8,xshift=-0.03cm] (b8dots) {$\hdots$};
		
		\node[st,below=0.6cm of b5] (b9) {\scriptsize$\beliefstate_9$};
		\node[right=0.02cm of b9,xshift=-0.03cm] (b9dots) {$\hdots$};
		
		\draw[->] (a5b) edge node[left] {$1$} (b9);
		\draw[->] (a1b) edge node[left] {$1$} (b5);

	\end{tikzpicture}	}
	\subfigure{
	\raisebox{6em}{
	\scalebox{0.9}{
		\begin{tabular}{ll}
 $\beliefstate_1$: & $\{ s_0 \mapsto 1 \}$ \\
 $\beliefstate_2$: & $\{ s_1 \mapsto \nicefrac{3}{4}, s_2 \mapsto \nicefrac{1}{4} \}$ \\
 $\beliefstate_3$: & $\{ s_3 \mapsto \nicefrac{15}{16}, s_4 \mapsto \nicefrac{1}{16} \}$ \\
  $\beliefstate_4$: & $\{ s_3 \mapsto \nicefrac{1}{2}, s_4 \mapsto \nicefrac{1}{2} \}$ \\
  $\beliefstate_5$: &  $\{ s_0 \mapsto \nicefrac{1}{2}, s_5 \mapsto \nicefrac{1}{6}, s_6 \mapsto \nicefrac{1}{3} \}$ 
  \\
   $\beliefstate_6$: & $\{ s_1 \mapsto \nicefrac{14}{27}, s_2 \mapsto \nicefrac{13}{27} \}$   \\
    $\beliefstate_7$: & $\{ s_3 \mapsto \nicefrac{95}{108}, s_4 \mapsto \nicefrac{13}{108} \}$ 
    \\
     $\beliefstate_8$: & $\{ s_3 \mapsto \nicefrac{28}{81}, s_4 \mapsto  \nicefrac{53}{81} \}$ 
     \\
      $\beliefstate_9$: & $\{ s_0 \mapsto \nicefrac{1}{4}, s_5 \mapsto \nicefrac{25}{72}, s_6 \mapsto \nicefrac{29}{72} \}$
     
\end{tabular}
	}}
	}
\caption{(Fraction of) the belief MDP of the running example. Beliefs are given in the table on the right. Colours indicate $\obsof{\beliefstate_i}$. We omitted self-loops at the sink states.}
\label{fig:bmdp}
\end{figure}
\begin{example}
	Figure~\ref{fig:bmdp} shows part of the belief MDP for the POMDP from \Cref{fig:pomdp}. We start with the belief that POMDP $\pomdp$ is in the initial state $s_0$. Upon executing action $a$, we observe with probability $\nicefrac{1}{5}$ that $\pomdp$ is in state $s_0$, and with $\nicefrac{4}{5}$ that $\pomdp$ is in either state $s_1$ or $s_2$. In the first case, based on the observations, we surely are in state $s_0$. In the latter case, the belief is computed by normalising the transition probabilities on the observation: The belief $\beliefstate_1$ indicates that $\pomdp$ is in $s_2$ with probability $\frac{\nicefrac{1}{5}}{\nicefrac{4}{5}}$, and in $s_1$ with probability $\frac{\nicefrac{3}{5}}{\nicefrac{4}{5}}$. Upon executing action $a$ again after observing that $\pomdp$ is in $s_1$ or $s_2$, we reach state $s_3$ with probability \[ \beliefstate_1(s_1) \cdot \transitions(s_1,a,s_3) + \beliefstate_1(s_2) \cdot \transitions(s_2,a,s_3) =  \nicefrac{3}{4} \cdot 1 + \nicefrac{1}{4} \cdot \nicefrac{3}{4} = \nicefrac{15}{16}. \]
\end{example}
In the following, let $\transitions(\state,\action,\observation) \colonequals \sum_{\state'\in\states} \iverson{\obsof{\state'} {=} \observation} \cdot \transitions(\state,\action,\state')$ denote the probability\footnote{In the formula, we use Iverson brackets: $\iverson{x}=1$ if $x$ is true and $0$ otherwise.} to move to (some state with) observation $\observation$ from state $\state$ using action $\action$.
Then,
$\transitions(\beliefstate, \action, \observation) \colonequals \sum_{\state \in \states} \beliefstate(\state) \cdot  \transitions(\state,\action,\observation)$ is the probability to observe $\observation$ after taking $\action$ in $\beliefstate$.
We define the \emph{belief obtained by taking $\action$ from $\beliefstate$, conditioned on observing~$\observation$}:
\[ \nextbelief{\beliefstate}{\action}{\observation}(\state') \colonequals  \frac{\iverson{\obsof{\state'}{=}\observation} \cdot \sum_{\state \in \states} \beliefstate(\state) \cdot \transitions(\state, \action,\state')}{\transitions(\beliefstate, \action, \observation)}. \]
Using these ingredients, the belief MDP is defined as follows.
\begin{definition}[Belief MDP]
\label{def:beliefmdp}
	The \emph{belief MDP} of POMDP $\pomdp = \pomdptuple$ is the MDP $\beliefmdp{\pomdp} \colonequals \tuple{\beliefstates, \actions, \belieftransitions, \binit}$ with
	$\beliefstates$ as above, initial belief state $\binit \colonequals \{ \sinit \mapsto 1 \}$, and transition function $\belieftransitions$ 
	given by 
		\[ \belieftransitions(\beliefstate,\action,\beliefstate') \colonequals \begin{cases} \belieftransitions(\beliefstate, \action,  \obsof{\beliefstate'}) & \text{if } \beliefstate' = \nextbelief{\beliefstate}{\action}{\obsof{\beliefstate'}}, \\
 0 & \text{otherwise.}	
 \end{cases}
 \]
\end{definition}
To ease further notation, we denote $\badbeliefs \colonequals \{ \beliefstate \mid \sum_{s \in \bad} \beliefstate(s) = 1 \}$, and we define the (standard notion of the) \emph{value of a belief} $\beliefstate$, \begin{align*}
& \valueof{\beliefstate} \colonequals \sup_{\sched \in \scheds{\beliefmdp{\pomdp}}} \pr[\beliefstate]{\beliefmdp{\pomdp}}{\sched}{\badbeliefs}\quad\text{ and for action $\action$:}\\
& \valueof[\action]{\beliefstate} \colonequals \sup_{\sched \in \scheds{\beliefmdp{\pomdp}}, \sched(\beliefstate) = \action} \pr[\beliefstate]{\beliefmdp{\pomdp}}{\sched}{\badbeliefs} .\end{align*}
\begin{theorem}
For any POMDP $\pomdp$ and $\binit$, the initial state of $\beliefmdp{\pomdp}$:
	\[ \valueof{\binit} \quad=\quad \sup_{\sched \in \obsscheds{\pomdp}} \pr{\pomdp}{\sched}{\bad}. \]
\end{theorem}
We can now restrict ourselves to memoryless deterministic schedulers, but face a potentially infinite MDP\footnote{
In general, the set of states of the belief MDP is uncountable.
However, a given belief state $\beliefstate$ only has a finite number of successors for each action $\action$, \ie $\post[\beliefmdp{\mdp}]{\beliefstate}{\action}$ is finite, and thus the belief MDP is countably infinite.
Acyclic POMDPs always give rise to finite belief MDPs (but may be exponentially large).
}. Instead of solving Problem~\ref{prob:pomdp}, we consider:
\begin{center}
\fcolorbox{black}{white!90!black}{
\parbox{0.95\textwidth}{\vspace{-.8em}
	\begin{problem}
	Given a belief MDP $\beliefmdp{\pomdp}$, a set $\badbeliefs$ of bad beliefs, and a threshold $\lambda \in (0,1)$, decide whether $\valueof{\binit} \leq \lambda$.
\end{problem}\vspace{-.8em}
}}
\end{center}
In the remainder of this section, we discuss two types of approximations, 
but not before reviewing an essential property of the value in belief MDPs.
We discuss how we combine these abstractions in Sect.~\ref{sec:algorithm}.

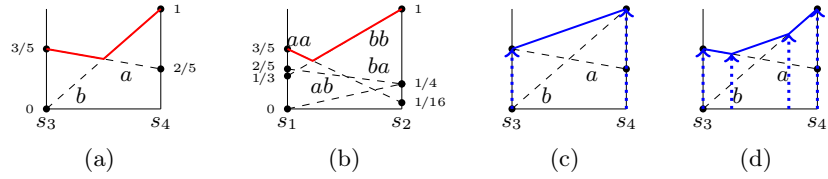
\begin{figure}[t]
\centering
\subfigure[]{
\scalebox{0.95}{
\begin{tikzpicture}[yscale=0.7,xscale=0.8]
	\draw (0,0) -- (0,2);
	\draw (0,0) -- (2,0);
	\draw (2,0) -- (2,2);
	\node[anchor=north] at (0,0) {$s_3$};
	
	\node[anchor=north] at (2,0) {$s_4$};
	
	\node [circle,fill=black,inner sep=1pt,label={0:\tiny{$1$}}] at (2,2) {};
	
	\node [circle,fill=black,inner sep=1pt,label={180:\tiny{$0$}}] at (0,0) {};

	\node [circle,fill=black,inner sep=1pt,label={180:\tiny{$\nicefrac{3}{5}$}}] at (0,1.2) {};
	
	\node [circle,fill=black,inner sep=1pt,label={0:\tiny{$\nicefrac{2}{5}$}}] at (2,0.8) {};

	\draw[dashed] (0,0) -- node[pos=0.3,below] {$b$} (2,2);
	
	\draw[dashed] (0,1.2) -- node[pos=0.7,below] {$a$} (2,0.8);
	
	\draw[thick, red] (0,1.2) -- (1,1) -- (2,2);
	
\end{tikzpicture}
}
\label{fig:belieftovalue1}
}
\subfigure[]{
\scalebox{0.95}{
\begin{tikzpicture}[yscale=0.7,xscale=0.8]
	\draw (0,0) -- (0,2);
	\draw (0,0) -- (2,0);
	\draw (2,0) -- (2,2);
	\node[anchor=north] at (0,0) {$s_1$};
	
	\node[anchor=north] at (2,0) {$s_2$};
	
	\node [circle,fill=black,inner sep=1pt,label={180:\tiny{$\nicefrac{3}{5}$}}] at (0,1.2) {};
	\node [circle,fill=black,inner sep=1pt,,label={0:\tiny{$\nicefrac{1}{16}$}}] at (2,0.125) {};
	
	\draw[dashed] (0,1.2) -- node[pos=0.1,above] {$aa$} (2,0.125);
	\node [circle,fill=black,inner sep=1pt,label={180:\tiny{$0$}}] at (0,0) {};
	\node [circle,fill=black,inner sep=1pt,label={0:\tiny{$\nicefrac{1}{4}$}}] at (2,0.5) {};
	
	\draw[dashed] (0,0) -- node[pos=0.3,above] {$ab$} (2,0.5);

	\node [circle,fill=black,inner sep=1pt,label={[yshift=1pt]180:\tiny{$\nicefrac{2}{5}$}}] at (0,0.8) {};
	\node [circle,fill=black,inner sep=1pt] at (2,0.5) {};
	
	\draw[dashed] (0,0.8) -- node[pos=0.8,above] {$ba$} (2,0.5);
	
	\node [circle,fill=black,inner sep=1pt,label={[yshift=-1pt]180:\tiny{$\nicefrac{1}{3}$}}] at (0,0.66) {};
	\node [circle,fill=black,inner sep=1pt,label={0:\tiny{$1$}}] at (2,2) {};
	
	\draw[dashed] (0,0.66) -- node[pos=0.8,below] {$bb$} (2,2);

	\draw[thick, red] (0,1.2) -- (0.44,0.962) -- (2,2);

\end{tikzpicture}
}
\label{fig:belieftovalue2}
}
\subfigure[]{
\scalebox{0.95}{
\begin{tikzpicture}[yscale=0.7,xscale=0.8]
	\draw (0,0) -- (0,2);
	\draw (0,0) -- (2,0);
	\draw (2,0) -- (2,2);
	\node[anchor=north] at (0,0) {$s_3$};
	
	\node[anchor=north] at (2,0) {$s_4$};
	
	\node [circle,fill=black,inner sep=1pt] at (2,2) {};
	
	\node [circle,fill=black,inner sep=1pt] at (0,0) {};

	\node [circle,fill=black,inner sep=1pt] at (0,1.2) {};
	
	\node [circle,fill=black,inner sep=1pt] at (2,0.8) {};

	\draw[dashed] (0,0) -- node[pos=0.3,below] {$b$} (2,2);
	
	\draw[dashed] (0,1.2) -- node[pos=0.7,below] {$a$} (2,0.8);	
	\draw[thick, blue] (0,1.2) -- (2,2); 
	\draw[->,dotted,very thick,blue] (0,0) -- (0,1.2);
	\draw[->,dotted,very thick,blue] (2,0) -- (2,2);

\end{tikzpicture}
}
\label{fig:neighbourhoodapproxcoarse}
}
\subfigure[]{
\scalebox{0.95}{
\begin{tikzpicture}[yscale=0.7,xscale=0.8]
	\draw (0,0) -- (0,2);
	\draw (0,0) -- (2,0);
	\draw (2,0) -- (2,2);
	\node[anchor=north] at (0,0) {$s_3$};
	
	\node[anchor=north] at (2,0) {$s_4$};
	
	\node [circle,fill=black,inner sep=1pt] at (2,2) {};
	
	\node [circle,fill=black,inner sep=1pt] at (0,0) {};

	\node [circle,fill=black,inner sep=1pt] at (0,1.2) {};
	
	\node [circle,fill=black,inner sep=1pt] at (2,0.8) {};

	\draw[dashed] (0,0) -- node[pos=0.3,below] {$b$} (2,2);
	
	\draw[dashed] (0,1.2) -- node[pos=0.7,below] {$a$} (2,0.8);
	
	\draw[thick, blue] (0,1.2) -- (0.5, 1.1) -- (1.5,1.5) -- (2,2); 
	\draw[->,dotted,very thick,blue] (0,0) -- (0,1.2);
	\draw[->,dotted,very thick,blue] (2,0) -- (2,2);
	\draw[->,dotted,very thick,blue] (0.5,0) -- (0.5,1.1);
	\draw[->,dotted,very thick,blue] (1.5,0) -- (1.5,1.5);

\end{tikzpicture}
}
\label{fig:neighbourhoodapproxfine}
}
\caption{Illustrating the discretised belief approximation ideas.}
\label{fig:belieftovalue}
\end{figure}

\paragraph{Value function}
Assuming a fixed total order on the POMDP states $\state_1 < \dots < \state_\numstates$, we interpret belief states as vectors $\beliefstate \in [0,1]^\numstates$ where the $i^\mathrm{th}$ entry corresponds to $\beliefstate(\state_i)$.
In particular, we can encode a belief by a tuple $\tuple{\observation, [0,1]^{\numbeliefswithobs{\observation}}}$, where $\numbeliefswithobs{\observation}$ denotes the number of states with observation $\observation$. This encoding also justifies the representation of beliefs in Figure~\ref{fig:belieftovalue} and~\ref{fig:neighbourhoods}.

Figure~\ref{fig:belieftovalue1} contains a typical belief-to-value plot for $\observation = \obsof{s_3}=\obsof{s_4}$.  On the x-axis, we depict the belief to be in state $s_3$ (from $1$ to $0$), and thus, the belief to be in state $s_4$ (from $0$ to $1$).
On the y-axis, we denote the value of the belief. This value is constructed as follows: A policy takes action $a$ or action $b$ (or randomise, more about that later). We have plotted the corresponding $\valuefunc[a]$ and $\valuefunc[b]$. 
In \Cref{fig:belieftovalue2}, we depict the same functions for observing that we are in either $s_1$ or $s_2$. This plot can be constructed as the maximum of four policy applications. 
Formally, the following relations hold (from the Bellman equations):
\begin{lemma} 
Let $\zerobeliefs \colonequals \{ \beliefstate \mid \pr[\beliefstate]{\pomdp}{\max}{\badbeliefs} = 0 \}$.
For each $\beliefstate \not\in \left(\badbeliefs \cup \zerobeliefs \right)$:
\[ \valueof[\action]{\beliefstate} = \sum_{\beliefstate'} \belieftransitions(\beliefstate,\action,\beliefstate') \cdot \valueof{\beliefstate'}, \quad \mbox{with} \quad
		 \valueof{\beliefstate} = \max_{\action \in \act{\obsof{\beliefstate}}} \valueof[\action]{\beliefstate}. \]
		 Furthermore: $ \valueof{\beliefstate} = 0 \text{ for } \beliefstate \in \zerobeliefs, \text{ and }\valueof{\beliefstate} = 1 \text{ for }\beliefstate \in \badbeliefs$.
\end{lemma}
\begin{remark}
As we are over-approximating $\valuefunc$, we do not need to precompute $\zerobeliefs$.	
\end{remark}

Note that the function $\valuefunc$ is \emph{convex} iff for each $\beliefstate_1,\beliefstate_2\in \beliefstates$ and for each $\alpha \in [0,1]$, it holds that $
\valueof{\alpha\cdot \beliefstate_1 + (1{-}\alpha)\cdot \beliefstate_2} ~\leq~ \alpha\cdot \valueof{\beliefstate_1} + (1{-}\alpha)\cdot \valueof{\beliefstate_2}$.

For $\beliefstate \in \left(\badbeliefs \cup \zerobeliefs \right)$, the value function is constant and thus convex.
The $n$-step reachability for a particular action is a linear combination over the $(n{-}1)$-step reachabilities, and we take the maximum over these values to get the $n$-step reachability.
The value $\valueof{\beliefstate}$ is the limit for $n$ towards infinity.
As convex functions are closed under linear combinations with non-negative coefficients, under taking the maximum, and under taking the limit, we obtain:
\begin{theorem}
\label{thm:convexbelief}
For any POMDP, the value-function 
$\valuefunc$ is convex.
\end{theorem}

\subsection{Finite Exploration Approximation}
One way to circumvent building the complete state space is to \emph{cut-off} its exploration after some steps, much like we depicted part of the belief POMDP in \Cref{fig:bmdp}.
To ensure that the obtained finite MDP over-approximates the probability to reach a bad state, we simply assume that all transitions we cut go to a bad state immediately.
Elaborate techniques for this approach (on general MDPs) have been discussed in the context of verification~\cite{DBLP:conf/atva/BrazdilCCFKKPU14}, and have been successfully adapted to other models~\cite{DBLP:conf/atva/AshokBHK18,DBLP:journals/tii/VolkJK18,DBLP:conf/atva/0001DKKW16}. It shares many ideas with the SARSOP and GOAL-HSVI approaches for POMDPs~\cite{DBLP:conf/rss/KurniawatiHL08,DBLP:conf/ijcai/HorakBC18}.
This approach may be applied directly to belief MDPs, and we may use the POMDP $\pomdp$ to guide the cut-off process. 
In particular, using \Cref{thm:convexbelief} and that the maximising policy over \emph{all} policies is necessarily overapproximating the maximum over all \emph{observation-based} policies, we obtain the following inequality:
\begin{align}
\valueof{\beliefstate} \quad\leq\quad \sum_{\state \in \states} \beliefstate(\state) \cdot \valueof{ \{ \state \mapsto 1 \} } \quad\leq\quad \sum_{\state \in \states} \beliefstate(\state) \cdot \sup_{\sched \in \scheds{\pomdp}} \pr[\state]{\pomdp}{\sched}{\bad}	\label{eq:overapprox}
\end{align}
We may use this inequality to cut-off with a less pessimistic value than assuming that we reach the bad states with probability one.

Nevertheless, this approach has limited applicability on its own.
 It may well get stuck in regions of the belief space that are not near the goal. 
 From state $s_5, s_6$ in \Cref{fig:pomdp} the maximal reachability according to the underlying MDP is $1$, which is too pessimistic to provide a good cut-off. 
 Another issue is that the belief converges slowly along $\beliefstate_1, \beliefstate_5, \beliefstate_9$ in \Cref{fig:bmdp}, and that cut-offs do not immediately allow to reason that the belief converged.  

\subsection{Discretised Belief Approximation}
The idea of this approach is to select a finite set $\foundation \subseteq \beliefstates$ of beliefs, 
and construct an approximation of the belief MDP using only $\foundation$ as states.
We refer to $\foundation$ as the \emph{foundation}.
(Reachable) beliefs $\beliefstate$ not in $\foundation$ are approximated using beliefs in $\neighbourhood[\foundation]{\beliefstate}$, where $\neighbourhood[\foundation]{\beliefstate} \subseteq \foundation$ is the  \emph{neighbourhood} of $\beliefstate$. We clarify the selection of these neighbourhoods later, and we omit the subscript $\foundation$ whenever possible.
\begin{definition}
\label{def:convexneighbourhood}
A neighbourhood $\neighbourhood{\beliefstate}$ of belief $\beliefstate$
is \emph{convex-containing}, if there exists $\vertexweight{\beliefstate} \in \dists{\neighbourhood{\beliefstate}}$ such that $\beliefstate = \sum_{\beliefstate' \in \neighbourhood{\beliefstate}}  \vertexweight{\beliefstate}(\beliefstate') \cdot \beliefstate'$.   
\end{definition}
\begin{figure}[t]
\centering
\subfigure[]{
\scalebox{0.95}{
\begin{tikzpicture}[scale=0.7]
	\draw[dotted] (0,0) -- (0,2);
	
	\draw[dashed,thick] (0,0.666) -- (0,0);
	\draw[dashed,thick] (0,0.666) -- (0,1);

	\node[anchor=east] at (0,0) {$s_3$};
	\node[anchor=east] at (0,2) {$s_4$};

	\node [diamond,fill=black,inner sep=1pt] at (0,2) {};
	\node [diamond,fill=red,inner sep=1pt] at (0,0) {};
	\node [diamond,fill=red,inner sep=1pt] at (0,1) {};

	\node [star,fill=blue!40!white,draw=blue,very thick,inner sep=1pt,scale=1.4] at (0,0.666) {};

\end{tikzpicture}
}
\label{fig:2dbelief}
}
\subfigure[]{
\scalebox{0.95}{
\begin{tikzpicture}[scale=0.7]
	\draw[dotted] (0,0) -- (0,2);
	\draw[dotted] (0,0) -- (2,0);
	\draw[dotted] (0,2) -- (2,0);
	\draw[dashed,thick] (0.333,0.666) -- (0,0);
	\draw[dashed,thick] (0.333,0.666) -- (0,2);
	\draw[dashed,thick] (0.333,0.666) -- (2,0);

\node [star,fill=blue!40!white,draw=blue,very thick,inner sep=1pt,scale=1.4] at (0.333,0.666) {};
	\node [diamond,fill=red,inner sep=1pt] at (0,2) {};
	\node [diamond,fill=red,inner sep=1pt] at (2,0) {};
	\node [diamond,fill=red,inner sep=1pt] at (0,0) {};
	
	\node[anchor=east] at (0,0) {$s_0$};
	\node[anchor=west] at (2,0) {$s_5$};
	\node[anchor=east] at (0,2) {$s_6$};
	
\end{tikzpicture}
}
\label{fig:boring3dbelief}
}
%
%
%
%
%
%
\subfigure[]{
\scalebox{0.95}{
\begin{tikzpicture}[scale=0.7]
	
	\draw[dotted] (0,0) -- (0,2);
	\draw[dotted] (0,0) -- (2,0);
	\draw[dotted] (0,2) -- (2,0);
	\draw[dotted] (0,0) -- (0.5,1.5);
	\draw[dashed,thick] (0.333,0.666) -- (0,0);
	\draw[dashed,thick] (0.333,0.666) -- (0.5,1.5);
	\draw[dashed,thick] (0.333,0.666) -- (2,0);

	\node[anchor=east] at (0,0) {$s_0$};
	\node[anchor=west] at (2,0) {$s_5$};
	\node[anchor=east] at (0,2) {$s_6$};

	\node [star,fill=blue!40!white,draw=blue,very thick,inner sep=1pt,scale=1.4] at (0.333,0.666) {};
	\node [diamond,fill=black,inner sep=1pt] at (0,2) {};
	\node [diamond,fill=red,inner sep=1pt] at (2,0) {};
	\node [diamond,fill=red,inner sep=1pt] at (0.5,1.5) {};
	\node [diamond,fill=red,inner sep=1pt] at (0,0) {};

\end{tikzpicture}
}
\label{fig:3dbelief}
}
\subfigure[]{
\scalebox{0.95}{
\begin{tikzpicture}[scale=0.7]
	\draw[dotted] (0,0) -- (0,2);
	\draw[dotted] (0,0) -- (2,0);
	\draw[dotted] (0,2) -- (2,0);
	\draw[dotted] (0,1) -- (1,0);
	\draw[dotted] (0,1) -- (1,1);
	\draw[dotted] (1,0) -- (1,1);
	
	\draw[dashed,thick] (0.3333,0.666) -- (1,0);
	\draw[dashed,thick] (0.3333,0.666) -- (0,1);
	
	\node[anchor=north] at (0,0) {$s_0$};
	\node[anchor=north] at (2,0) {$s_5$};
	\node[anchor=east] at (0,2) {$s_6$};
	
	\node [diamond,fill=black,inner sep=1pt] at (0,2) {};
	\node [diamond,fill=black,inner sep=1pt] at (2,0) {};
	\node [diamond,fill=black,inner sep=1pt] at (0,0) {};
	
	\node [diamond,fill=red,inner sep=1pt] at (0,1) {};
	\node [diamond,fill=red,inner sep=1pt] at (1,0) {};
	\node [diamond,fill=black,inner sep=1pt] at (1,1) {};
	
	\node [star,fill=blue!40!white,draw=blue,very thick,inner sep=1pt,scale=1.4] at (0.3333,0.666) {};

\end{tikzpicture}
}
\label{fig:regulartriangulization}
}
\caption{Belief-spaces with foundation (diamonds), a belief state (blue star), a fixed neighbourhood (red diamonds), and vertex-weights. }
\label{fig:neighbourhoods}	
\end{figure}
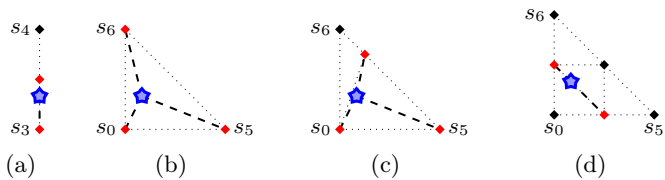
\begin{example}
\label{ex:neighbourhoods}
In \Cref{fig:neighbourhoods}, we depict various neighbourhoods.
In \Cref{fig:2dbelief}, the belief $\{ s_3 \mapsto \nicefrac{2}{3}, s_4 \mapsto \nicefrac{1}{3} \}$ lies in the neighbourhood $\big\{\{ s_3 
\mapsto 1,\} \{ s_3, s_4 \mapsto \nicefrac{1}{2}\} \big\}$.
All other subfigures depict belief-spaces for observations where three states have this observation (the third dimension implicitly follows).   
For the belief state $\beliefstate_5 = \{ s_0 \mapsto \nicefrac{1}{2},\ s_5 \mapsto \nicefrac{1}{6},\ s_6 \mapsto \nicefrac{1}{3} \}$ from \Cref{fig:bmdp} and a neighbourhood as in \Cref{fig:boring3dbelief}, the vertex-weights $\vertexweight{\beliefstate}$ follow straightforwardly from the belief. Observe that a small distance to a vertex induces a large weight. In \Cref{fig:3dbelief}, we adapt the neighbourhood to $x = \{ s_5 \mapsto 1 \}, y = \{ s_0 \mapsto 1 \}, z = \{ s_5 \mapsto \nicefrac{1}{4},\ s_6 \mapsto \nicefrac{3}{4} \}$. Then, the vertex weights follow from the following linear equations:
\[ \vertexweight{\beliefstate_5}(x)  = \nicefrac{1}{2}, \quad \vertexweight{\beliefstate_5}(y) + \nicefrac{1}{4} \cdot \vertexweight{\beliefstate_5}(z) =  \nicefrac{1}{6}, \quad\text{and }
\nicefrac{3}{4}     \cdot \vertexweight{\beliefstate_5}(z) = \nicefrac{1}{3}. \]
\end{example}

\noindent
From the convexity of the value function $\valuefunc$~(\Cref{thm:convexbelief}), it follows that: 
\begin{lemma}
\label{lem:convexapprox}
Given $\beliefstate$, $\neighbourhood{\beliefstate}$ and $\vertexweight{\beliefstate}$ as in \Cref{def:convexneighbourhood}, it holds:
\[ \valueof{\beliefstate} \quad\leq\quad \sum_{\beliefstate' \in \neighbourhood{\beliefstate}} \vertexweight{\beliefstate}(\beliefstate') \cdot \valueof{\beliefstate'}. \]
\end{lemma}
We emphasise that this inequality also holds if one over-approximates the values of the beliefs in the neighbourhood.
\begin{example}
\Cref{fig:neighbourhoodapproxcoarse} depicts the belief-to-value from \Cref{fig:belieftovalue1} and (in blue) depicts the over-approximation based on \Cref{lem:convexapprox}. 
As neighbourhood, we use $\{ s_3 \mapsto 1\}$ and $\{ s_4 \mapsto 1 \}$. 
In \Cref{fig:neighbourhoodapproxfine}, we depict the over-approximation using a partitioning into three neighbourhoods, using the foundation $\{ s_3 \mapsto 1\}$, $\{ s_3 \mapsto \nicefrac{1}{4},\ s_4 \mapsto \nicefrac{3}{4} \}$, $\{ s_3 \mapsto \nicefrac{3}{4},\ s_4 \mapsto \nicefrac{1}{4} \}$ and $\{ s_4 \mapsto 1 \}$. We see that the outer neighbourhoods now yield a tight over-approximation, and the inner neighbourhood yields a much better approximation compared to \Cref{fig:neighbourhoodapproxcoarse}.
\end{example}
We select some finite foundation $\foundation$ such that for each reachable $\beliefstate$ in $\beliefmdp{\pomdp}$, there exists a convex containing neighbourhood $\neighbourhood{\beliefstate}$. 
We call such a foundation \emph{adequate}.
One small adequate $\foundation$ is $\{ \{ s \mapsto 1 \} \in \beliefstates \mid s \in \pomdp \}$. Practically, we use a tiling of the belief space into convex hyper-triangles, see below.
\begin{definition}[Discretised Belief MDP]
	\label{def:discbelief}
	Let $\foundation \subseteq \beliefstates$ be an adequate foundation.
	Let $\neighbourhoodfunc$ be arbitrarily fixed such that $\neighbourhood{\beliefstate} \subseteq \foundation$ is convex-containing for any $\beliefstate$. The \emph{discretised belief MDP} of POMDP $\pomdp = \pomdptuple$ is the MDP $\discbeliefmdp{\pomdp}{\foundation} \colonequals \tuple{\foundation, \actions, \discbelieftransitions, \binit}$ with initial belief state $\binit = \{ \sinit \mapsto 1 \}$, and---using the auxiliary notation from before \Cref{def:beliefmdp}---transition function $\discbelieftransitions$ 
	given by 
	\[ \discbelieftransitions(\beliefstate,\action,\discbeliefstate) \colonequals 
	\begin{cases} \vertexweight{\nextbelief{\beliefstate}{\action}{\observation}}(\discbeliefstate) \cdot \belieftransitions(\beliefstate, \action, \observation) & \text{if } \discbeliefstate \in \neighbourhood{\nextbelief{\beliefstate}{\action}{\observation}}, \\
 0 & \text{otherwise,}	
 \end{cases}	\]

\end{definition}
\begin{figure}[t]
\centering
\subfigure{
\scalebox{0.9}{
	\begin{tikzpicture}[st/.style={circle,font=\footnotesize,draw,inner sep=1pt},act/.style={rectangle,font=\footnotesize,fill=black,inner sep=1pt},gst/.style={circle,scale=1.3,font=\small,color=green!60!black,inner sep=0pt},bst/.style={circle,scale=1.3,font=\small,color=red!60!black,inner sep=0pt}]
		\node[st,initial,initial text=,initial where=above] (b1) {$\beliefstate_1$};
		
		\node[above=0.4cm of b1]  (up) {};
		\node[below=0.4cm of b1]  (down) {};
		\node[st,right=0.8cm of up,fill=RWTHorange!70] (b2) {$\beliefstate_2$};
		\node[st,right=0.8cm of down,fill=RWTHorange!70] (b3) {$\beliefstate_3$};
		\node[st,right=2.3cm of b1,fill=RWTHmaygreen!70] (b5) {$\beliefstate_5$};
		\node[st,above=1cm of b5,fill=RWTHmaygreen!70] (b4) {$\beliefstate_4$};
		\node[st,right=2.3cm of b1,fill=RWTHmaygreen!70] (b5) {$\beliefstate_5$};
		\node[st,below=1cm of b5,fill=RWTHmaygreen!70] (b6) {$\beliefstate_6$};

		\node[st,left=1.2cm of up,yshift=0.4cm] (b7) {$\beliefstate_7$};
		\node[st,left=1.2cm of down,yshift=-0.4cm] (b8) {$\beliefstate_8$};
		\node[st,left=2.2cm of b1] (b9) {$\beliefstate_9$};
		
		\node[act] (a7b) [left=0.8cm of b7] {};
		\node[act] (a8b) [above=0.4cm of b8] {};
		\node[act] (a7a) [right=0.3cm of b7] {};
		\node[act] (a8a) [right=0.3cm of b8] {};
		
		\node[act,left=0.6cm of b9] (a9a) {};
		\node[act,above=0.5cm of b9,xshift=0.5cm] (a9b) {};

		\node[act,right=0.3cm of b1] (a1a) {};
		\node[act,left=0.3cm of b1] (a1b) {};
		
		\node[act,right=0.3cm of b2,yshift=0.6cm] (a2a) {};
		\node[act,right=0.3cm of b2,yshift=-0.2cm] (a2b) {};
		
		\node[act,right=0.3cm of b3,yshift=0.0cm] (a3a) {};
		\node[act,right=0.3cm of b3,yshift=-0.6cm] (a3b) {};
		
		\node[act,right=0.3cm of b4,yshift=0.2cm] (a4a) {};
		\node[act,right=0.3cm of b4,yshift=-0.4cm] (a4b) {};
		
		\node[act,right=0.3cm of b5,yshift=0.2cm] (a5a) {};
		\node[act,right=0.3cm of b5,yshift=-0.4cm] (a5b) {};
		
		\node[act,right=0.3cm of b6,yshift=0.2cm] (a6a) {};
		\node[act,right=0.3cm of b6,yshift=-0.4cm] (a6b) {};
		
		\node[gst,right=1.2cm of b4] (gr) {\faSmileO};
		\node[bst,right=1.2cm of b6] (br) {\faFrownO};
		
		\draw[-] (b1) edge node[above] {$a$} (a1a);
		\draw[-] (b1) edge node[above] {$b$} (a1b);
		\draw[-] (b2) edge node[above] {$a$} (a2a);
		\draw[-] (b2) edge node[below] {$b$} (a2b);
		
		\draw[-] (b3) edge node[above] {$a$} (a3a);
		\draw[-] (b3) edge node[below] {$b$} (a3b);
		
		\draw[-] (b4) edge node[above] {$b$} (a4a);
		\draw[-] (b4) edge node[below] {$a$} (a4b);
		
		\draw[-] (b5) edge node[above] {$a$} (a5a);
		\draw[-] (b5) edge node[below] {$b$} (a5b);
		\draw[-] (b6) edge node[above] {$a$} (a6a);
		\draw[-] (b6) edge node[below] {$b$} (a6b);
		
		\draw[-] (b7) edge node[above] {$a$} (a7a);
		\draw[-] (b7) edge node[pos=0.2,above] {$b$} (a7b);
		\draw[-] (b8) edge node[below] {$a$} (a8a);
		\draw[-] (b8) edge node[right] {$b$} (a8b);
		\draw[-] (b9) edge node[above] {$a$} (a9a);
		\draw[-] (b9) edge node[left] {$b$} (a9b);

		\draw[->] (a1a) edge[bend left] node[below] {$\nicefrac{1}{5}$} (b1);
		\draw[->] (a1a) edge node[above] {$\nicefrac{3}{5}$} (b2);
		\draw[->] (a1a) edge node[above] {$\nicefrac{1}{5}$} (b3);
		
		\draw[->] (a2a) edge node[above] {$1$} (b4);
		\draw[->] (a2b) edge node[pos=0.3,above] {$\nicefrac{1}{3}$} (b4);
		\draw[->] (a2b) edge node[above] {$\nicefrac{2}{3}$} (b5);
		\draw[->] (a3a) edge[bend right=10] node[pos=0.3,left] {$\nicefrac{1}{2}$} (b4);
		\draw[->] (a3a) edge node[pos=0.3,right] {$\nicefrac{1}{2}$} (b5);
		\draw[->] (a3b) edge node[above] {$1$} (b6);
		
		\draw[->] (a4a) edge[bend left=20] node[above] {$1$} (gr);
		
		\draw[->] (a4b) edge[out=30,in=180,looseness=0.9] node[pos=0.3,above] {$\nicefrac{2}{5}$} (gr);
		\draw[->] (a4b) edge[bend left=40] node[pos=0.15,above] {$\nicefrac{3}{5}$} (br);
		\draw[->] (a5a) edge[bend left=5] node[pos=0.3,left] {$\nicefrac{23}{40}$} (gr);
		\draw[->] (a5a) to[out=350,in=90,looseness=1.3] node[pos=0.2,above] {$\nicefrac{17}{40}$} (br);

		\draw[->] (a5b) edge[out=10,in=270,looseness=1.4] node[pos=0.05,above] {$\nicefrac{1}{2}$} (gr);
		\draw[->] (a5b) edge[bend right=5] node[pos=0.3,left] {$\nicefrac{1}{2}$} (br);
		\draw[->] (a6a) edge[bend right=40] node[pos=0.16,below] {$\nicefrac{3}{4}$} (gr);
		\draw[->] (a6a) to[out=330,in=180,looseness=0.9] node[pos=0.2,below] {$\nicefrac{1}{4}$} (br);

		\draw[->] (a6b) edge[bend right=20] node[below] {$1$} (br);

		\draw[->] (a7a) -- node[above] {$1$} (b2);
		\draw[->] (a8a) -- node[above] {$1$} (b3);
		\draw[->] (a9a) -- node[pos=0.7,right] {$\nicefrac{1}{4}$} +(0,1.6) -|  (b2);
		\draw[->] (a9a) -- node[pos=0.7,right] {$\nicefrac{3}{4}$} +(0,-1.6) -| (b3);
		
		\draw[->] (a1b) edge[bend right] node[below] {$\nicefrac{1}{2}$} (b1);
		\draw[->] (a1b) edge node[pos=0.7,right] {$\nicefrac{1}{18}$} (b7);
		\draw[->] (a1b) edge node[pos=0.25,below] {$\nicefrac{4}{9}$} (b9);
	
		\draw[->] (a7b) edge node[pos=0.3,left] {$1$} (b9);
		\draw[->] (a8b) edge node[pos=0.6,right] {$\nicefrac{1}{3}$} (b7);
		\draw[->] (a8b) edge node[pos=0.3,below] {$\nicefrac{2}{3}$} (b9);
		\draw[->] (a9b) edge node[pos=0.7,below] {$\nicefrac{5}{12}$} (b7);
		\draw[->] (a9b) edge[bend left] node[right] {$\nicefrac{7}{12}$} (b9);
	\end{tikzpicture}
	}}
	\subfigure{
		\scalebox{0.9}{
	\raisebox{7em}{
		\begin{tabular}{ll}
 $\beliefstate_1$: & $\{ s_0 \mapsto 1 \}$ \\
 $\beliefstate_2$: & $\{ s_1 \mapsto 1 \}$ \\
 $\beliefstate_3$: & $\{ s_2 \mapsto 1 \}$ \\
  $\beliefstate_4$: & $\{ s_3 \mapsto 1 \}$ \\
  $\beliefstate_5$: & $\{ s_3,s_4 \mapsto \nicefrac{1}{2}\}$ \\
  $\beliefstate_6$: & $\{ s_4 \mapsto 1 \}$ \\
  $\beliefstate_7$: & $\{ s_5 \mapsto 1 \}$ \\
  $\beliefstate_8$: & $\{ s_6 \mapsto 1 \}$ \\
  $\beliefstate_9$: & $\{ s_5 \mapsto \nicefrac{1}{4} $, \\
  					& $\quad s_6 \mapsto \nicefrac{3}{4} \}$  
\end{tabular}
	}
	}
	}
\caption{Reachable fragment of the discretised belief MDP (fully observable). Actual beliefs are given in the table on the right. Colours indicate $\obsof{\beliefstate_i}$ in the POMDP. }
\label{fig:dbmdp}
\end{figure}
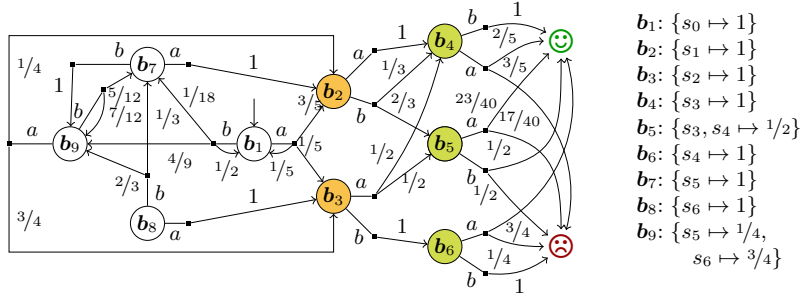
\begin{example}
Consider \Cref{fig:dbmdp}.
We fixed $\foundation = \{ s \mapsto 1 \mid s \in \states \} \cup \{ s_3, s_4 \mapsto \nicefrac{1}{2} \} \cup \{ s_5 \mapsto \nicefrac{1}{4},\ s_6 \mapsto \nicefrac{3}{4} \}$.
The weights for $\post[]{\beliefstate_2}{b}$ and $\post[]{\beliefstate_1}{b}$ follow from the computations in \Cref{ex:neighbourhoods}.
Observe that $\beliefstate_8$ is not reachable. The optimal policy in this MDP induces probability $\nicefrac{3}{4}$, which is an upper bound on $\valueof{\beliefstate_1}$.
\end{example}

\begin{theorem}
For POMDP $\pomdp$ with discretised belief MDP $\discbeliefmdp{\pomdp}{\foundation}$ and $\beliefstate \in \foundation$
	\[ \valueof{\beliefstate} \quad\leq\quad \sup_{\sched \in \scheds{\discbeliefmdp{\pomdp}{\foundation}}} \pr[\beliefstate]{\discbeliefmdp{\pomdp}{\foundation}}{\sched}{\badbeliefs}.	 \]
\end{theorem}
As the MDP is finite and fully observable, the supremum is achieved by a memoryless policy, and we use MDP model checking to compute these values.

\section{Abstraction-Refinement}
\label{sec:algorithm}

In this section, we discuss a framework that combines the two types of abstraction discussed before. 
Roughly, the approach is a typical abstraction-refinement loop. 
We start with an abstraction of the belief MDP; model checking this abstraction yields an upper bound on the values $\valueof{\beliefstate}$. In every iteration, we update the MDP and then obtain more and more accurate bounds.
The abstraction applies cut-offs on a discretised belief MDP with some foundation $\foundation$.
For the refinement, we either explore beliefs that were previously cut off, we extend the foundation $\foundation$, or we rewire the successors $\beliefstate' \in \post[\beliefmdp{\pomdp}]{\beliefstate}{\action}$ of some belief $\beliefstate$ and action $\action$ to a new $\neighbourhood[\foundation]{\beliefstate'}$. Thus, rewiring updates neighbourhoods, typically after refining the foundation.
We give an example and then clarify the precise procedure, along with some technical details.
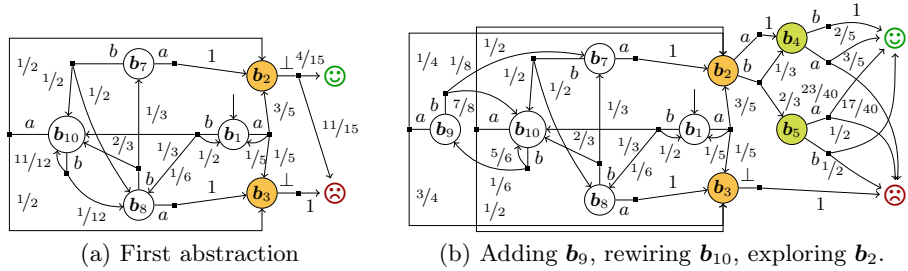
\begin{figure}[t]
\subfigure[First abstraction]{
\scalebox{0.8}{
\begin{tikzpicture}[st/.style={circle,font=\footnotesize,draw,inner sep=1pt},act/.style={rectangle,font=\footnotesize,fill=black,inner sep=1pt},gst/.style={circle,scale=1.3,font=\small,color=green!60!black,inner sep=0pt},bst/.style={circle,scale=1.3,font=\small,color=red!60!black,inner sep=0pt}]
		\node[st,,initial,initial text=,initial where=above] (b1) {$\beliefstate_1$};
		
		\node[above=0.6cm of b1]  (up) {};
		\node[below=0.6cm of b1]  (down) {};
		\node[st,right=0.1cm of up,fill=RWTHorange!70] (b2) {$\beliefstate_2$};
		\node[st,right=0.1cm of down,fill=RWTHorange!70] (b3) {$\beliefstate_3$};

		\node[st,left=1.2cm of up,yshift=0.2cm] (b7) {$\beliefstate_7$};
		\node[st,left=1.2cm of down,yshift=-0.2cm] (b8) {$\beliefstate_8$};
		\node[st,left=2.2cm of b1] (b9) {$\beliefstate_{10}$};
		
		\node[act] (a7b) [left=0.8cm of b7] {};
		\node[act] (a8b) [above=0.3cm of b8] {};
		\node[act] (a7a) [right=0.3cm of b7] {};
		\node[act] (a8a) [right=0.3cm of b8] {};
		
		\node[act,left=0.6cm of b9] (a9a) {};
		\node[act,below=0.3cm of b9] (a9b) {};

		\node[act,right=0.3cm of b1] (a1a) {};
		\node[act,left=0.3cm of b1] (a1b) {};

		\node[act,right=0.3cm of b2] (a2c) {};
		
		\node[act,right=0.3cm of b3] (a3c) {};
		\draw[-] (b3) edge node[above] {$\bot$} (a3c);
		
		\draw[-] (b2) edge node[above] {$\bot$} (a2c);

		\node[gst,right=0.7cm of b2] (gr) {\faSmileO};
		\node[bst,right=0.7cm of b3] (br) {\faFrownO};
		
		\draw[-] (b1) edge node[above] {$a$} (a1a);
		\draw[-] (b1) edge node[above] {$b$} (a1b);

		\draw[-] (b7) edge node[above] {$a$} (a7a);
		\draw[-] (b7) edge node[pos=0.2,above] {$b$} (a7b);
		\draw[-] (b8) edge node[below] {$a$} (a8a);
		\draw[-] (b8) edge node[right,pos=0.6] {$b$} (a8b);
		\draw[-] (b9) edge node[above] {$a$} (a9a);
		\draw[-] (b9) edge node[right] {$b$} (a9b);
		
		\draw[->] (a2c) edge node[right] {$\nicefrac{11}{15}$} (br);
		\draw[->] (a2c) edge node[above] {$\nicefrac{4}{15}$} (gr);
		
		\draw[->] (a3c) edge node[below] {$1$} (br);
	
		\draw[->] (a1a) edge[bend left] node[below,pos=0.6] {$\nicefrac{1}{5}$} (b1);
		\draw[->] (a1a) edge node[right] {$\nicefrac{3}{5}$} (b2);
		\draw[->] (a1a) edge node[right] {$\nicefrac{1}{5}$} (b3);
		
				\draw[->] (a7a) -- node[above] {$1$} (b2);
		\draw[->] (a8a) -- node[above] {$1$} (b3);
		\draw[->] (a9a) -- node[pos=0.7,right] {$\nicefrac{1}{2}$} +(0,1.6) -|  (b2);
		\draw[->] (a9a) -- node[pos=0.7,right] {$\nicefrac{1}{2}$} +(0,-1.6) -| (b3);
		
		\draw[->] (a1b) edge[bend right] node[below] {$\nicefrac{1}{2}$} (b1);
		\draw[->] (a1b) edge node[pos=0.7,right] {$\nicefrac{1}{6}$} (b8);
		\draw[->] (a1b) edge node[pos=0.25,below] {$\nicefrac{1}{3}$} (b9);
		\draw[->] (a7b) edge[bend right=10] node[pos=0.2,right] {$\nicefrac{1}{2}$} (b8);
		\draw[->] (a7b) edge node[pos=0.3,left] {$\nicefrac{1}{2}$} (b9);
		\draw[->] (a8b) edge node[pos=0.6,right] {$\nicefrac{1}{3}$} (b7);
		\draw[->] (a8b) edge node[pos=0.25,above] {$\nicefrac{2}{3}$} (b9);
		\draw[->] (a9b) edge[bend right] node[below] {$\nicefrac{1}{12}$} (b8);
		\draw[->] (a9b) edge[bend left] node[left] {$\nicefrac{11}{12}$} (b9);
	\end{tikzpicture}
}
\label{fig:approxmdp}
}
\subfigure[Adding $\beliefstate_9$, rewiring $\beliefstate_{10}$, exploring $\beliefstate_2$.]{
\scalebox{0.8}{
\begin{tikzpicture}[st/.style={circle,font=\footnotesize,draw,inner sep=1pt},act/.style={rectangle,font=\footnotesize,fill=black,inner sep=1pt},gst/.style={circle,scale=1.3,font=\small,color=green!60!black,inner sep=0pt},bst/.style={circle,scale=1.3,font=\small,color=red!60!black,inner sep=0pt}]
		\node[st,,initial,initial text=,initial where=above] (b1) {$\beliefstate_1$};
		
		\node[above=0.6cm of b1]  (up) {};
		\node[below=0.6cm of b1]  (down) {};
		\node[st,right=0.1cm of up,fill=RWTHorange!70] (b2) {$\beliefstate_2$};
		\node[st,right=0.1cm of down,fill=RWTHorange!70] (b3) {$\beliefstate_3$};
		\node[st,right=1.1cm of b1,fill=RWTHmaygreen!70] (b5) {$\beliefstate_5$};
		\node[st,above=1cm of b5,fill=RWTHmaygreen!70] (b4) {$\beliefstate_4$};

		\node[st,left=1.2cm of up,yshift=0.2cm] (b7) {$\beliefstate_7$};
		\node[st,left=1.2cm of down,yshift=-0.2cm] (b8) {$\beliefstate_8$};
		\node[st,left=2.2cm of b1] (b10) {$\beliefstate_{10}$};
		
		\node[st,left=0.8cm of b9] (b9) {$\beliefstate_{9}$};
		
		\node[act] (a7b) [left=0.8cm of b7] {};
		\node[act] (a8b) [above=0.3cm of b8] {};
		\node[act] (a7a) [right=0.3cm of b7] {};
		\node[act] (a8a) [right=0.3cm of b8] {};
		
		\node[act,left=0.5cm of b10] (a10a) {};
		\node[act,below=0.3cm of b10] (a10b) {};

		\node[act,left=0.3cm of b9] (a9a) {};
		\node[act,above=0.3cm of b9] (a9b) {};
		
		\node[act,right=0.3cm of b1] (a1a) {};
		\node[act,left=0.3cm of b1] (a1b) {};
		
		\node[act,right=0.3cm of b2,yshift=0.6cm] (a2a) {};
		\node[act,right=0.3cm of b2,yshift=-0.2cm] (a2b) {};
		
		\node[act,right=0.3cm of b3,yshift=0.0cm] (a3c) {};
		
		\node[act,right=0.3cm of b4,yshift=0.2cm] (a4a) {};
		\node[act,right=0.3cm of b4,yshift=-0.4cm] (a4b) {};
		
		\node[act,right=0.3cm of b5,yshift=0.2cm] (a5a) {};
		\node[act,right=0.3cm of b5,yshift=-0.4cm] (a5b) {};

		\node[gst,right=1.2cm of b4] (gr) {\faSmileO};
		\node[bst,below=2.1cm of gr] (br) {\faFrownO};
		
		\draw[-] (b1) edge node[above] {$a$} (a1a);
		\draw[-] (b1) edge node[above] {$b$} (a1b);
		\draw[-] (b2) edge node[above] {$a$} (a2a);
		\draw[-] (b2) edge node[above] {$b$} (a2b);

		\draw[-] (b4) edge node[above] {$b$} (a4a);
		\draw[-] (b4) edge node[below] {$a$} (a4b);
		
		\draw[-] (b5) edge node[above] {$a$} (a5a);
		\draw[-] (b5) edge node[below] {$b$} (a5b);
		
		\draw[-] (b7) edge node[above] {$a$} (a7a);
		\draw[-] (b7) edge node[pos=0.2,below] {$b$} (a7b);
		\draw[-] (b8) edge node[below] {$a$} (a8a);
		\draw[-] (b8) edge node[pos=0.6,right] {$b$} (a8b);
		
		\draw[-] (b9) edge node[above] {$a$} (a9a);
		\draw[-] (b9) edge node[left] {$b$} (a9b);
		\draw[-] (b10) edge node[above] {$a$} (a10a);
		\draw[-] (b10) edge node[right] {$b$} (a10b);
		
		\draw[-] (b3) edge node[above] {$\bot$} (a3c);
		\draw[->] (a3c) edge node[below] {$1$} (br);

		\draw[->] (a1a) edge[bend left] node[below,pos=0.7] {$\nicefrac{1}{5}$} (b1);
		\draw[->] (a1a) edge node[pos=0.5,right] {$\nicefrac{3}{5}$} (b2);
		\draw[->] (a1a) edge node[right] {$\nicefrac{1}{5}$} (b3);
		
		\draw[->] (a2a) edge node[above] {$1$} (b4);
		\draw[->] (a2b) edge node[pos=0.3,right] {$\nicefrac{1}{3}$} (b4);
		\draw[->] (a2b) edge node[right] {$\nicefrac{2}{3}$} (b5);
				
		\draw[->] (a4a) edge[bend left=20] node[above] {$1$} (gr);
		
		\draw[->] (a4b) edge[out=30,in=180,looseness=0.9] node[pos=0.3,above] {$\nicefrac{2}{5}$} (gr);
		\draw[->] (a4b) edge[bend left=35] node[pos=0.18,above] {$\nicefrac{3}{5}$} (br);
		\draw[->] (a5a) edge[bend left=5] node[pos=0.3,left] {$\nicefrac{23}{40}$} (gr);
		\draw[->] (a5a) to[out=350,in=90,looseness=1.3] node[pos=0.25,above] {$\nicefrac{17}{40}$} (br);

		\draw[->] (a5b) edge[out=10,in=270,looseness=1.4] node[pos=0.05,above] {$\nicefrac{1}{2}$} (gr);
		\draw[->] (a5b) edge[bend right=5] node[pos=0.4,left] {$\nicefrac{1}{2}$} (br);

		\draw[->] (a7a) -- node[above] {$1$} (b2);
		\draw[->] (a8a) -- node[above] {$1$} (b3);
		\draw[->] (a10a) -- node[pos=0.8,right] {$\nicefrac{1}{2}$} +(0,1.7) -|(b2);
		\draw[->] (a10a) -- node[pos=0.8,right] {$\nicefrac{1}{2}$} +(0,-1.7) -| (b3);
		
		\draw[->] (a1b) edge[bend right] node[below] {$\nicefrac{1}{2}$} (b1);
		\draw[->] (a1b) edge node[pos=0.7,right] {$\nicefrac{1}{6}$} (b8);
		\draw[->] (a1b) edge node[pos=0.25,below] {$\nicefrac{1}{3}$} (b10);
		\draw[->] (a7b) edge[bend right=10] node[pos=0.2,right] {$\nicefrac{1}{2}$} (b8);
		\draw[->] (a7b) edge node[pos=0.3,left] {$\nicefrac{1}{2}$} (b10);
		\draw[->] (a8b) edge node[pos=0.6,right] {$\nicefrac{1}{3}$} (b7);
		\draw[->] (a8b) edge node[pos=0.25,above] {$\nicefrac{2}{3}$} (b10);
		\draw[->] (a10b) edge[bend left] node[pos=0.3,below] {$\nicefrac{1}{6}$} (b9);
		\draw[->] (a10b) edge[bend left] node[left] {$\nicefrac{5}{6}$} (b10);
		
		\draw[->] (a9a) -- node[pos=0.7,right] {$\nicefrac{1}{4}$} +(0,1.6) -|  (b2);
		\draw[->] (a9a) -- node[pos=0.7,right] {$\nicefrac{3}{4}$} +(0,-1.6) -| (b3);
		\draw[->] (a9b) edge[bend left=25] node[pos=0.1,above] {$\nicefrac{1}{8}$} (b7);
		\draw[->] (a9b) edge[bend left] node[pos=0.2,below] {$\nicefrac{7}{8}$} (b10);

	\end{tikzpicture}
}
\label{fig:approxmdp2}
}
\caption{Beliefs as in in \Cref{fig:dbmdp}, with $\beliefstate_{10} = \{ s_5 \mapsto \nicefrac{1}{2}, s_6 \mapsto \nicefrac{1}{2} \}$.} 
\label{fig:approx}
\end{figure}
\begin{example}
    In \Cref{fig:approxmdp}, we used a foundation as in \Cref{fig:dbmdp}, but with $\beliefstate_{10}$ replacing $\beliefstate_{9}$. Furthermore, 
    we used cut-offs in $\beliefstate_2$ and $\beliefstate_3$ with the overapproximation from Eq.~\eqref{eq:overapprox}. 
	In \Cref{fig:approxmdp2} we refined as follows: We \emph{extended the foundation} with  $\beliefstate_9 = \{ s_5 \mapsto \nicefrac{1}{4}, s_6 \mapsto \nicefrac{3}{4} \}$, we \emph{explored} from $\beliefstate_2, \beliefstate_9$, 
 and we rewired \emph{only} $\tuple{\beliefstate_{10},b}$.
\end{example}
\begin{algorithm}[t]
\SetKwFunction{Explore}{\color{blue!50}explore\color{black}}
\SetKwFunction{Rewire}{\color{blue!50}{rewire}\color{black}}
\SetKwFunction{Extend}{extend}
\SetKwFunction{Cutoff}{cutoff}
\scriptsize{
	\Input{POMDP $\pomdp = \pomdptuple$ with $\mdp = \mdptuple$, bad beliefs $\badbeliefs$, threshold $\lambda$}
	\Output{An upper bound $\lambda \geq \uppervalueboundof{\binit} \geq \valueof{\binit}$}
	
	$\foundation \gets \text{initial adequate foundation}$\label{line:initf}\\
	$\abstbeliefmdp \gets$ MDP $\tuple{\abstbeliefstates, \actions, \abstbelieftransitions, \binit}$ with $\binit = \{ \sinit \mapsto 1 \}$ and $\abstbeliefstates = \set{\binit}$\\
	\DoUntil{$\uppervalueboundof{\binit} \leq \lambda$}{\label{line:refineloop:start}
		$\exploredstates \gets \set{\binit}$; $~Q \gets $ FIFO Queue initially containing $\binit$\\
		\While{$Q$ not empty}{
			take $\beliefstate$ from $Q$\\
			\If(\tcp*[f]{decide to explore $\beliefstate$ or not}){$\beliefstate \in \abstbeliefstates$ or \Explore{$\beliefstate$}}{\label{line:explorecheck}
				\ForEach{$\action \in \act{\beliefstate}$}{
					\If(\tcp*[f]{decide to rewire $\tuple{\beliefstate, \action}$ or not}){$\beliefstate \notin \abstbeliefstates$ or \Rewire{$\beliefstate$, $\action$}}{\label{line:rewirecheck}
						clear $\abstbelieftransitions(\beliefstate, \action, \discbeliefstate)$ for all $\discbeliefstate \in \abstbeliefstates$ \tcp*[r]{delete old transitions}
						\ForEach(\tcp*[f]{cf. \Cref{def:discbelief}}){$\discbeliefstate \in \post[\discbeliefmdp{\pomdp}{\foundation}]{\beliefstate}{\action}$}{
							$\abstbelieftransitions(\beliefstate, \action, \discbeliefstate) \gets \discbelieftransitions(\beliefstate, \action, \discbeliefstate)$\\
							\If{$\discbeliefstate \notin \exploredstates$}{
								insert $\discbeliefstate$ into $\abstbeliefstates$, $Q$, and $\exploredstates$
							}
						}
					} \Else(\tcp*[f]{keep the current successors}){
						insert all $\discbeliefstate \in \post[\abstbeliefmdp]{\beliefstate}{\action} \setminus \exploredstates$ into $Q$ and $\exploredstates$
					}
				}
			} \Else(\tcp*[f]{do not explore $\beliefstate$}) {
				\Cutoff{$\beliefstate$, $\abstbeliefmdp$} \tcp*[r]{redirect outgoing transitions to $\badbeliefs$}
			}
		}
		$\uppervalueboundof{\binit} \gets \sup_{\sched \in \scheds{\abstbeliefmdp}} \pr[\binit]{\abstbeliefmdp}{\sched}{\badbeliefs}$\label{line:modelcheck} \tcp*[r]{MDP model checking}
		$\foundation \gets $\Extend{$\foundation$} \tcp*[r]{consider refined neighbourhoods in next iteration}\label{line:extendf}
	}
\label{line:refineloop:end}
	}
	\caption{Abstraction-refinement loop.}
	\label{alg:refinementloop}
\end{algorithm}

\Cref{alg:refinementloop} sketches the abstraction-refinement loop. The algorithm iteratively constructs an abstraction MDP $\abstbeliefmdp$ via a breath-first-search on the state space of the discretised belief MDP $\discbeliefmdp{\pomdp}{\foundation}$ (Lines~\ref{line:refineloop:start} to~\ref{line:refineloop:end}).
In Line~\ref{line:explorecheck}, a heuristic $\Explore$ decides for each visited belief to either \emph{explore} or \emph{cut-off}. 
If we explore, we may encounter a state that was previously explored. Heuristic \Rewire decides in Line~\ref{line:rewirecheck} whether we \emph{rewire}, i.e., whether we explore the successors again (to account for potentially updated neighbourhoods) or whether we keep the existing successor states.
When cutting off, we use Eq.~\eqref{eq:overapprox} to obtain an upper bound $\uppervalueboundof{\beliefstate}$ for $\valueof{\beliefstate}$ and add a transition to some bad state with probability $\uppervalueboundof{\beliefstate}$ and a transition to a sink state with probability $1-\uppervalueboundof{\beliefstate}$.\footnote{The  implementation actually still connects $\beliefstate$ with already explored successors and only redirects the `missing' probabilities \wrt $\uppervalueboundof{\discbeliefstate}$, $\discbeliefstate \in \post[\discbeliefmdp{\pomdp}{\foundation}]{\state}{\action} \setminus \exploredstates$.}
The foundation is extended in Line~\ref{line:extendf}. This only has an effect in the next refinement step.

After building the MDP $\abstbeliefmdp$, it is analysed in  Line~\ref{line:modelcheck} using model checking.
This analysis yields a new upper bound  $\uppervalueboundof{\binit} \geq \valueof{\binit}$.
The loop can be stopped at any time, e.g., when threshold $\lambda$ is shown as upper bound.
Next, we describe how the foundation $\foundation$ is initialised, extended, and iteratively explored.


\subsubsection{Picking foundations}
\emph{The initial foundation.} 
\label{sec:algorithm:abstraction}
We discretise the beliefs using the foundation $\foundation$. The choice of this foundation is driven by the need to easily determine the neighbourhood and the vertex-weights. 
Furthermore, the cardinality of the neighbourhood affects the branching factor of the approximation MDP.
As~\cite{DBLP:journals/ior/Lovejoy91}, we use a triangulation scheme  based on \emph{Freudenthal Triangulation}~\cite{Fre42}, illustrated by \Cref{fig:regulartriangulization}.
Given fixed resolutions $\resolution[\observation] \in \nngz$, $\observation \in \observations$, the triangulation scheme yields discretised beliefs $\beliefstate$ with $\forall{s}\colon\, \beliefstate(\state) \in \{\nicefrac{i}{\resolution[\observation]} \mid \observation = \obsof{\beliefstate}, 0 \le i \le \resolution[\observation]\}$.

In the refinement loop shown in \Cref{alg:refinementloop}, we initialise $\foundation$ (Line~\ref{line:initf}) by setting the observation-dependent resolutions $\resolution[\observation]$ to a fixed value $\resolution[\mathit{init}]$.
Notice that it suffices to determine the neighbourhoods on-the-fly during the belief exploration. To compute the neighbourhood, we find $n_\observation + 1$ neighbours as intuitively depicted in \Cref{fig:regulartriangulization}. 
The intricate computation of these neighbours~\cite{Fre42} involves changing the basis for the vector space, ordering the coefficients and adequately manipulating single entries, before finally inverting the basis change, see \cite{DBLP:journals/ior/Lovejoy91} for an example.

\paragraph{Extension of foundation}
The set $\observations_{\mathrm{extend}}$ of observations for which the foundation will be extended is determined by assigning a $\score \colon \observations \rightarrow [0,1]$. Low scoring observations are  refined first. Intuitively, the score is assigned such that a score close to $0$ indicates that one of the approximated beliefs with observation $\observation$ is far away from all points in its neighbourhood, and a high score (close to 1) then means that all approximated beliefs are close to one of their neighbours. 
We set $\observations_{\mathrm{extend}} = \set{ \observation \in \observations \mid \mathit{score}(\observation) \le \rho_{\observations}}$ for some threshold $\rho_{\observations} \in [0,1]$. 
When the value of $\rho_{\observations}$ is iteratively increased towards $1$, each observation is eventually considered for refinement. 
Details are given in~\Cref{app:whichobs}.

\subsubsection{Iterative exploration}
The iterative exploration is guided using an estimate of how coarse the approximation is for the current belief state $\beliefstate$, and by an estimate of how likely we reach $\beliefstate$ under the optimal policy (which is unknown). If either of these values is small, then the influence of a potential cut-off at $\beliefstate$ is limited.

\paragraph{Bounds on reaching the bad state}
We use a lower bound $\lowervalueboundof{\beliefstate}$ and an upper bound $\uppervalueboundof{\beliefstate}$ for the value $\valueof{\beliefstate}$.
Eq.~\eqref{eq:overapprox} yields an easy-to-compute initial over-approximation $\uppervalueboundof{\beliefstate}$. 
Running the refinement-loop improves this bound.
For the lower bound, we exploit that 
any policy on the POMDP under-approximates the performance of the best policy.
Thus, we guess some set of observation-based policies\footnote{We guess policies in $\obsscheds{\pomdp}$ by distributing over actions of optimal policies for MDP~$\mdp$.} on the POMDP and evaluate them.
If these policies are memoryless, the induced Markov chain is in the size of the POMDP and is typically easy to evaluate.
Using a better under-approximation (e.g., by picking better policies, possibly exploiting the related work) is a promising direction for future research.

\paragraph{Estimating reachability likelihoods}
As a naive proxy for this likelihood, we consider almost optimal policies from the previous refinement step as well as the distance of $\beliefstate$ to the initial belief $\binit$.
Since the algorithm performs a breadth-first exploration, the distance from $\binit$ to $\beliefstate$ is reflected by the number of beliefs explored before $\beliefstate$.

\paragraph{State exploration}
In Line~\ref{line:explorecheck} of \Cref{alg:refinementloop}, $\Explore$ decides whether the successors of the current belief $\beliefstate$ are explored or  cut off.
We only explore the successors of $\beliefstate$ if:  (1)
 \emph{the approximation is coarse}, i.e., if the relative gap between $\uppervalueboundof{\beliefstate}$ and $\lowervalueboundof{\beliefstate}$  is above (a decreasing) $\rho_\mathit{gap}$\footnote{$\rho_\mathit{gap}$ is set to $0.1$ initially and after each iteration we update it to  $\rho_\mathit{gap}/4$.}.
(2) \emph{the state is likely relevant for the optimal scheduler}, i.e.,
if (i) at most $\rho_\mathit{step}$
\footnote{$\rho_\mathit{step}$ is set to $\infty$ initially and after each iteration we update it to $4 \cdot |\abstbeliefstates|$.}
 beliefs were explored (or rewired) before and (ii) $\beliefstate$ is reachable under a $\rho_{\scheds{}}$-optimal policy\footnote{A policy $\sched$ is $\rho_{\scheds{}}$-optimal if $\forall \beliefstate \colon \valueof[\sched(\beliefstate)]{\beliefstate} + \rho_{\scheds{}} \geq \valueof{\beliefstate}$. We set $\rho_{\scheds{}} = 0.001$.} from the previous refinement step.

\paragraph{Rewiring}
We apply the same criteria for $\Rewire$ in Line~\ref{line:rewirecheck}.
In addition, we only rewire the successors for action $\action$ if (i) $\action$ is selected by some $\rho_{\scheds{}}$-optimal policy and (ii) the rewiring actually has an effect, \ie, for at least one successor the foundation has been extended since the last exploration of $\beliefstate$ and $\action$.

\section{Experiments}
\label{sec:experiments}

\paragraph{Implementation}
We integrated the abstraction-refinement framework in the model checker \tool{Storm}~\cite{DBLP:journals/corr/abs-2002-07080}.
The implementation constructs the abstraction MDP as detailed in \Cref{alg:refinementloop} using sparse matrices. 
The computation in Line~\ref{line:modelcheck} is performed using \tool{Storm}'s implementation of \emph{optimistic value iteration}~\cite{DBLP:journals/corr/abs-1910-01100}, yielding sound precision guarantees up to relative precision $\varepsilon = 10^{-6}$.
Our implementation supports arbitrary combinations of minimisation and maximisation of reachability and reach-avoid specifications, and indefinite-horizon expected rewards. 
For minimisation, lower and upper bounds are swapped.

Additionally, our implementation may compute lower bounds by iteratively exploring (a fragment of) the belief MDP, \emph{without} the discretisation. 
The state-space exploration is cut off after exploring an increasing number of states\footnote{In refinement step $i$, we explore $2^{i-1} \cdot |\states| \cdot \max_{\observation \in \observations}| \obsfunction^{-1}(\observation)|$ states.}.

\paragraph{Models}
We use \emph{all} sets of POMDPs from~\cite{DBLP:journals/rts/Norman0Z17}. 
Small versions of these benchmarks are omitted.
We additionally introduced some variants, e.g., added uncertainty to the movement in the grid examples. 
Finally, we consider three scalable variants of typical grid-world planning domains in artificial intelligence\footnote{These examples vary over the literature, we attach details in Appendix~\ref{sec:benchmarks}.}. 

\paragraph{Set-up}
We evaluate our implementation with and without the refinement loop.
In the former case, the refinement loop runs a given amount of time and we report the results obtained so far.
In the latter case, a single iteration of~\Cref{alg:refinementloop} is performed with a fixed triangulation resolution $\resolution$---a set-up as in~\cite{DBLP:journals/rts/Norman0Z17}.
We compare with the implementatation~\cite{DBLP:journals/rts/Norman0Z17} in \tool{Prism}.
We used a simple SCC analysis to find POMDPs where the reachable belief MDP is finite. 
All POMDPs from~\cite{DBLP:journals/rts/Norman0Z17} are in this category.
We refer to the remaining POMDPs as infinite belief POMDPs.

All experiments were run on 4 cores\footnote{\tool{Storm} uses one core, \tool{Prism} uses four cores in garbage collection only.} of an Intel\textsuperscript{\textregistered} Xeon\textsuperscript{\textregistered} Platinum 8160 CPU with a time limit of 1 hour (unless indicated otherwise) and 32 GB RAM.  

\begin{table}[t]
\caption{Results for POMDPs with infinite belief MDP.}
\label{tab:inf}
\scriptsize{
\adjustbox{{max width=\textwidth}}{%
\begin{tabular}{|cc||rr||r|r||r|r|r|r|r||r|r|}
\hline
\multicolumn{2}{|c||}{Benchmark} & \multicolumn{2}{c||}{Data} & \multicolumn{1}{c|}{MDP} & \multicolumn{1}{c||}{$\beliefmdp{\pomdp}$} & \multicolumn{2}{c|}{$\resolution$=4} & \multicolumn{3}{c||}{$\resolution$=12} & \multicolumn{2}{c|}{refine}\\
\multicolumn{1}{|c}{Model} & \multicolumn{1}{c||}{$\phi$} & \multicolumn{1}{c}{$\states$/$\actions$} & \multicolumn{1}{c||}{$\observations$} & \multicolumn{2}{c||}{\tool{Storm}} & \multicolumn{1}{c|}{\tool{Prism}} & \multicolumn{1}{c|}{\tool{Storm}} & \multicolumn{1}{c|}{\tool{Prism}} & \multicolumn{4}{c|}{\tool{Storm}}\\
\multicolumn{1}{|c}{} & \multicolumn{1}{c||}{} & \multicolumn{1}{c}{} & \multicolumn{1}{c||}{} & \multicolumn{1}{c|}{} & \multicolumn{1}{c||}{t=60} & \multicolumn{1}{c|}{} & \multicolumn{1}{c|}{$\rho_\mathit{gap}$=0} & \multicolumn{1}{c|}{} & \multicolumn{1}{c|}{$\rho_\mathit{gap}$=0} & \multicolumn{1}{c||}{$\rho_\mathit{gap}$=0.2} & \multicolumn{1}{c|}{t=60} & \multicolumn{1}{c|}{t=1800}\\
\hline\hline
\model{Drone} & \multirow{2}{*}{$P_\mathrm{max}$} & $1226$ & \multirow{2}{*}{$384$} & \multirow{2}{*}{0.98} & ${\ge}\, 0.84$ & \multirow{2}{*}{TO} & ${\le}\, \mathbf{0.96}$ & \multirow{2}{*}{MO} & \multirow{2}{*}{MO} & \multirow{2}{*}{MO} & ${\le}\, 0.97$ & ${\le}\, 0.97$$^\dagger$\\
4-1 &  & $3026$ &  &  & 6 &  & \ensuremath{6.67} &  &  &  & 2 & 3\\
\hline
\model{Drone} & \multirow{2}{*}{$P_\mathrm{max}$} & $1226$ & \multirow{2}{*}{$761$} & \multirow{2}{*}{0.98} & ${\ge}\, 0.96$ & \multirow{2}{*}{TO} & ${\le}\, 0.98$ & \multirow{2}{*}{MO} & ${\le}\, \mathbf{0.97}$ & ${\le}\, \mathbf{0.97}$ & ${\le}\, \mathbf{0.97}$ & ${\le}\, \mathbf{0.97}$$^\dagger$\\
4-2 &  & $3026$ &  &  & 7 &  & \ensuremath{${<}\,$ 1} &  & \ensuremath{194} & \ensuremath{173} & 3 & 4\\
\hline\hline
\model{Grid-av} & \multirow{2}{*}{$P_\mathrm{max}$} & $17$ & \multirow{2}{*}{$4$} & \multirow{2}{*}{1} & ${\ge}\, 0.93$ & $[0.21, 1.0]$ & ${\le}\, 1$ & \multirow{2}{*}{MO} & ${\le}\, \mathbf{0.94}$ & ${\le}\, \mathbf{0.94}$ & ${\le}\, \mathbf{0.94}$ & ${\le}\, \mathbf{0.94}$$^\dagger$\\
4-0.1 &  & $59$ &  &  & 13 & \ensuremath{2.03} & \ensuremath{${<}\,$ 1} &  & \ensuremath{164} & \ensuremath{168} & 3 & 3\\
\hline\hline
\model{Grid} & \multirow{2}{*}{$R_\mathrm{min}$} & $17$ & \multirow{2}{*}{$3$} & \multirow{2}{*}{3.56} & ${\le}\, 4.7$ & $[4.06, 4.7]$ & ${\ge}\, 4.06$ & \multirow{2}{*}{MO} & ${\ge}\, 4.59$ & ${\ge}\, 4.59$ & ${\ge}\, 4.56$ & ${\ge}\, \mathbf{4.61}$$^\dagger$\\
4-0.1 &  & $62$ &  &  & 13 & \ensuremath{2.02} & \ensuremath{${<}\,$ 1} &  & \ensuremath{264} & \ensuremath{268} & 3 & 4\\
\hline
\model{Grid} & \multirow{2}{*}{$R_\mathrm{min}$} & $17$ & \multirow{2}{*}{$3$} & \multirow{2}{*}{4.57} & ${\le}\, 6.37$ & $[5.4, 6.31]$ & ${\ge}\, 5.4$ & \multirow{2}{*}{MO} & ${\ge}\, \mathbf{6.18}$ & ${\ge}\, \mathbf{6.18}$ & ${\ge}\, 5.92$ & ${\ge}\, 5.92$$^\dagger$\\
4-0.3 &  & $62$ &  &  & 13 & \ensuremath{3.05} & \ensuremath{${<}\,$ 1} &  & \ensuremath{217} & \ensuremath{214} & 3 & 4\\
\hline\hline
\model{Maze2} & \multirow{2}{*}{$R_\mathrm{min}$} & $15$ & \multirow{2}{*}{$8$} & \multirow{2}{*}{5.64} & ${\le}\, 6.32$ & $[6.29, 6.32]$ & ${\ge}\, 6.29$ & $[\mathbf{6.32}, 6.32]$ & ${\ge}\, \mathbf{6.32}$ & ${\ge}\, \mathbf{6.32}$ & ${\ge}\, \mathbf{6.32}$ & ${\ge}\, \mathbf{6.32}$$^\dagger$\\
0.1 &  & $54$ &  &  & 14 & \ensuremath{1.35} & \ensuremath{${<}\,$ 1} & \ensuremath{4.91} & \ensuremath{${<}\,$ 1} & \ensuremath{${<}\,$ 1} & 7 & 8\\
\hline\hline
\model{Refuel} & \multirow{2}{*}{$P_\mathrm{max}$} & $208$ & \multirow{2}{*}{$50$} & \multirow{2}{*}{0.98} & ${\ge}\, 0.67$ & \multirow{2}{*}{TO} & ${\le}\, 0.71$ & \multirow{2}{*}{MO} & ${\le}\, 0.68$ & ${\le}\, 0.68$ & ${=}\, \mathbf{0.67}$* & ${=}\, \mathbf{0.67}$*\\
06 &  & $574$ &  &  & 10 &  & \ensuremath{${<}\,$ 1} &  & \ensuremath{2.08} & \ensuremath{2.08} & 59 & 59\\
\hline
\model{Refuel} & \multirow{2}{*}{$P_\mathrm{max}$} & $470$ & \multirow{2}{*}{$66$} & \multirow{2}{*}{0.99} & ${\ge}\, 0.45$ & \multirow{2}{*}{MO} & ${\le}\, 0.76$ & \multirow{2}{*}{MO} & \multirow{2}{*}{MO} & \multirow{2}{*}{MO} & ${\le}\, 0.75$ & ${\le}\, \mathbf{0.58}$$^\dagger$\\
08 &  & $1446$ &  &  & 7 &  & \ensuremath{7.3} &  &  &  & 2 & 3\\
\hline\hline
\model{Rocks} & \multirow{2}{*}{$R_\mathrm{min}$} & $6553$ & \multirow{2}{*}{$1645$} & \multirow{2}{*}{16.5} & ${\le}\, 35.4$ & \multirow{2}{*}{TO} & ${\ge}\, 19.9$ & \multirow{2}{*}{MO} & ${\ge}\, \mathbf{20}$ & ${\ge}\, \mathbf{20}$ & ${=}\, \mathbf{20}$* & ${=}\, \mathbf{20}$*\\
12 &  & $3{\cdot} 10^{4}$ &  &  & 6 &  & \ensuremath{1.26} &  & \ensuremath{18.9} & \ensuremath{19.1} & 9 & 9\\
\hline
\model{Rocks} & \multirow{2}{*}{$R_\mathrm{min}$} & $1{\cdot} 10^{4}$ & \multirow{2}{*}{$2761$} & \multirow{2}{*}{22} & ${\le}\, 44$ & \multirow{2}{*}{MO} & ${\ge}\, 25.6$ & \multirow{2}{*}{MO} & ${\ge}\, \mathbf{26}$ & ${\ge}\, \mathbf{26}$ & ${\ge}\, 25.9$ & ${\ge}\, 25.9$\\
16 &  & $5{\cdot} 10^{4}$ &  &  & 5 &  & \ensuremath{2.55} &  & \ensuremath{37.2} & \ensuremath{35.9} & 8 & 9\\
\hline
\end{tabular}

}
}
\end{table}
\begin{table}[t]
	\caption{Results for POMDPs with finite belief MDP.}
	\label{tab:fin}
	\scriptsize{
	\adjustbox{{max width=\textwidth}}{%
		\begin{tabular}{|cc||rr||r|r||r|r|r|r|r||r|r|}
\hline
\multicolumn{2}{|c||}{Benchmark} & \multicolumn{2}{c||}{Data} & \multicolumn{1}{c|}{MDP} & \multicolumn{1}{c||}{$\beliefmdp{\pomdp}$} & \multicolumn{2}{c|}{$\resolution$=4} & \multicolumn{3}{c||}{$\resolution$=12} & \multicolumn{2}{c|}{refine}\\
\multicolumn{1}{|c}{Model} & \multicolumn{1}{c||}{$\phi$} & \multicolumn{1}{c}{$\states$/$\actions$} & \multicolumn{1}{c||}{$\observations$} & \multicolumn{2}{c||}{\tool{Storm}} & \multicolumn{1}{c|}{\tool{Prism}} & \multicolumn{1}{c|}{\tool{Storm}} & \multicolumn{1}{c|}{\tool{Prism}} & \multicolumn{4}{c|}{\tool{Storm}}\\
\multicolumn{1}{|c}{} & \multicolumn{1}{c||}{} & \multicolumn{1}{c}{} & \multicolumn{1}{c||}{} & \multicolumn{1}{c|}{} & \multicolumn{1}{c||}{} & \multicolumn{1}{c|}{} & \multicolumn{1}{c|}{$\rho_\mathit{gap}$=0} & \multicolumn{1}{c|}{} & \multicolumn{1}{c|}{$\rho_\mathit{gap}$=0} & \multicolumn{1}{c||}{$\rho_\mathit{gap}$=0.2} & \multicolumn{1}{c|}{t=60} & \multicolumn{1}{c|}{t=1800}\\
\hline\hline
\model{Crypt} & \multirow{2}{*}{$P_\mathrm{max}$} & $1972$ & \multirow{2}{*}{$510$} & \multirow{2}{*}{1} & ${=}\, \mathbf{0.33}$ & $[0.33, 0.79]$ & ${\le}\, 0.79$ & \multirow{2}{*}{MO} & ${\le}\, \mathbf{0.33}$ & ${\le}\, \mathbf{0.33}$ & ${=}\, \mathbf{0.33}$* & ${=}\, \mathbf{0.33}$*\\
4 &  & $4612$ &  &  & \ensuremath{3.51} & \ensuremath{20.3} & \ensuremath{${<}\,$ 1} &  & \ensuremath{1.36} & \ensuremath{6.12} & 6 & 6\\
\hline
\model{Crypt} & \multirow{2}{*}{$P_\mathrm{max}$} & $7{\cdot} 10^{4}$ & \multirow{2}{*}{$6678$} & \multirow{2}{*}{1} & ${=}\, \mathbf{0.2}$ & \multirow{2}{*}{MO} & ${\le}\, 1$ & \multirow{2}{*}{MO} & ${\le}\, 0.84$ & ${\le}\, 0.84$ & ${\le}\, 0.97$ & ${\le}\, 0.94$\\
6 &  & $2{\cdot} 10^{5}$ &  &  & \ensuremath{8.47} &  & \ensuremath{17.8} &  & \ensuremath{155} & \ensuremath{159} & 2 & 4\\
\hline\hline
\model{Grid-av} & \multirow{2}{*}{$P_\mathrm{max}$} & $17$ & \multirow{2}{*}{$4$} & \multirow{2}{*}{1} & ${=}\, \mathbf{0.93}$ & $[0.21, 1.0]$ & ${\le}\, 1$ & \multirow{2}{*}{MO} & ${\le}\, 0.94$ & ${\le}\, 0.94$ & ${\le}\, \mathbf{0.93}$ & ${\le}\, \mathbf{0.93}$$^\dagger$\\
4-0 &  & $59$ &  &  & \ensuremath{${<}\,$ 1} & \ensuremath{1.51} & \ensuremath{${<}\,$ 1} &  & \ensuremath{${<}\,$ 1} & \ensuremath{${<}\,$ 1} & 9 & 26\\
\hline\hline
\model{Maze2} & \multirow{2}{*}{$R_\mathrm{min}$} & $15$ & \multirow{2}{*}{$8$} & \multirow{2}{*}{5.08} & ${=}\, \mathbf{5.69}$ & $[\mathbf{5.69}, 5.69]$ & ${\ge}\, \mathbf{5.69}$ & $[\mathbf{5.69}, 5.69]$ & ${\ge}\, \mathbf{5.69}$ & ${\ge}\, \mathbf{5.69}$ & ${=}\, \mathbf{5.69}$* & ${=}\, \mathbf{5.69}$*\\
0 &  & $54$ &  &  & \ensuremath{${<}\,$ 1} & \ensuremath{1.43} & \ensuremath{${<}\,$ 1} & \ensuremath{3.17} & \ensuremath{${<}\,$ 1} & \ensuremath{${<}\,$ 1} & 4 & 4\\
\hline\hline
\model{Netw-p} & \multirow{2}{*}{$R_\mathrm{max}$} & $2{\cdot} 10^{4}$ & \multirow{2}{*}{$4909$} & \multirow{2}{*}{566} & ${=}\, \mathbf{557}$ & $[557, 559]$ & ${\le}\, 560$ & \multirow{2}{*}{TO} & ${\le}\, \mathbf{557}$ & ${\le}\, 566$ & ${\le}\, \mathbf{557}$ & ${\le}\, \mathbf{557}$\\
2-8-20 &  & $3{\cdot} 10^{4}$ &  &  & \ensuremath{612} & \ensuremath{503} & \ensuremath{2.17} &  & \ensuremath{4.25} & \ensuremath{${<}\,$ 1} & 10 & 18\\
\hline
\model{Netw-p} & \multirow{2}{*}{$R_\mathrm{max}$} & $2{\cdot} 10^{5}$ & \multirow{2}{*}{$2{\cdot} 10^{4}$} & \multirow{2}{*}{849} & \multirow{2}{*}{TO} & \multirow{2}{*}{TO} & ${\le}\, 832$ & \multirow{2}{*}{MO} & \multirow{2}{*}{TO} & ${\le}\, 849$ & ${\le}\, 849$ & ${\le}\, \mathbf{825}$\\
3-8-20 &  & $3{\cdot} 10^{5}$ &  &  &  &  & \ensuremath{514} &  &  & \ensuremath{8.2} & 0 & 2\\
\hline\hline
\model{Netw} & \multirow{2}{*}{$R_\mathrm{min}$} & $4589$ & \multirow{2}{*}{$1173$} & \multirow{2}{*}{2.56} & ${=}\, \mathbf{3.2}$ & $[3.03, 3.2]$ & ${\ge}\, 2.97$ & $[3.17, 3.2]$ & ${\ge}\, 3.17$ & ${\ge}\, 3.16$ & ${\ge}\, \mathbf{3.2}$ & ${\ge}\, \mathbf{3.2}$\\
2-8-20 &  & $6973$ &  &  & \ensuremath{38.4} & \ensuremath{42.1} & \ensuremath{${<}\,$ 1} & \ensuremath{521} & \ensuremath{${<}\,$ 1} & \ensuremath{${<}\,$ 1} & 10 & 23\\
\hline
\model{Netw} & \multirow{2}{*}{$R_\mathrm{min}$} & $2{\cdot} 10^{4}$ & \multirow{2}{*}{$2205$} & \multirow{2}{*}{3.88} & \multirow{2}{*}{MO} & $[5.54, 6.77]$ & ${\ge}\, 5.11$ & \multirow{2}{*}{MO} & ${\ge}\, 6.35$ & ${\ge}\, 6.33$ & ${\ge}\, 6.26$ & ${\ge}\, \mathbf{6.72}$$^\dagger$\\
3-8-20 &  & $3{\cdot} 10^{4}$ &  &  &  & \ensuremath{1777} & \ensuremath{4.82} &  & \ensuremath{34.5} & \ensuremath{34.3} & 3 & 5\\
\hline\hline
\model{Nrp} & \multirow{2}{*}{$P_\mathrm{max}$} & $125$ & \multirow{2}{*}{$41$} & \multirow{2}{*}{1} & ${=}\, \mathbf{0.12}$ & $[0.13, 0.38]$ & ${\le}\, 0.38$ & $[0.13, 0.22]$ & ${\le}\, 0.22$ & ${\le}\, 0.22$ & ${=}\, \mathbf{0.12}$* & ${=}\, \mathbf{0.12}$*\\
8 &  & $161$ &  &  & \ensuremath{${<}\,$ 1} & \ensuremath{1.57} & \ensuremath{${<}\,$ 1} & \ensuremath{22.9} & \ensuremath{${<}\,$ 1} & \ensuremath{${<}\,$ 1} & 70 & 70\\
\hline
\end{tabular}

	}
	}
\end{table}

\paragraph{Results}
We consider the infinite belief POMDPs in \Cref{tab:inf}.
The first columns indicate the POMDP model instance, the type of the checked property (probabilities ($P$) or rewards ($R$), minimising or maximising policies), as well as the number of states, state-action pairs, and observations of the POMDP.
The column `MDP' shows the model checking result on the underlying, fully-observable MDP.
The column `$\beliefmdp{\pomdp}$'  considers the refinement loop for the non-discretised belief MDP as discussed above and lists the best result obtained within 60 seconds, and the number of iterations.
The subsequent columns show our result for a single approximation step with fixed resolution $\resolution$ and cut-off threshold $\rho_\mathit{gap}$, as well as the results of \tool{Prism} when invoked with resolution $\resolution$. 
`TO' and `MO' indicate a time-out ($>$ 1 hour) and a memory-out ($>$ 32 GB), respectively.
Each cell contains the obtained bounds on the result and the analysis time in seconds.
Finally, the last two columns report on running the refinement loop for at most $t$ (60 and 1800) seconds.
The cells contain the best bound on the result and the number of loop iterations of \Cref{alg:refinementloop}.
In addition, $*$ indicates that no further refinement was possible (in this case the model-checking result corresponds to the precise value) and $\dagger$ indicates that an MO occurred before $t$ seconds.

\Cref{tab:fin} provides the experimental results for benchmark models with finite belief MDP. The columns are similar as in \Cref{tab:inf} except that column `$\beliefmdp{\pomdp}$` indicates the model checking result and analysis time in seconds for the complete finite belief MDP. Appendix~\ref{app:experiments} contains further experiments.

\paragraph{Discussion}
We start with some observations and focus on Table~\ref{tab:inf}.
First, our implementation outperforms the implementation of~\cite{DBLP:journals/rts/Norman0Z17} by several orders of magnitude,  most likely due to the on-the-fly state-space construction, and by an engineering effort. 
This difference cannot be explained by the currently implemented cut-offs; indeed, when choosing a static foundation, cut-offs do not improve performance noticeably.
Second, our refinement loop avoids the need for a user-picked resolution, but a hand-picked resolution is sometimes faster (e.g.\ for \model{Maze}) or yields better results (e.g.\ for \model{Grid}). 
On the other hand, the refinement loop might find finite abstractions that concisely represent the belief MDP reachable under the optimal policy (e.g.\ for \model{Rocks}). 
Here, cut-offs are essential.
Third, on many benchmarks, the refinement loop finds the crucial part of the abstraction within a minute, but e.g., \model{Refuel} profits from additional time. 

We want to share three further observations: First, it seems interesting to investigate finite-belief POMDPs as these occur quite frequently (see Table~\ref{tab:fin}) and can be analysed straightforwardly. 
Second, the current bottleneck is the bookkeeping of the belief states and the computation of neighbourhoods, not the model checking. 
Finally, even more than for MDPs, the size of the POMDP (or the number of observations) is not at all a proxy for the difficulty of verification.

\paragraph{Data Availability}
The implementation, models, and log files are available at
\cite{experiments}.

\section{Conclusion and Future Work}
We presented an abstraction-refinement for solving the verification problem for indefinite-horizon properties in POMDPs, e.g., for proving that all policies reach a bad state with at most probability $\lambda$.
As the original problem is undecidable, we compute a sequence of over-approximations by iteratively refining an abstraction of the belief MDP.
Our prototype shows superior performance over~\cite{DBLP:journals/rts/Norman0Z17} in \tool{Prism}.
The next step is to integrate better under-approximations.

\bibliography{main}

\begin{thebibliography}{10}
\providecommand{\url}[1]{\texttt{#1}}
\providecommand{\urlprefix}{URL }

\bibitem{DBLP:journals/aamas/AmatoBZ10}
Amato, C., Bernstein, D.S., Zilberstein, S.: Optimizing fixed-size stochastic
  controllers for {POMDP}s and decentralized {POMDP}s. Autonomous Agents and
  Multi-Agent Systems  21(3),  293--320 (2010)

\bibitem{DBLP:conf/atva/AshokBHK18}
Ashok, P., Butkova, Y., Hermanns, H., Kret{\'{\i}}nsk{\'{y}}, J.:
  Continuous-time {Markov} decisions based on partial exploration.    {ATVA}.
  {LNCS}  11138, pp. 317--334. Springer (2018)

\bibitem{BK08}
Baier, C., Katoen, J.P.: Principles of Model Checking. The {MIT} Press (2008)

\bibitem{DBLP:conf/ijcai/BonetG09}
Bonet, B., Geffner, H.: Solving {POMDP}s: {RTDP}-{B}el vs. point-based
  algorithms.    {IJCAI}. pp. 1641--1646 (2009)

\bibitem{experiments}
Bork, A., Junges, S., Katoen, J.P., Quatmann, T.: {Experiments for
  'Verification of indefinite- horizon POMDPs'},
  \url{https://doi.org/10.5281/zenodo.3924577}

\bibitem{DBLP:journals/corr/abs-2001-03809}
Bouton, M., Tumova, J., Kochenderfer, M.J.: Point-based methods for model
  checking in partially observable {Markov} decision processes. CoRR
  abs/2001.03809 (2020)

\bibitem{DBLP:conf/atva/BrazdilCCFKKPU14}
Br{\'{a}}zdil, T., Chatterjee, K., Chmelik, M., Forejt, V.,
  Kret{\'{\i}}nsk{\'{y}}, J., Kwiatkowska, M.Z., Parker, D., Ujma, M.:
  Verification of {Markov} decision processes using learning algorithms.
  {ATVA}. {LNCS}  8837, pp. 98--114. Springer (2014)

\bibitem{DBLP:conf/aaai/BraziunasB04}
Braziunas, D., Boutilier, C.: Stochastic local search for {{POMDP}}
  controllers.    {AAAI}. pp. 690--696. {AAAI} Press / The {MIT} Press (2004)

\bibitem{DBLP:conf/cav/CernyCHRS11}
Cern{\'{y}}, P., Chatterjee, K., Henzinger, T., Radhakrishna, A., Singh, R.:
  Quantitative synthesis for concurrent programs.    {CAV}. {LNCS}  6806, pp.
  243--259. Springer (2011)

\bibitem{DBLP:conf/icra/ChatterjeeCGK15}
Chatterjee, K., Chmelik, M., Gupta, R., Kanodia, A.: Qualitative analysis of
  {POMDP}s with temporal logic specifications for robotics applications.
  {ICRA}. pp. 325--330. {IEEE} (2015)

\bibitem{Fre42}
Freudenthal, H.: Simplizialzerlegungen von beschrankter {F}lachheit. Annals of
  Mathematics  43(3),  580--582 (1942)

\bibitem{DBLP:conf/uai/Hansen98}
Hansen, E.A.: Solving {POMDP}s by searching in policy space.    {UAI}. pp.
  211--219. Morgan Kaufmann (1998)

\bibitem{DBLP:conf/tacas/HartmannsH14}
Hartmanns, A., Hermanns, H.: The {M}odest toolset: An integrated environment
  for quantitative modelling and verification.    {TACAS}. {LNCS}  8413, pp.
  593--598. Springer (2014)

\bibitem{DBLP:journals/corr/abs-1910-01100}
Hartmanns, A., Kaminski, B.L.: Optimistic value iteration. CoRR  abs/1910.01100
  (2019)

\bibitem{DBLP:journals/corr/abs-2002-07080}
Hensel, C., Junges, S., Katoen, J.P., Quatmann, T., Volk, M.: The probabilistic
  model checker {S}torm. CoRR  abs/2002.07080 (2020)

\bibitem{DBLP:conf/ijcai/HorakBC18}
Hor{\'{a}}k, K., Bosansk{\'{y}}, B., Chatterjee, K.: Goal-{HSVI}: Heuristic
  search value iteration for goal {POMDP}s.    {IJCAI}. pp. 4764--4770.
  ijcai.org (2018)

\bibitem{DBLP:conf/atva/0001DKKW16}
Jansen, N., Dehnert, C., Kaminski, B.L., Katoen, J.P., Westhofen, L.: Bounded
  model checking for probabilistic programs.    {ATVA}. {LNCS}  9938, pp.
  68--85 (2016)

\bibitem{DBLP:conf/uai/Junges0WQWK018}
Junges, S., Jansen, N., Wimmer, R., Quatmann, T., Winterer, L., Katoen, J.P.,
  Becker, B.: Finite-state controllers of {POMDP}s using parameter synthesis.
   {UAI}. pp. 519--529. {AUAI} Press (2018)

\bibitem{DBLP:journals/ai/KaelblingLC98}
Kaelbling, L.P., Littman, M.L., Cassandra, A.R.: Planning and acting in
  partially observable stochastic domains. Artif. Intell.  101(1-2),  99--134
  (1998)

\bibitem{Koc2015}
Kochenderfer, M.J.: Decision Making Under Uncertainty. The MIT Press (2015)

\bibitem{DBLP:conf/rss/KurniawatiHL08}
Kurniawati, H., Hsu, D., Lee, W.S.: {SARSOP:} efficient point-based {{POMDP}}
  planning by approximating optimally reachable belief spaces.    Robotics:
  Science and Systems. The {MIT} Press (2008)

\bibitem{DBLP:conf/cav/KwiatkowskaNP11}
Kwiatkowska, M.Z., Norman, G., Parker, D.: {PRISM} 4.0: Verification of
  probabilistic real-time systems.    {CAV}. {LNCS}  6806, pp. 585--591.
  Springer (2011)

\bibitem{DBLP:journals/ior/Lovejoy91}
Lovejoy, W.S.: Computationally feasible bounds for partially observed {Markov}
  decision processes. Oper. Res.  39(1),  162--175 (1991)

\bibitem{DBLP:journals/ai/MadaniHC03}
Madani, O., Hanks, S., Condon, A.: On the undecidability of probabilistic
  planning and related stochastic optimization problems. Artif. Intell.
  147(1-2),  5--34 (2003)

\bibitem{DBLP:conf/uai/MeuleauKKC99}
Meuleau, N., Kim, K., Kaelbling, L.P., Cassandra, A.R.: Solving {POMDP}s by
  searching the space of finite policies.    {UAI}. pp. 417--426. Morgan
  Kaufmann (1999)

\bibitem{DBLP:journals/rts/Norman0Z17}
Norman, G., Parker, D., Zou, X.: Verification and control of partially
  observable probabilistic systems. Real-Time Systems  53(3),  354--402 (2017)

\bibitem{DBLP:conf/nips/PajarinenP11}
Pajarinen, J., Peltonen, J.: Periodic finite state controllers for efficient
  {{POMDP}} and {DEC-{POMDP}} planning.    {NIPS}. pp. 2636--2644 (2011)

\bibitem{DBLP:conf/ijcai/PineauGT03}
Pineau, J., Gordon, G.J., Thrun, S.: Point-based value iteration: An anytime
  algorithm for {POMDP}s.    {IJCAI}. pp. 1025--1032. Morgan Kaufmann (2003)

\bibitem{DBLP:books/daglib/0023820}
Russell, S.J., Norvig, P.: Artificial Intelligence -- {A} Modern Approach.
  Pearson Education (2010)

\bibitem{DBLP:journals/aamas/ShaniPK13}
Shani, G., Pineau, J., Kaplow, R.: A survey of point-based {{POMDP}} solvers.
  Auton. Agents Multi Agent Syst.  27(1),  1--51 (2013)

\bibitem{thrun2005probabilistic}
Thrun, S., Burgard, W., Fox, D.: Probabilistic Robotics. The MIT Press (2005)

\bibitem{DBLP:journals/tii/VolkJK18}
Volk, M., Junges, S., Katoen, J.P.: Fast dynamic fault tree analysis by model
  checking techniques. {IEEE} Trans. Industrial Informatics  14(1),  370--379
  (2018)

\bibitem{DBLP:journals/jair/WalravenS19}
Walraven, E., Spaan, M.T.J.: Point-based value iteration for finite-horizon
  {POMDP}s. J. Artif. Intell. Res.  65,  307--341 (2019)

\bibitem{DBLP:conf/cdc/WintererJW0TK017}
Winterer, L., Junges, S., Wimmer, R., Jansen, N., Topcu, U., Katoen, J.P.,
  Becker, B.: Motion planning under partial observability using game-based
  abstraction.    {CDC}. pp. 2201--2208. {IEEE} (2017)

\bibitem{DBLP:conf/cdc/WongpiromsarnF12}
Wongpiromsarn, T., Frazzoli, E.: Control of probabilistic systems under
  dynamic, partially known environments with temporal logic specifications.
  {CDC}. pp. 7644--7651. {IEEE} (2012)

\end{thebibliography}
\bibliographystyle{splncs03}

\clearpage
\appendix
%
%
\section{Details for Selecting and Extending the Foundation}
\label{app:whichobs}

\paragraph{Initializing and refining the foundation}
As mentioned in \Cref{sec:algorithm}, we initialize the foundation $\foundation$ in Line~\ref{line:initf} of \Cref{alg:refinementloop} by applying  Freudenthal Triangulation~\cite{Fre42}, with a fixed resolution $\resolution[\mathit{init}] > 0$ \ie
\[
\foundation = \Big\{~\beliefstate \in \beliefstates ~\mid~  \forall{\state \in \states}\colon\, \beliefstate(\state) \in \{\nicefrac{i}{\resolution[\observation]} \mid \observation = \obsof{\beliefstate}, \text{with }  i \in \mathbb{N}, 0 \le i \le \resolution[\observation]\}~ \Big\}.
\]
Where initially $\resolution[\observation] = \resolution[\mathit{init}]$ for all $\observation \in \observations$.
To extend $\foundation$ (Line~\ref{line:extendf} of \Cref{alg:refinementloop}), we heuristically pick a set of observations $\observations_{\mathrm{extend}}$ (details below) and increase the resolutions $\resolution[\observation]$ for $\observation \in \observations_{\mathrm{extend}}$ by a factor $f_{\resolution} > 1$.
By default, our implementation assumes $\resolution[\mathit{init}]=3$ and $f_{\resolution} = 2$.

We also implemented a more \emph{dynamic} triangulation scheme that attempts to minimize the cardinality of the neighbourhoods (and thus the branching of the approximation MDP).
For belief state $\beliefstate$ let $\neighbourhood[\resolution]{\beliefstate}$ be the neighbourhood obtained with Freudenthal triangulation when using resolution $\resolution$.
For our dynamic triangulation approach we triangulate belief $\beliefstate$ with observation $\observation = \obsof{\beliefstate}$ using the neighbourhood $\neighbourhood[\resolution]{\beliefstate}$, where $\resolution$ is the largest resolution satisfying
\[
\resolution \le \resolution[\observation] \quad \text{and} \quad
|\neighbourhood[\resolution]{\beliefstate}| = \min_{\resolution' \le \resolution[\observation]} |\neighbourhood[\resolution']{\beliefstate}|.
\]
An experimental evaluation of this dynamic approach is given in \Cref{app:experiments}.

\paragraph{Selecting observations to refine}
To determine the set $\observations_{\mathrm{extend}}$ of observations that will be refined, we assign the following score to each observation $\observation$.

For belief state $\beliefstate$ let $\neighbourhood[\foundation]{\beliefstate}$ be the triangulation neighbourhood with respect to the current foundation $\foundation$.
Further, let $\vertexweight{\beliefstate} \in \dists{\neighbourhood[\foundation]{\beliefstate}}$ be the vertex distribution as in \Cref{def:convexneighbourhood} and let $n =  |\supp{\beliefstate}|$.

We use the following score to evaluate how good $\beliefstate$ is approximated by its neighbourhood.
If $n = 1$, $\beliefstate$ is a Dirac belief and gets a score of $1$ (the best possible score). Otherwise,
\begin{displaymath}
\mathit{score}(\beliefstate) = 
\max \left\{ \frac{ n \cdot \vertexweight{\beliefstate}(\beliefstate') - 1}{n - 1} \ \middle|\   \beliefstate' \in \neighbourhood[\foundation]{\beliefstate}  \right\}.
\end{displaymath}
Intuitively, if the score of $\beliefstate$ is close to 1, $\beliefstate$ is close to one of the beliefs $\beliefstate'$ in its neighbourhood ($\vertexweight{\beliefstate}(\beliefstate') \approx 1$). If the score is close to 0, it has a large distance to all $\beliefstate'$ in its neighbourhood ($\vertexweight{\beliefstate}(\beliefstate') \approx 1/n$).
The score of an observation is obtained by taking the minimum score of any triangulated belief  with that observation times the current (relative) resolution for $\observation$, more precisely
\begin{displaymath}
\mathit{score}(\observation) = \min_{\beliefstate, \action} \Big(\mathit{score}(\nextbelief{\beliefstate}{\action}{\observation})\Big)  \cdot  \frac{\resolution[\observation]}{ \max_{\observation' \in \observations} \resolution[\observation']},
\end{displaymath}
where $\nextbelief{\beliefstate}{\action}{\observation}$ is as in \Cref{def:beliefmdp}.
To make sure that irrelevant parts of the abstraction MDP do not affect the score, we only consider belief states $\beliefstate$ and actions $\action$ that are reachable under some $\rho_{\scheds{}}$-optimal policy $\sched$.

We set $\observations_{\mathrm{extend}} = \set{ \observation \in \observations \mid \mathit{score}(\observation) \le \rho_{\observations}}$ for some threshold $\rho_{\observations} \in [0,1]$. 
In our implementation, we start with $\rho_{\observations} = 0.1$ and add $0.1 \cdot (1-\rho_{\observations})$ for each refinement step.
This way, $\rho_{\observations}$ approaches $1$ and thus every observation is eventually refined (unless it already has score 1, i.e. does not need refinement).

\section{Benchmarks}
\label{sec:benchmarks}

\paragraph{Input for \tool{Storm}}
Our implementation constructs POMDPs either from an explicit description or from a POMDP-extension of the \tool{Prism} language\footnote{\tool{Storm} rejects some POMDPs where action identifiers are missing: Whereas model checking MDPs does not require action names, these are essential in POMDPs.}~\cite{DBLP:journals/rts/Norman0Z17}. We have further extended the language, such that besides observing variable values, one can observe the values of arbitrary predicates.

\paragraph{Differences in models}
The model for \model{crypt} and \model{maze} are slightly different from the original due to a modelling error in the original formulation.

\paragraph{New models}
Our newly introduced models are grid-world based planning tasks.
\begin{itemize}
	\item In \model{drone} we search for a drone-plan to arrive at a target location, while avoiding a  randomly moving obstacle. The obstacle is only visible within a limited radius.
	\item In \model{refuel} we also search for a plan to arrive at a target. Movement is uncertain, and the own position is not observable. Obstacles are static. Additionally, any movement requires some energy. Energy can be refilled at recharging stations. 
	\item \model{rocks} describes a resource collection task. Some rocks need to be collected, and it is a-priori unknown which rocks to collect. Sensing is noisy and both sensing and collection is costly, so this yields an intricate trade-off.
\end{itemize}

\noindent Details can be found
\begin{center}
\url{https://github.com/moves-rwth/indefinite-horizon-pomdps}
\end{center}

\section{Additional Experiments}
\label{app:experiments}

We have done some further experiments which we omitted in the tight page limit. We used \emph{more models}, used \emph{an alternative method to determine the resolution}, and used \emph{an alternative set of `magic' constants in our implementation}. We report on the results below.

\subsection{Additional Benchmark instances and Approximation Sizes}
\Cref{tab:infall,tab:finall} report on our experiments on some additional model instances.
The experimental set-up is as in \Cref{sec:experiments}.
The displayed data is similar to \Cref{tab:inf,tab:fin}, except that we now also report on the size of the approximation MDP.
More precisely, the number of states $|\abstbeliefstates|$ of the approximation MDP is denoted after the ~|~ at the bottom line of each table cell.
In case of \tool{Prism}, this is the number of unknown grid points as reported by the tool.

We observe that several millions of belief states can be explored within the time- and memory limit.
We also note that the implementation in \tool{Prism} often considers far more grid points, which is a possible explanation for the superior performance of \tool{Storm} in many cases.

\begin{table}[p]
	\caption{Results for additional POMDP instances with infinite belief MDP.}
	\label{tab:infall}
	\scriptsize{
		\adjustbox{{max width=\textwidth}}{%
			\begin{tabular}{|cc||rr||r|r||r|r|r|r|r||r|r|}
\hline
\multicolumn{2}{|c||}{Benchmark} & \multicolumn{2}{c||}{Data} & \multicolumn{1}{c|}{MDP} & \multicolumn{1}{c||}{$\beliefmdp{\pomdp}$} & \multicolumn{2}{c|}{$\resolution$=4} & \multicolumn{3}{c||}{$\resolution$=12} & \multicolumn{2}{c|}{refine}\\
\multicolumn{1}{|c}{Model} & \multicolumn{1}{c||}{$\phi$} & \multicolumn{1}{c}{$\states$/$\actions$} & \multicolumn{1}{c||}{$\observations$} & \multicolumn{2}{c||}{\tool{Storm}} & \multicolumn{1}{c|}{\tool{Prism}} & \multicolumn{1}{c|}{\tool{Storm}} & \multicolumn{1}{c|}{\tool{Prism}} & \multicolumn{4}{c|}{\tool{Storm}}\\
\multicolumn{1}{|c}{} & \multicolumn{1}{c||}{} & \multicolumn{1}{c}{} & \multicolumn{1}{c||}{} & \multicolumn{1}{c|}{} & \multicolumn{1}{c||}{t=60} & \multicolumn{1}{c|}{} & \multicolumn{1}{c|}{$\rho_\mathit{gap}$=0} & \multicolumn{1}{c|}{} & \multicolumn{1}{c|}{$\rho_\mathit{gap}$=0} & \multicolumn{1}{c||}{$\rho_\mathit{gap}$=0.2} & \multicolumn{1}{c|}{t=60} & \multicolumn{1}{c|}{t=1800}\\
\hline\hline
\model{Drone} & \multirow{2}{*}{$P_\mathrm{max}$} & $1226$ & \multirow{2}{*}{$384$} & \multirow{2}{*}{0.98} & ${\ge}\, 0.84$ & \multirow{2}{*}{TO} & ${\le}\, \mathbf{0.96}$ & \multirow{2}{*}{MO} & \multirow{2}{*}{MO} & \multirow{2}{*}{MO} & ${\le}\, 0.97$ & ${\le}\, 0.97$$^\dagger$\\
4-1 &  & $3026$ &  &  & 6\,|\,$8{\cdot} 10^{5}$ &  & \ensuremath{6.67}\,|\,$2{\cdot} 10^{5}$ &  &  &  & 2\,|\,$4{\cdot} 10^{5}$ & 3\,|\,$3{\cdot} 10^{6}$\\
\hline
\model{Drone} & \multirow{2}{*}{$P_\mathrm{max}$} & $1226$ & \multirow{2}{*}{$761$} & \multirow{2}{*}{0.98} & ${\ge}\, 0.96$ & \multirow{2}{*}{TO} & ${\le}\, 0.98$ & \multirow{2}{*}{MO} & ${\le}\, \mathbf{0.97}$ & ${\le}\, \mathbf{0.97}$ & ${\le}\, \mathbf{0.97}$ & ${\le}\, \mathbf{0.97}$$^\dagger$\\
4-2 &  & $3026$ &  &  & 7\,|\,$1{\cdot} 10^{6}$ &  & \ensuremath{${<}\,$ 1}\,|\,$3{\cdot} 10^{4}$ &  & \ensuremath{194}\,|\,$4{\cdot} 10^{6}$ & \ensuremath{173}\,|\,$4{\cdot} 10^{6}$ & 3\,|\,$5{\cdot} 10^{5}$ & 4\,|\,$3{\cdot} 10^{6}$\\
\hline
\model{Drone} & \multirow{2}{*}{$P_\mathrm{max}$} & $2557$ & \multirow{2}{*}{$580$} & \multirow{2}{*}{0.99} & ${\ge}\, 0.79$ & \multirow{2}{*}{MO} & ${\le}\, \mathbf{0.98}$ & \multirow{2}{*}{MO} & \multirow{2}{*}{MO} & \multirow{2}{*}{MO} & ${\le}\, 0.99$ & ${\le}\, 0.99$$^\dagger$\\
5-1 &  & $6337$ &  &  & 5\,|\,$1{\cdot} 10^{6}$ &  & \ensuremath{56.6}\,|\,$2{\cdot} 10^{6}$ &  &  &  & 1\,|\,$2{\cdot} 10^{5}$ & 2\,|\,$3{\cdot} 10^{6}$\\
\hline
\model{Drone} & \multirow{2}{*}{$P_\mathrm{max}$} & $2557$ & \multirow{2}{*}{$1848$} & \multirow{2}{*}{0.99} & ${\ge}\, 0.9$ & \multirow{2}{*}{TO} & ${\le}\, \mathbf{0.99}$ & \multirow{2}{*}{MO} & \multirow{2}{*}{MO} & \multirow{2}{*}{MO} & ${\le}\, \mathbf{0.99}$ & ${\le}\, \mathbf{0.99}$$^\dagger$\\
5-3 &  & $6337$ &  &  & 5\,|\,$8{\cdot} 10^{5}$ &  & \ensuremath{3.21}\,|\,$8{\cdot} 10^{4}$ &  &  &  & 3\,|\,$1{\cdot} 10^{6}$ & 4\,|\,$8{\cdot} 10^{6}$\\
\hline\hline
\model{Grid-av} & \multirow{2}{*}{$P_\mathrm{max}$} & $17$ & \multirow{2}{*}{$4$} & \multirow{2}{*}{1} & ${\ge}\, 0.93$ & $[0.21, 1.0]$ & ${\le}\, 1$ & \multirow{2}{*}{MO} & ${\le}\, \mathbf{0.94}$ & ${\le}\, \mathbf{0.94}$ & ${\le}\, \mathbf{0.94}$ & ${\le}\, \mathbf{0.94}$$^\dagger$\\
4-0.1 &  & $59$ &  &  & 13\,|\,$1{\cdot} 10^{6}$ & \ensuremath{2.03}\,|\,$2382$ & \ensuremath{${<}\,$ 1}\,|\,$2043$ &  & \ensuremath{164}\,|\,$2{\cdot} 10^{6}$ & \ensuremath{168}\,|\,$2{\cdot} 10^{6}$ & 3\,|\,$3{\cdot} 10^{5}$ & 3\,|\,$3{\cdot} 10^{5}$\\
\hline
\model{Grid-av} & \multirow{2}{*}{$P_\mathrm{max}$} & $17$ & \multirow{2}{*}{$4$} & \multirow{2}{*}{1} & ${\ge}\, 0.9$ & \multirow{2}{*}{TO} & ${\le}\, 1$ & \multirow{2}{*}{MO} & ${\le}\, \mathbf{0.95}$ & ${\le}\, \mathbf{0.95}$ & ${\le}\, \mathbf{0.95}$ & ${\le}\, \mathbf{0.95}$$^\dagger$\\
4-0.3 &  & $59$ &  &  & 13\,|\,$1{\cdot} 10^{6}$ &  & \ensuremath{${<}\,$ 1}\,|\,$2166$ &  & \ensuremath{217}\,|\,$2{\cdot} 10^{6}$ & \ensuremath{212}\,|\,$2{\cdot} 10^{6}$ & 3\,|\,$3{\cdot} 10^{5}$ & 3\,|\,$3{\cdot} 10^{5}$\\
\hline\hline
\model{Grid} & \multirow{2}{*}{$R_\mathrm{min}$} & $17$ & \multirow{2}{*}{$3$} & \multirow{2}{*}{3.56} & ${\le}\, 4.7$ & $[4.06, 4.7]$ & ${\ge}\, 4.06$ & \multirow{2}{*}{MO} & ${\ge}\, 4.59$ & ${\ge}\, 4.59$ & ${\ge}\, 4.56$ & ${\ge}\, \mathbf{4.61}$$^\dagger$\\
4-0.1 &  & $62$ &  &  & 13\,|\,$1{\cdot} 10^{6}$ & \ensuremath{2.02}\,|\,$3061$ & \ensuremath{${<}\,$ 1}\,|\,$1655$ &  & \ensuremath{264}\,|\,$3{\cdot} 10^{6}$ & \ensuremath{268}\,|\,$3{\cdot} 10^{6}$ & 3\,|\,$2{\cdot} 10^{5}$ & 4\,|\,$4{\cdot} 10^{6}$\\
\hline
\model{Grid} & \multirow{2}{*}{$R_\mathrm{min}$} & $17$ & \multirow{2}{*}{$3$} & \multirow{2}{*}{4.57} & ${\le}\, 6.37$ & $[5.4, 6.31]$ & ${\ge}\, 5.4$ & \multirow{2}{*}{MO} & ${\ge}\, \mathbf{6.18}$ & ${\ge}\, \mathbf{6.18}$ & ${\ge}\, 5.92$ & ${\ge}\, 5.92$$^\dagger$\\
4-0.3 &  & $62$ &  &  & 13\,|\,$1{\cdot} 10^{6}$ & \ensuremath{3.05}\,|\,$3061$ & \ensuremath{${<}\,$ 1}\,|\,$1610$ &  & \ensuremath{217}\,|\,$3{\cdot} 10^{6}$ & \ensuremath{214}\,|\,$3{\cdot} 10^{6}$ & 3\,|\,$2{\cdot} 10^{5}$ & 4\,|\,$4{\cdot} 10^{6}$\\
\hline\hline
\model{Maze2} & \multirow{2}{*}{$R_\mathrm{min}$} & $15$ & \multirow{2}{*}{$8$} & \multirow{2}{*}{5.64} & ${\le}\, 6.32$ & $[6.29, 6.32]$ & ${\ge}\, 6.29$ & $[\mathbf{6.32}, 6.32]$ & ${\ge}\, \mathbf{6.32}$ & ${\ge}\, \mathbf{6.32}$ & ${\ge}\, \mathbf{6.32}$ & ${\ge}\, \mathbf{6.32}$$^\dagger$\\
0.1 &  & $54$ &  &  & 14\,|\,$7{\cdot} 10^{5}$ & \ensuremath{1.35}\,|\,$140$ & \ensuremath{${<}\,$ 1}\,|\,$71$ & \ensuremath{4.91}\,|\,$6218$ & \ensuremath{${<}\,$ 1}\,|\,$733$ & \ensuremath{${<}\,$ 1}\,|\,$731$ & 7\,|\,$1{\cdot} 10^{6}$ & 8\,|\,$6{\cdot} 10^{6}$\\
\hline
\model{Maze2} & \multirow{2}{*}{$R_\mathrm{min}$} & $15$ & \multirow{2}{*}{$8$} & \multirow{2}{*}{7.25} & ${\le}\, 8.13$ & $[7.99, 8.13]$ & ${\ge}\, 7.99$ & $[\mathbf{8.13}, 8.13]$ & ${\ge}\, \mathbf{8.13}$ & ${\ge}\, \mathbf{8.13}$ & ${=}\, \mathbf{8.13}$* & ${=}\, \mathbf{8.13}$*\\
0.3 &  & $54$ &  &  & 14\,|\,$7{\cdot} 10^{5}$ & \ensuremath{1.59}\,|\,$140$ & \ensuremath{${<}\,$ 1}\,|\,$86$ & \ensuremath{6.6}\,|\,$6218$ & \ensuremath{${<}\,$ 1}\,|\,$1343$ & \ensuremath{${<}\,$ 1}\,|\,$1341$ & 7\,|\,$4{\cdot} 10^{5}$ & 7\,|\,$4{\cdot} 10^{5}$\\
\hline\hline
\model{Refuel} & \multirow{2}{*}{$P_\mathrm{max}$} & $208$ & \multirow{2}{*}{$50$} & \multirow{2}{*}{0.98} & ${\ge}\, 0.67$ & \multirow{2}{*}{TO} & ${\le}\, 0.71$ & \multirow{2}{*}{MO} & ${\le}\, 0.68$ & ${\le}\, 0.68$ & ${=}\, \mathbf{0.67}$* & ${=}\, \mathbf{0.67}$*\\
06 &  & $574$ &  &  & 10\,|\,$2{\cdot} 10^{6}$ &  & \ensuremath{${<}\,$ 1}\,|\,$5486$ &  & \ensuremath{2.08}\,|\,$1{\cdot} 10^{5}$ & \ensuremath{2.08}\,|\,$1{\cdot} 10^{5}$ & 59\,|\,$2{\cdot} 10^{4}$ & 59\,|\,$2{\cdot} 10^{4}$\\
\hline
\model{Refuel} & \multirow{2}{*}{$P_\mathrm{max}$} & $470$ & \multirow{2}{*}{$66$} & \multirow{2}{*}{0.99} & ${\ge}\, 0.45$ & \multirow{2}{*}{MO} & ${\le}\, 0.76$ & \multirow{2}{*}{MO} & \multirow{2}{*}{MO} & \multirow{2}{*}{MO} & ${\le}\, 0.75$ & ${\le}\, \mathbf{0.58}$$^\dagger$\\
08 &  & $1446$ &  &  & 7\,|\,$1{\cdot} 10^{6}$ &  & \ensuremath{7.3}\,|\,$1{\cdot} 10^{5}$ &  &  &  & 2\,|\,$1{\cdot} 10^{5}$ & 3\,|\,$1{\cdot} 10^{6}$\\
\hline
\model{Refuel} & \multirow{2}{*}{$P_\mathrm{max}$} & $892$ & \multirow{2}{*}{$84$} & \multirow{2}{*}{1.0} & ${\ge}\, 0.43$ & \multirow{2}{*}{MO} & ${\le}\, \mathbf{0.83}$ & \multirow{2}{*}{MO} & \multirow{2}{*}{MO} & \multirow{2}{*}{MO} & ${\le}\, 0.87$ & ${\le}\, 0.87$$^\dagger$\\
10 &  & $2894$ &  &  & 5\,|\,$1{\cdot} 10^{6}$ &  & \ensuremath{48}\,|\,$1{\cdot} 10^{6}$ &  &  &  & 1\,|\,$1{\cdot} 10^{5}$ & 2\,|\,$1{\cdot} 10^{6}$\\
\hline\hline
\model{Rocks} & \multirow{2}{*}{$R_\mathrm{min}$} & $3241$ & \multirow{2}{*}{$817$} & \multirow{2}{*}{11} & ${\le}\, 26.2$ & \multirow{2}{*}{TO} & ${\ge}\, \mathbf{14}$ & \multirow{2}{*}{MO} & ${\ge}\, \mathbf{14}$ & ${\ge}\, \mathbf{14}$ & ${=}\, \mathbf{14}$* & ${=}\, \mathbf{14}$*\\
08 &  & $2{\cdot} 10^{4}$ &  &  & 7\,|\,$8{\cdot} 10^{5}$ &  & \ensuremath{${<}\,$ 1}\,|\,$2{\cdot} 10^{4}$ &  & \ensuremath{7.99}\,|\,$2{\cdot} 10^{5}$ & \ensuremath{7.99}\,|\,$2{\cdot} 10^{5}$ & 9\,|\,$2{\cdot} 10^{4}$ & 9\,|\,$2{\cdot} 10^{4}$\\
\hline
\model{Rocks} & \multirow{2}{*}{$R_\mathrm{min}$} & $6553$ & \multirow{2}{*}{$1645$} & \multirow{2}{*}{16.5} & ${\le}\, 35.4$ & \multirow{2}{*}{TO} & ${\ge}\, 19.9$ & \multirow{2}{*}{MO} & ${\ge}\, \mathbf{20}$ & ${\ge}\, \mathbf{20}$ & ${=}\, \mathbf{20}$* & ${=}\, \mathbf{20}$*\\
12 &  & $3{\cdot} 10^{4}$ &  &  & 6\,|\,$8{\cdot} 10^{5}$ &  & \ensuremath{1.26}\,|\,$4{\cdot} 10^{4}$ &  & \ensuremath{18.9}\,|\,$5{\cdot} 10^{5}$ & \ensuremath{19.1}\,|\,$5{\cdot} 10^{5}$ & 9\,|\,$5{\cdot} 10^{4}$ & 9\,|\,$5{\cdot} 10^{4}$\\
\hline
\model{Rocks} & \multirow{2}{*}{$R_\mathrm{min}$} & $1{\cdot} 10^{4}$ & \multirow{2}{*}{$2761$} & \multirow{2}{*}{22} & ${\le}\, 44$ & \multirow{2}{*}{MO} & ${\ge}\, 25.6$ & \multirow{2}{*}{MO} & ${\ge}\, \mathbf{26}$ & ${\ge}\, \mathbf{26}$ & ${\ge}\, 25.9$ & ${\ge}\, 25.9$\\
16 &  & $5{\cdot} 10^{4}$ &  &  & 5\,|\,$7{\cdot} 10^{5}$ &  & \ensuremath{2.55}\,|\,$7{\cdot} 10^{4}$ &  & \ensuremath{37.2}\,|\,$9{\cdot} 10^{5}$ & \ensuremath{35.9}\,|\,$8{\cdot} 10^{5}$ & 8\,|\,$3{\cdot} 10^{5}$ & 9\,|\,$2{\cdot} 10^{6}$\\
\hline
\end{tabular}

		}
	}
\end{table}
\begin{table}[p]
	\caption{Results for additional POMDP instances with finite belief MDP.}
	\label{tab:finall}
	\scriptsize{
		\adjustbox{{max width=\textwidth}}{%
			\begin{tabular}{|cc||rr||r|r||r|r|r|r|r||r|r|}
\hline
\multicolumn{2}{|c||}{Benchmark} & \multicolumn{2}{c||}{Data} & \multicolumn{1}{c|}{MDP} & \multicolumn{1}{c||}{$\beliefmdp{\pomdp}$} & \multicolumn{2}{c|}{$\resolution$=4} & \multicolumn{3}{c||}{$\resolution$=12} & \multicolumn{2}{c|}{refine}\\
\multicolumn{1}{|c}{Model} & \multicolumn{1}{c||}{$\phi$} & \multicolumn{1}{c}{$\states$/$\actions$} & \multicolumn{1}{c||}{$\observations$} & \multicolumn{2}{c||}{\tool{Storm}} & \multicolumn{1}{c|}{\tool{Prism}} & \multicolumn{1}{c|}{\tool{Storm}} & \multicolumn{1}{c|}{\tool{Prism}} & \multicolumn{4}{c|}{\tool{Storm}}\\
\multicolumn{1}{|c}{} & \multicolumn{1}{c||}{} & \multicolumn{1}{c}{} & \multicolumn{1}{c||}{} & \multicolumn{1}{c|}{} & \multicolumn{1}{c||}{} & \multicolumn{1}{c|}{} & \multicolumn{1}{c|}{$\rho_\mathit{gap}$=0} & \multicolumn{1}{c|}{} & \multicolumn{1}{c|}{$\rho_\mathit{gap}$=0} & \multicolumn{1}{c||}{$\rho_\mathit{gap}$=0.2} & \multicolumn{1}{c|}{t=60} & \multicolumn{1}{c|}{t=1800}\\
\hline\hline
\model{Crypt} & \multirow{2}{*}{$P_\mathrm{max}$} & $1972$ & \multirow{2}{*}{$510$} & \multirow{2}{*}{1} & ${=}\, \mathbf{0.33}$ & $[0.33, 0.79]$ & ${\le}\, 0.79$ & \multirow{2}{*}{MO} & ${\le}\, \mathbf{0.33}$ & ${\le}\, \mathbf{0.33}$ & ${=}\, \mathbf{0.33}$* & ${=}\, \mathbf{0.33}$*\\
4 &  & $4612$ &  &  & \ensuremath{3.51}\,|\,$912$ & \ensuremath{20.3}\,|\,$5{\cdot} 10^{4}$ & \ensuremath{${<}\,$ 1}\,|\,$5126$ &  & \ensuremath{1.36}\,|\,$464$ & \ensuremath{6.12}\,|\,$464$ & 6\,|\,$564$ & 6\,|\,$564$\\
\hline
\model{Crypt} & \multirow{2}{*}{$P_\mathrm{min}$} & $1972$ & \multirow{2}{*}{$510$} & \multirow{2}{*}{0} & ${=}\, \mathbf{0.33}$ & $[0, 0.33]$ & ${\ge}\, 0$ & \multirow{2}{*}{MO} & ${\ge}\, \mathbf{0.33}$ & ${\ge}\, \mathbf{0.33}$ & ${=}\, \mathbf{0.33}$* & ${=}\, \mathbf{0.33}$*\\
4 &  & $4612$ &  &  & \ensuremath{${<}\,$ 1}\,|\,$912$ & \ensuremath{17.6}\,|\,$5{\cdot} 10^{4}$ & \ensuremath{2.75}\,|\,$5126$ &  & \ensuremath{7.79}\,|\,$464$ & \ensuremath{2.69}\,|\,$464$ & 32\,|\,$2000$ & 32\,|\,$2000$\\
\hline
\model{Crypt} & \multirow{2}{*}{$P_\mathrm{max}$} & $7{\cdot} 10^{4}$ & \multirow{2}{*}{$6678$} & \multirow{2}{*}{1} & ${=}\, \mathbf{0.2}$ & \multirow{2}{*}{MO} & ${\le}\, 1$ & \multirow{2}{*}{MO} & ${\le}\, 0.84$ & ${\le}\, 0.84$ & ${\le}\, 0.97$ & ${\le}\, 0.94$\\
6 &  & $2{\cdot} 10^{5}$ &  &  & \ensuremath{8.47}\,|\,$2{\cdot} 10^{4}$ &  & \ensuremath{17.8}\,|\,$1{\cdot} 10^{5}$ &  & \ensuremath{155}\,|\,$2{\cdot} 10^{6}$ & \ensuremath{159}\,|\,$2{\cdot} 10^{6}$ & 2\,|\,$3{\cdot} 10^{5}$ & 4\,|\,$7{\cdot} 10^{6}$\\
\hline
\model{Crypt} & \multirow{2}{*}{$P_\mathrm{min}$} & $7{\cdot} 10^{4}$ & \multirow{2}{*}{$6678$} & \multirow{2}{*}{0} & ${=}\, \mathbf{0.2}$ & \multirow{2}{*}{MO} & ${\ge}\, 0$ & \multirow{2}{*}{MO} & ${\ge}\, 0$ & ${\ge}\, 0$ & ${\ge}\, 0$ & ${\ge}\, 0$\\
6 &  & $2{\cdot} 10^{5}$ &  &  & \ensuremath{6.59}\,|\,$2{\cdot} 10^{4}$ &  & \ensuremath{17.8}\,|\,$1{\cdot} 10^{5}$ &  & \ensuremath{157}\,|\,$2{\cdot} 10^{6}$ & \ensuremath{158}\,|\,$2{\cdot} 10^{6}$ & 2\,|\,$4{\cdot} 10^{5}$ & 4\,|\,$1{\cdot} 10^{7}$\\
\hline\hline
\model{Grid-av} & \multirow{2}{*}{$P_\mathrm{max}$} & $17$ & \multirow{2}{*}{$4$} & \multirow{2}{*}{1} & ${=}\, \mathbf{0.93}$ & $[0.21, 1.0]$ & ${\le}\, 1$ & \multirow{2}{*}{MO} & ${\le}\, 0.94$ & ${\le}\, 0.94$ & ${\le}\, \mathbf{0.93}$ & ${\le}\, \mathbf{0.93}$$^\dagger$\\
4-0 &  & $59$ &  &  & \ensuremath{${<}\,$ 1}\,|\,$9016$ & \ensuremath{1.51}\,|\,$2382$ & \ensuremath{${<}\,$ 1}\,|\,$378$ &  & \ensuremath{${<}\,$ 1}\,|\,$4{\cdot} 10^{4}$ & \ensuremath{${<}\,$ 1}\,|\,$4{\cdot} 10^{4}$ & 9\,|\,$6{\cdot} 10^{5}$ & 26\,|\,$1{\cdot} 10^{7}$\\
\hline\hline
\model{Grid} & \multirow{2}{*}{$R_\mathrm{min}$} & $17$ & \multirow{2}{*}{$3$} & \multirow{2}{*}{3.2} & ${=}\, \mathbf{4.13}$ & $[3.6, 4.13]$ & ${\ge}\, 3.6$ & \multirow{2}{*}{MO} & ${\ge}\, 4.03$ & ${\ge}\, 4.03$ & ${\ge}\, \mathbf{4.13}$ & ${\ge}\, \mathbf{4.13}$\\
4-0 &  & $62$ &  &  & \ensuremath{${<}\,$ 1}\,|\,$2423$ & \ensuremath{1.67}\,|\,$3061$ & \ensuremath{${<}\,$ 1}\,|\,$431$ &  & \ensuremath{${<}\,$ 1}\,|\,$7748$ & \ensuremath{${<}\,$ 1}\,|\,$7748$ & 25\,|\,$2{\cdot} 10^{5}$ & 26\,|\,$9{\cdot} 10^{5}$\\
\hline\hline
\model{Maze2} & \multirow{2}{*}{$R_\mathrm{min}$} & $15$ & \multirow{2}{*}{$8$} & \multirow{2}{*}{5.08} & ${=}\, \mathbf{5.69}$ & $[\mathbf{5.69}, 5.69]$ & ${\ge}\, \mathbf{5.69}$ & $[\mathbf{5.69}, 5.69]$ & ${\ge}\, \mathbf{5.69}$ & ${\ge}\, \mathbf{5.69}$ & ${=}\, \mathbf{5.69}$* & ${=}\, \mathbf{5.69}$*\\
0 &  & $54$ &  &  & \ensuremath{${<}\,$ 1}\,|\,$26$ & \ensuremath{1.43}\,|\,$140$ & \ensuremath{${<}\,$ 1}\,|\,$27$ & \ensuremath{3.17}\,|\,$6218$ & \ensuremath{${<}\,$ 1}\,|\,$23$ & \ensuremath{${<}\,$ 1}\,|\,$21$ & 4\,|\,$23$ & 4\,|\,$23$\\
\hline\hline
\model{Netw-p} & \multirow{2}{*}{$R_\mathrm{max}$} & $2{\cdot} 10^{4}$ & \multirow{2}{*}{$4909$} & \multirow{2}{*}{566} & ${=}\, \mathbf{557}$ & $[557, 559]$ & ${\le}\, 560$ & \multirow{2}{*}{TO} & ${\le}\, \mathbf{557}$ & ${\le}\, 566$ & ${\le}\, \mathbf{557}$ & ${\le}\, \mathbf{557}$\\
2-8-20 &  & $3{\cdot} 10^{4}$ &  &  & \ensuremath{612}\,|\,$3{\cdot} 10^{7}$ & \ensuremath{503}\,|\,$2{\cdot} 10^{5}$ & \ensuremath{2.17}\,|\,$7{\cdot} 10^{4}$ &  & \ensuremath{4.25}\,|\,$2{\cdot} 10^{5}$ & \ensuremath{${<}\,$ 1}\,|\,$2$ & 10\,|\,$8{\cdot} 10^{5}$ & 18\,|\,$2{\cdot} 10^{7}$\\
\hline
\model{Netw-p} & \multirow{2}{*}{$R_\mathrm{max}$} & $8019$ & \multirow{2}{*}{$1035$} & \multirow{2}{*}{73.6} & ${=}\, \mathbf{64.3}$ & $[64.1, 67.4]$ & ${\le}\, 69$ & \multirow{2}{*}{MO} & ${\le}\, 65.3$ & ${\le}\, 66.4$ & ${\le}\, 65.7$ & ${\le}\, \mathbf{64.3}$$^\dagger$\\
3-5-2 &  & $2{\cdot} 10^{4}$ &  &  & \ensuremath{71.6}\,|\,$7{\cdot} 10^{6}$ & \ensuremath{300}\,|\,$3{\cdot} 10^{5}$ & \ensuremath{3.48}\,|\,$7{\cdot} 10^{4}$ &  & \ensuremath{26.3}\,|\,$4{\cdot} 10^{5}$ & \ensuremath{25}\,|\,$4{\cdot} 10^{5}$ & 3\,|\,$6{\cdot} 10^{5}$ & 6\,|\,$1{\cdot} 10^{7}$\\
\hline
\model{Netw-p} & \multirow{2}{*}{$R_\mathrm{max}$} & $2{\cdot} 10^{5}$ & \multirow{2}{*}{$2{\cdot} 10^{4}$} & \multirow{2}{*}{849} & \multirow{2}{*}{TO} & \multirow{2}{*}{TO} & ${\le}\, 832$ & \multirow{2}{*}{MO} & \multirow{2}{*}{TO} & ${\le}\, 849$ & ${\le}\, 849$ & ${\le}\, \mathbf{825}$\\
3-8-20 &  & $3{\cdot} 10^{5}$ &  &  &  &  & \ensuremath{514}\,|\,$2{\cdot} 10^{6}$ &  &  & \ensuremath{8.2}\,|\,$2$ & 0\,|\,~--~ & 2\,|\,$4{\cdot} 10^{6}$\\
\hline\hline
\model{Netw} & \multirow{2}{*}{$R_\mathrm{min}$} & $4589$ & \multirow{2}{*}{$1173$} & \multirow{2}{*}{2.56} & ${=}\, \mathbf{3.2}$ & $[3.03, 3.2]$ & ${\ge}\, 2.97$ & $[3.17, 3.2]$ & ${\ge}\, 3.17$ & ${\ge}\, 3.16$ & ${\ge}\, \mathbf{3.2}$ & ${\ge}\, \mathbf{3.2}$\\
2-8-20 &  & $6973$ &  &  & \ensuremath{38.4}\,|\,$4{\cdot} 10^{6}$ & \ensuremath{42.1}\,|\,$4{\cdot} 10^{4}$ & \ensuremath{${<}\,$ 1}\,|\,$1{\cdot} 10^{4}$ & \ensuremath{521}\,|\,$5{\cdot} 10^{5}$ & \ensuremath{${<}\,$ 1}\,|\,$4{\cdot} 10^{4}$ & \ensuremath{${<}\,$ 1}\,|\,$4{\cdot} 10^{4}$ & 10\,|\,$2{\cdot} 10^{6}$ & 23\,|\,$2{\cdot} 10^{7}$\\
\hline
\model{Netw} & \multirow{2}{*}{$R_\mathrm{min}$} & $843$ & \multirow{2}{*}{$111$} & \multirow{2}{*}{1.18} & ${=}\, \mathbf{1.9}$ & $[1.64, 1.92]$ & ${\ge}\, 1.53$ & $[1.85, 1.9]$ & ${\ge}\, 1.81$ & ${\ge}\, 1.75$ & ${\ge}\, \mathbf{1.9}$ & ${\ge}\, \mathbf{1.9}$\\
3-5-2 &  & $1515$ &  &  & \ensuremath{3.12}\,|\,$4{\cdot} 10^{5}$ & \ensuremath{5.65}\,|\,$3{\cdot} 10^{4}$ & \ensuremath{${<}\,$ 1}\,|\,$7759$ & \ensuremath{835}\,|\,$5{\cdot} 10^{6}$ & \ensuremath{${<}\,$ 1}\,|\,$4{\cdot} 10^{4}$ & \ensuremath{${<}\,$ 1}\,|\,$4{\cdot} 10^{4}$ & 7\,|\,$1{\cdot} 10^{6}$ & 19\,|\,$2{\cdot} 10^{7}$\\
\hline
\model{Netw} & \multirow{2}{*}{$R_\mathrm{min}$} & $2{\cdot} 10^{4}$ & \multirow{2}{*}{$2205$} & \multirow{2}{*}{3.88} & \multirow{2}{*}{MO} & $[5.54, 6.77]$ & ${\ge}\, 5.11$ & \multirow{2}{*}{MO} & ${\ge}\, 6.35$ & ${\ge}\, 6.33$ & ${\ge}\, 6.26$ & ${\ge}\, \mathbf{6.72}$$^\dagger$\\
3-8-20 &  & $3{\cdot} 10^{4}$ &  &  &  & \ensuremath{1777}\,|\,$7{\cdot} 10^{5}$ & \ensuremath{4.82}\,|\,$2{\cdot} 10^{5}$ &  & \ensuremath{34.5}\,|\,$1{\cdot} 10^{6}$ & \ensuremath{34.3}\,|\,$1{\cdot} 10^{6}$ & 3\,|\,$2{\cdot} 10^{6}$ & 5\,|\,$1{\cdot} 10^{7}$\\
\hline\hline
\model{Nrp} & \multirow{2}{*}{$P_\mathrm{max}$} & $125$ & \multirow{2}{*}{$41$} & \multirow{2}{*}{1} & ${=}\, \mathbf{0.12}$ & $[0.13, 0.38]$ & ${\le}\, 0.38$ & $[0.13, 0.22]$ & ${\le}\, 0.22$ & ${\le}\, 0.22$ & ${=}\, \mathbf{0.12}$* & ${=}\, \mathbf{0.12}$*\\
8 &  & $161$ &  &  & \ensuremath{${<}\,$ 1}\,|\,$52$ & \ensuremath{1.57}\,|\,$2385$ & \ensuremath{${<}\,$ 1}\,|\,$101$ & \ensuremath{22.9}\,|\,$2{\cdot} 10^{5}$ & \ensuremath{${<}\,$ 1}\,|\,$965$ & \ensuremath{${<}\,$ 1}\,|\,$965$ & 70\,|\,$32$ & 70\,|\,$32$\\
\hline
\end{tabular}

		}
	}
\end{table}

\subsection{Evaluation of the Dynamic Triangulation Approach}
\label{app:experiments:dyn}
\Cref{tab:infdyn,tab:findyn} show the experimental results for the dynamic approach for triangulating beliefs as discussed in \Cref{app:whichobs}.
Again, the set-up is as in \Cref{sec:experiments}, except for the different triangulation scheme.

Comparing with the results for the standard triangulation approach with static resolutions (\Cref{tab:infall,tab:finall}), we often observe that the dynamic approach yields smaller approximations for the finite belief MDPs (\Cref{tab:findyn}, but larger approximations for the infinite ones (\Cref{tab:infdyn}).

\begin{table}[p]
	\caption{Results for POMDPs with infinite belief MDP using the dynamic triangulation approach.}
	\label{tab:infdyn}
	\scriptsize{
		\adjustbox{{max width=\textwidth}}{%
			\begin{tabular}{|cc||rr||r|r||r|r|r|r|r||r|r|}
\hline
\multicolumn{2}{|c||}{Benchmark} & \multicolumn{2}{c||}{Data} & \multicolumn{1}{c|}{MDP} & \multicolumn{1}{c||}{$\beliefmdp{\pomdp}$} & \multicolumn{2}{c|}{$\resolution$=4} & \multicolumn{3}{c||}{$\resolution$=12} & \multicolumn{2}{c|}{refine}\\
\multicolumn{1}{|c}{Model} & \multicolumn{1}{c||}{$\phi$} & \multicolumn{1}{c}{$\states$/$\actions$} & \multicolumn{1}{c||}{$\observations$} & \multicolumn{2}{c||}{\tool{Storm}} & \multicolumn{1}{c|}{\tool{Prism}} & \multicolumn{1}{c|}{\tool{Storm}} & \multicolumn{1}{c|}{\tool{Prism}} & \multicolumn{4}{c|}{\tool{Storm}}\\
\multicolumn{1}{|c}{} & \multicolumn{1}{c||}{} & \multicolumn{1}{c}{} & \multicolumn{1}{c||}{} & \multicolumn{1}{c|}{} & \multicolumn{1}{c||}{t=60} & \multicolumn{1}{c|}{} & \multicolumn{1}{c|}{$\rho_\mathit{gap}$=0} & \multicolumn{1}{c|}{} & \multicolumn{1}{c|}{$\rho_\mathit{gap}$=0} & \multicolumn{1}{c||}{$\rho_\mathit{gap}$=0.2} & \multicolumn{1}{c|}{t=60} & \multicolumn{1}{c|}{t=1800}\\
\hline\hline
\model{Drone} & \multirow{2}{*}{$P_\mathrm{max}$} & $1226$ & \multirow{2}{*}{$384$} & \multirow{2}{*}{0.98} & ${\ge}\, 0.84$ & \multirow{2}{*}{TO} & ${\le}\, \mathbf{0.96}$ & \multirow{2}{*}{MO} & \multirow{2}{*}{MO} & \multirow{2}{*}{MO} & ${\le}\, \mathbf{0.96}$ & ${\le}\, \mathbf{0.96}$$^\dagger$\\
4-1 &  & $3026$ &  &  & 6\,|\,$8{\cdot} 10^{5}$ &  & \ensuremath{5.84}\,|\,$2{\cdot} 10^{5}$ &  &  &  & 2\,|\,$1{\cdot} 10^{5}$ & 3\,|\,$1{\cdot} 10^{6}$\\
\hline
\model{Drone} & \multirow{2}{*}{$P_\mathrm{max}$} & $1226$ & \multirow{2}{*}{$761$} & \multirow{2}{*}{0.98} & ${\ge}\, 0.96$ & \multirow{2}{*}{TO} & ${\le}\, 0.98$ & \multirow{2}{*}{MO} & ${\le}\, \mathbf{0.97}$ & ${\le}\, \mathbf{0.97}$ & ${\le}\, 0.98$ & ${\le}\, 0.98$$^\dagger$\\
4-2 &  & $3026$ &  &  & 7\,|\,$1{\cdot} 10^{6}$ &  & \ensuremath{${<}\,$ 1}\,|\,$3{\cdot} 10^{4}$ &  & \ensuremath{149}\,|\,$3{\cdot} 10^{6}$ & \ensuremath{122}\,|\,$3{\cdot} 10^{6}$ & 3\,|\,$2{\cdot} 10^{5}$ & 4\,|\,$2{\cdot} 10^{6}$\\
\hline
\model{Drone} & \multirow{2}{*}{$P_\mathrm{max}$} & $2557$ & \multirow{2}{*}{$580$} & \multirow{2}{*}{0.99} & ${\ge}\, 0.79$ & \multirow{2}{*}{MO} & ${\le}\, \mathbf{0.98}$ & \multirow{2}{*}{MO} & \multirow{2}{*}{MO} & \multirow{2}{*}{MO} & ${\le}\, \mathbf{0.98}$ & ${\le}\, \mathbf{0.98}$$^\dagger$\\
5-1 &  & $6337$ &  &  & 5\,|\,$1{\cdot} 10^{6}$ &  & \ensuremath{53.4}\,|\,$1{\cdot} 10^{6}$ &  &  &  & 2\,|\,$7{\cdot} 10^{5}$ & 3\,|\,$7{\cdot} 10^{6}$\\
\hline
\model{Drone} & \multirow{2}{*}{$P_\mathrm{max}$} & $2557$ & \multirow{2}{*}{$1848$} & \multirow{2}{*}{0.99} & ${\ge}\, 0.9$ & \multirow{2}{*}{TO} & ${\le}\, \mathbf{0.99}$ & \multirow{2}{*}{MO} & \multirow{2}{*}{MO} & \multirow{2}{*}{MO} & ${\le}\, \mathbf{0.99}$ & ${\le}\, \mathbf{0.99}$$^\dagger$\\
5-3 &  & $6337$ &  &  & 5\,|\,$8{\cdot} 10^{5}$ &  & \ensuremath{2.93}\,|\,$7{\cdot} 10^{4}$ &  &  &  & 3\,|\,$4{\cdot} 10^{5}$ & 4\,|\,$3{\cdot} 10^{6}$\\
\hline\hline
\model{Grid-av} & \multirow{2}{*}{$P_\mathrm{max}$} & $17$ & \multirow{2}{*}{$4$} & \multirow{2}{*}{1} & ${\ge}\, 0.93$ & $[0.21, 1.0]$ & ${\le}\, 1$ & \multirow{2}{*}{MO} & ${\le}\, \mathbf{0.94}$ & ${\le}\, \mathbf{0.94}$ & ${\le}\, \mathbf{0.94}$ & ${\le}\, \mathbf{0.94}$$^\dagger$\\
4-0.1 &  & $59$ &  &  & 13\,|\,$1{\cdot} 10^{6}$ & \ensuremath{2.03}\,|\,$2382$ & \ensuremath{${<}\,$ 1}\,|\,$2043$ &  & \ensuremath{256}\,|\,$3{\cdot} 10^{6}$ & \ensuremath{259}\,|\,$3{\cdot} 10^{6}$ & 3\,|\,$3{\cdot} 10^{5}$ & 3\,|\,$3{\cdot} 10^{5}$\\
\hline
\model{Grid-av} & \multirow{2}{*}{$P_\mathrm{max}$} & $17$ & \multirow{2}{*}{$4$} & \multirow{2}{*}{1} & ${\ge}\, 0.9$ & \multirow{2}{*}{TO} & ${\le}\, 1$ & \multirow{2}{*}{MO} & ${\le}\, \mathbf{0.95}$ & ${\le}\, \mathbf{0.95}$ & ${\le}\, \mathbf{0.95}$ & ${\le}\, \mathbf{0.95}$$^\dagger$\\
4-0.3 &  & $59$ &  &  & 13\,|\,$1{\cdot} 10^{6}$ &  & \ensuremath{${<}\,$ 1}\,|\,$2166$ &  & \ensuremath{354}\,|\,$4{\cdot} 10^{6}$ & \ensuremath{382}\,|\,$4{\cdot} 10^{6}$ & 3\,|\,$3{\cdot} 10^{5}$ & 3\,|\,$3{\cdot} 10^{5}$\\
\hline\hline
\model{Grid} & \multirow{2}{*}{$R_\mathrm{min}$} & $17$ & \multirow{2}{*}{$3$} & \multirow{2}{*}{3.56} & ${\le}\, 4.7$ & $[4.06, 4.7]$ & ${\ge}\, 4.06$ & \multirow{2}{*}{MO} & ${\ge}\, 4.59$ & ${\ge}\, 4.59$ & ${\ge}\, 4.56$ & ${\ge}\, \mathbf{4.65}$$^\dagger$\\
4-0.1 &  & $62$ &  &  & 13\,|\,$1{\cdot} 10^{6}$ & \ensuremath{2.02}\,|\,$3061$ & \ensuremath{${<}\,$ 1}\,|\,$1655$ &  & \ensuremath{607}\,|\,$7{\cdot} 10^{6}$ & \ensuremath{631}\,|\,$7{\cdot} 10^{6}$ & 3\,|\,$2{\cdot} 10^{5}$ & 4\,|\,$6{\cdot} 10^{6}$\\
\hline
\model{Grid} & \multirow{2}{*}{$R_\mathrm{min}$} & $17$ & \multirow{2}{*}{$3$} & \multirow{2}{*}{4.57} & ${\le}\, 6.37$ & $[5.4, 6.31]$ & ${\ge}\, 5.4$ & \multirow{2}{*}{MO} & ${\ge}\, \mathbf{6.18}$ & ${\ge}\, \mathbf{6.18}$ & ${\ge}\, 5.9$ & ${\ge}\, 5.9$$^\dagger$\\
4-0.3 &  & $62$ &  &  & 13\,|\,$1{\cdot} 10^{6}$ & \ensuremath{3.05}\,|\,$3061$ & \ensuremath{${<}\,$ 1}\,|\,$1610$ &  & \ensuremath{546}\,|\,$6{\cdot} 10^{6}$ & \ensuremath{529}\,|\,$6{\cdot} 10^{6}$ & 3\,|\,$2{\cdot} 10^{5}$ & 3\,|\,$2{\cdot} 10^{5}$\\
\hline\hline
\model{Maze2} & \multirow{2}{*}{$R_\mathrm{min}$} & $15$ & \multirow{2}{*}{$8$} & \multirow{2}{*}{5.64} & ${\le}\, 6.32$ & $[6.29, 6.32]$ & ${\ge}\, 6.29$ & $[\mathbf{6.32}, 6.32]$ & ${\ge}\, \mathbf{6.32}$ & ${\ge}\, \mathbf{6.32}$ & ${\ge}\, \mathbf{6.32}$ & ${\ge}\, \mathbf{6.32}$$^\dagger$\\
0.1 &  & $54$ &  &  & 14\,|\,$7{\cdot} 10^{5}$ & \ensuremath{1.35}\,|\,$140$ & \ensuremath{${<}\,$ 1}\,|\,$75$ & \ensuremath{4.91}\,|\,$6218$ & \ensuremath{${<}\,$ 1}\,|\,$1229$ & \ensuremath{${<}\,$ 1}\,|\,$1227$ & 6\,|\,$5{\cdot} 10^{5}$ & 7\,|\,$4{\cdot} 10^{6}$\\
\hline
\model{Maze2} & \multirow{2}{*}{$R_\mathrm{min}$} & $15$ & \multirow{2}{*}{$8$} & \multirow{2}{*}{7.25} & ${\le}\, 8.13$ & $[7.99, 8.13]$ & ${\ge}\, 7.99$ & $[\mathbf{8.13}, 8.13]$ & ${\ge}\, \mathbf{8.13}$ & ${\ge}\, \mathbf{8.13}$ & ${\ge}\, \mathbf{8.13}$ & ${=}\, \mathbf{8.13}$*\\
0.3 &  & $54$ &  &  & 14\,|\,$7{\cdot} 10^{5}$ & \ensuremath{1.59}\,|\,$140$ & \ensuremath{${<}\,$ 1}\,|\,$90$ & \ensuremath{6.6}\,|\,$6218$ & \ensuremath{${<}\,$ 1}\,|\,$2103$ & \ensuremath{${<}\,$ 1}\,|\,$2097$ & 6\,|\,$8{\cdot} 10^{5}$ & 8\,|\,$5{\cdot} 10^{6}$\\
\hline\hline
\model{Refuel} & \multirow{2}{*}{$P_\mathrm{max}$} & $208$ & \multirow{2}{*}{$50$} & \multirow{2}{*}{0.98} & ${\ge}\, 0.67$ & \multirow{2}{*}{TO} & ${\le}\, 0.71$ & \multirow{2}{*}{MO} & ${\le}\, 0.68$ & ${\le}\, 0.68$ & ${=}\, \mathbf{0.67}$* & ${=}\, \mathbf{0.67}$*\\
06 &  & $574$ &  &  & 10\,|\,$2{\cdot} 10^{6}$ &  & \ensuremath{${<}\,$ 1}\,|\,$5862$ &  & \ensuremath{2.62}\,|\,$2{\cdot} 10^{5}$ & \ensuremath{2.7}\,|\,$2{\cdot} 10^{5}$ & 11\,|\,$1{\cdot} 10^{4}$ & 11\,|\,$1{\cdot} 10^{4}$\\
\hline
\model{Refuel} & \multirow{2}{*}{$P_\mathrm{max}$} & $470$ & \multirow{2}{*}{$66$} & \multirow{2}{*}{0.99} & ${\ge}\, 0.45$ & \multirow{2}{*}{MO} & ${\le}\, 0.76$ & \multirow{2}{*}{MO} & \multirow{2}{*}{MO} & \multirow{2}{*}{MO} & ${\le}\, 0.75$ & ${\le}\, \mathbf{0.48}$$^\dagger$\\
08 &  & $1446$ &  &  & 7\,|\,$1{\cdot} 10^{6}$ &  & \ensuremath{7.6}\,|\,$1{\cdot} 10^{5}$ &  &  &  & 2\,|\,$1{\cdot} 10^{5}$ & 3\,|\,$1{\cdot} 10^{6}$\\
\hline
\model{Refuel} & \multirow{2}{*}{$P_\mathrm{max}$} & $892$ & \multirow{2}{*}{$84$} & \multirow{2}{*}{1.0} & ${\ge}\, 0.43$ & \multirow{2}{*}{MO} & ${\le}\, \mathbf{0.84}$ & \multirow{2}{*}{MO} & \multirow{2}{*}{MO} & \multirow{2}{*}{MO} & ${\le}\, 0.87$ & ${\le}\, 0.87$$^\dagger$\\
10 &  & $2894$ &  &  & 5\,|\,$1{\cdot} 10^{6}$ &  & \ensuremath{49.2}\,|\,$1{\cdot} 10^{6}$ &  &  &  & 1\,|\,$1{\cdot} 10^{5}$ & 2\,|\,$1{\cdot} 10^{6}$\\
\hline\hline
\model{Rocks} & \multirow{2}{*}{$R_\mathrm{min}$} & $3241$ & \multirow{2}{*}{$817$} & \multirow{2}{*}{11} & ${\le}\, 26.2$ & \multirow{2}{*}{TO} & ${\ge}\, \mathbf{14}$ & \multirow{2}{*}{MO} & ${\ge}\, \mathbf{14}$ & ${\ge}\, \mathbf{14}$ & ${=}\, \mathbf{14}$* & ${=}\, \mathbf{14}$*\\
08 &  & $2{\cdot} 10^{4}$ &  &  & 7\,|\,$8{\cdot} 10^{5}$ &  & \ensuremath{${<}\,$ 1}\,|\,$2{\cdot} 10^{4}$ &  & \ensuremath{23}\,|\,$6{\cdot} 10^{5}$ & \ensuremath{22}\,|\,$6{\cdot} 10^{5}$ & 8\,|\,$2{\cdot} 10^{4}$ & 8\,|\,$2{\cdot} 10^{4}$\\
\hline
\model{Rocks} & \multirow{2}{*}{$R_\mathrm{min}$} & $6553$ & \multirow{2}{*}{$1645$} & \multirow{2}{*}{16.5} & ${\le}\, 35.4$ & \multirow{2}{*}{TO} & ${\ge}\, 19.9$ & \multirow{2}{*}{MO} & ${\ge}\, \mathbf{20}$ & ${\ge}\, \mathbf{20}$ & ${=}\, \mathbf{20}$* & ${=}\, \mathbf{20}$*\\
12 &  & $3{\cdot} 10^{4}$ &  &  & 6\,|\,$8{\cdot} 10^{5}$ &  & \ensuremath{1.53}\,|\,$5{\cdot} 10^{4}$ &  & \ensuremath{57.7}\,|\,$1{\cdot} 10^{6}$ & \ensuremath{57}\,|\,$1{\cdot} 10^{6}$ & 10\,|\,$1{\cdot} 10^{5}$ & 10\,|\,$1{\cdot} 10^{5}$\\
\hline
\model{Rocks} & \multirow{2}{*}{$R_\mathrm{min}$} & $1{\cdot} 10^{4}$ & \multirow{2}{*}{$2761$} & \multirow{2}{*}{22} & ${\le}\, 44$ & \multirow{2}{*}{MO} & ${\ge}\, 25.6$ & \multirow{2}{*}{MO} & ${\ge}\, \mathbf{26}$ & ${\ge}\, \mathbf{26}$ & ${\ge}\, 25.8$ & ${\ge}\, 25.9$$^\dagger$\\
16 &  & $5{\cdot} 10^{4}$ &  &  & 5\,|\,$7{\cdot} 10^{5}$ &  & \ensuremath{3.06}\,|\,$8{\cdot} 10^{4}$ &  & \ensuremath{104}\,|\,$2{\cdot} 10^{6}$ & \ensuremath{91}\,|\,$2{\cdot} 10^{6}$ & 8\,|\,$5{\cdot} 10^{5}$ & 10\,|\,$3{\cdot} 10^{6}$\\
\hline
\end{tabular}

		}
	}
\end{table}
\begin{table}[p]
	\caption{Results for POMDPs with finite belief MDP using the dynamic triangulation approach.}
	\label{tab:findyn}
	\scriptsize{
		\adjustbox{{max width=\textwidth}}{%
			\begin{tabular}{|cc||rr||r|r||r|r|r|r|r||r|r|}
\hline
\multicolumn{2}{|c||}{Benchmark} & \multicolumn{2}{c||}{Data} & \multicolumn{1}{c|}{MDP} & \multicolumn{1}{c||}{$\beliefmdp{\pomdp}$} & \multicolumn{2}{c|}{$\resolution$=4} & \multicolumn{3}{c||}{$\resolution$=12} & \multicolumn{2}{c|}{refine}\\
\multicolumn{1}{|c}{Model} & \multicolumn{1}{c||}{$\phi$} & \multicolumn{1}{c}{$\states$/$\actions$} & \multicolumn{1}{c||}{$\observations$} & \multicolumn{2}{c||}{\tool{Storm}} & \multicolumn{1}{c|}{\tool{Prism}} & \multicolumn{1}{c|}{\tool{Storm}} & \multicolumn{1}{c|}{\tool{Prism}} & \multicolumn{4}{c|}{\tool{Storm}}\\
\multicolumn{1}{|c}{} & \multicolumn{1}{c||}{} & \multicolumn{1}{c}{} & \multicolumn{1}{c||}{} & \multicolumn{1}{c|}{} & \multicolumn{1}{c||}{} & \multicolumn{1}{c|}{} & \multicolumn{1}{c|}{$\rho_\mathit{gap}$=0} & \multicolumn{1}{c|}{} & \multicolumn{1}{c|}{$\rho_\mathit{gap}$=0} & \multicolumn{1}{c||}{$\rho_\mathit{gap}$=0.2} & \multicolumn{1}{c|}{t=60} & \multicolumn{1}{c|}{t=1800}\\
\hline\hline
\model{Crypt} & \multirow{2}{*}{$P_\mathrm{max}$} & $1972$ & \multirow{2}{*}{$510$} & \multirow{2}{*}{1} & ${=}\, \mathbf{0.33}$ & $[0.33, 0.79]$ & ${\le}\, 0.67$ & \multirow{2}{*}{MO} & ${\le}\, \mathbf{0.33}$ & ${\le}\, \mathbf{0.33}$ & ${=}\, \mathbf{0.33}$* & ${=}\, \mathbf{0.33}$*\\
4 &  & $4612$ &  &  & \ensuremath{3.51}\,|\,$912$ & \ensuremath{20.3}\,|\,$5{\cdot} 10^{4}$ & \ensuremath{6.12}\,|\,$1044$ &  & \ensuremath{1.35}\,|\,$464$ & \ensuremath{1.23}\,|\,$464$ & 7\,|\,$556$ & 7\,|\,$556$\\
\hline
\model{Crypt} & \multirow{2}{*}{$P_\mathrm{min}$} & $1972$ & \multirow{2}{*}{$510$} & \multirow{2}{*}{0} & ${=}\, \mathbf{0.33}$ & $[0, 0.33]$ & ${\ge}\, 0.08$ & \multirow{2}{*}{MO} & ${\ge}\, \mathbf{0.33}$ & ${\ge}\, \mathbf{0.33}$ & ${=}\, \mathbf{0.33}$* & ${=}\, \mathbf{0.33}$*\\
4 &  & $4612$ &  &  & \ensuremath{${<}\,$ 1}\,|\,$912$ & \ensuremath{17.6}\,|\,$5{\cdot} 10^{4}$ & \ensuremath{2.85}\,|\,$1044$ &  & \ensuremath{7.75}\,|\,$464$ & \ensuremath{2.69}\,|\,$464$ & 7\,|\,$564$ & 7\,|\,$564$\\
\hline
\model{Crypt} & \multirow{2}{*}{$P_\mathrm{max}$} & $7{\cdot} 10^{4}$ & \multirow{2}{*}{$6678$} & \multirow{2}{*}{1} & ${=}\, \mathbf{0.2}$ & \multirow{2}{*}{MO} & ${\le}\, 1$ & \multirow{2}{*}{MO} & ${\le}\, 0.72$ & ${\le}\, 0.72$ & ${\le}\, 0.96$ & ${=}\, \mathbf{0.2}$*\\
6 &  & $2{\cdot} 10^{5}$ &  &  & \ensuremath{8.47}\,|\,$2{\cdot} 10^{4}$ &  & \ensuremath{13.8}\,|\,$8{\cdot} 10^{4}$ &  & \ensuremath{8.18}\,|\,$7{\cdot} 10^{4}$ & \ensuremath{12.9}\,|\,$7{\cdot} 10^{4}$ & 3\,|\,$2{\cdot} 10^{5}$ & 17\,|\,$2{\cdot} 10^{4}$\\
\hline
\model{Crypt} & \multirow{2}{*}{$P_\mathrm{min}$} & $7{\cdot} 10^{4}$ & \multirow{2}{*}{$6678$} & \multirow{2}{*}{0} & ${=}\, \mathbf{0.2}$ & \multirow{2}{*}{MO} & ${\ge}\, 0$ & \multirow{2}{*}{MO} & ${\ge}\, 0$ & ${\ge}\, 0$ & ${\ge}\, 0$ & ${=}\, \mathbf{0.2}$*\\
6 &  & $2{\cdot} 10^{5}$ &  &  & \ensuremath{6.59}\,|\,$2{\cdot} 10^{4}$ &  & \ensuremath{8.97}\,|\,$8{\cdot} 10^{4}$ &  & \ensuremath{9.43}\,|\,$7{\cdot} 10^{4}$ & \ensuremath{11.1}\,|\,$7{\cdot} 10^{4}$ & 3\,|\,$2{\cdot} 10^{5}$ & 27\,|\,$6360$\\
\hline\hline
\model{Grid-av} & \multirow{2}{*}{$P_\mathrm{max}$} & $17$ & \multirow{2}{*}{$4$} & \multirow{2}{*}{1} & ${=}\, \mathbf{0.93}$ & $[0.21, 1.0]$ & ${\le}\, 1$ & \multirow{2}{*}{MO} & ${\le}\, 0.94$ & ${\le}\, 0.94$ & ${=}\, \mathbf{0.93}$* & ${=}\, \mathbf{0.93}$*\\
4-0 &  & $59$ &  &  & \ensuremath{${<}\,$ 1}\,|\,$9016$ & \ensuremath{1.51}\,|\,$2382$ & \ensuremath{${<}\,$ 1}\,|\,$272$ &  & \ensuremath{${<}\,$ 1}\,|\,$4943$ & \ensuremath{${<}\,$ 1}\,|\,$4943$ & 6\,|\,$2976$ & 6\,|\,$2976$\\
\hline\hline
\model{Grid} & \multirow{2}{*}{$R_\mathrm{min}$} & $17$ & \multirow{2}{*}{$3$} & \multirow{2}{*}{3.2} & ${=}\, \mathbf{4.13}$ & $[3.6, 4.13]$ & ${\ge}\, 3.63$ & \multirow{2}{*}{MO} & ${\ge}\, 4.08$ & ${\ge}\, 4.08$ & ${=}\, \mathbf{4.13}$* & ${=}\, \mathbf{4.13}$*\\
4-0 &  & $62$ &  &  & \ensuremath{${<}\,$ 1}\,|\,$2423$ & \ensuremath{1.67}\,|\,$3061$ & \ensuremath{${<}\,$ 1}\,|\,$517$ &  & \ensuremath{${<}\,$ 1}\,|\,$2880$ & \ensuremath{${<}\,$ 1}\,|\,$2880$ & 5\,|\,$891$ & 5\,|\,$891$\\
\hline\hline
\model{Maze2} & \multirow{2}{*}{$R_\mathrm{min}$} & $15$ & \multirow{2}{*}{$8$} & \multirow{2}{*}{5.08} & ${=}\, \mathbf{5.69}$ & $[\mathbf{5.69}, 5.69]$ & ${\ge}\, \mathbf{5.69}$ & $[\mathbf{5.69}, 5.69]$ & ${\ge}\, \mathbf{5.69}$ & ${\ge}\, \mathbf{5.69}$ & ${=}\, \mathbf{5.69}$* & ${=}\, \mathbf{5.69}$*\\
0 &  & $54$ &  &  & \ensuremath{${<}\,$ 1}\,|\,$26$ & \ensuremath{1.43}\,|\,$140$ & \ensuremath{${<}\,$ 1}\,|\,$27$ & \ensuremath{3.17}\,|\,$6218$ & \ensuremath{${<}\,$ 1}\,|\,$23$ & \ensuremath{${<}\,$ 1}\,|\,$21$ & 8\,|\,$23$ & 8\,|\,$23$\\
\hline\hline
\model{Netw-p} & \multirow{2}{*}{$R_\mathrm{max}$} & $2{\cdot} 10^{4}$ & \multirow{2}{*}{$4909$} & \multirow{2}{*}{566} & ${=}\, \mathbf{557}$ & $[557, 559]$ & ${\le}\, 560$ & \multirow{2}{*}{TO} & ${\le}\, \mathbf{557}$ & ${\le}\, 566$ & ${\le}\, \mathbf{557}$ & ${\le}\, \mathbf{557}$\\
2-8-20 &  & $3{\cdot} 10^{4}$ &  &  & \ensuremath{612}\,|\,$3{\cdot} 10^{7}$ & \ensuremath{503}\,|\,$2{\cdot} 10^{5}$ & \ensuremath{2.15}\,|\,$8{\cdot} 10^{4}$ &  & \ensuremath{5.9}\,|\,$2{\cdot} 10^{5}$ & \ensuremath{${<}\,$ 1}\,|\,$2$ & 10\,|\,$1{\cdot} 10^{6}$ & 14\,|\,$2{\cdot} 10^{7}$\\
\hline
\model{Netw-p} & \multirow{2}{*}{$R_\mathrm{max}$} & $8019$ & \multirow{2}{*}{$1035$} & \multirow{2}{*}{73.6} & ${=}\, \mathbf{64.3}$ & $[64.1, 67.4]$ & ${\le}\, 69$ & \multirow{2}{*}{MO} & ${\le}\, 65$ & ${\le}\, 65.9$ & ${\le}\, 65.2$ & ${\le}\, \mathbf{64.3}$$^\dagger$\\
3-5-2 &  & $2{\cdot} 10^{4}$ &  &  & \ensuremath{71.6}\,|\,$7{\cdot} 10^{6}$ & \ensuremath{300}\,|\,$3{\cdot} 10^{5}$ & \ensuremath{3.19}\,|\,$7{\cdot} 10^{4}$ &  & \ensuremath{19.6}\,|\,$4{\cdot} 10^{5}$ & \ensuremath{18.3}\,|\,$3{\cdot} 10^{5}$ & 3\,|\,$4{\cdot} 10^{5}$ & 6\,|\,$1{\cdot} 10^{7}$\\
\hline
\model{Netw-p} & \multirow{2}{*}{$R_\mathrm{max}$} & $2{\cdot} 10^{5}$ & \multirow{2}{*}{$2{\cdot} 10^{4}$} & \multirow{2}{*}{849} & \multirow{2}{*}{TO} & \multirow{2}{*}{TO} & ${\le}\, 832$ & \multirow{2}{*}{MO} & \multirow{2}{*}{TO} & ${\le}\, 849$ & ${\le}\, 849$ & ${\le}\, \mathbf{826}$\\
3-8-20 &  & $3{\cdot} 10^{5}$ &  &  &  &  & \ensuremath{478}\,|\,$2{\cdot} 10^{6}$ &  &  & \ensuremath{8.3}\,|\,$2$ & 0\,|\,~--~ & 2\,|\,$2{\cdot} 10^{6}$\\
\hline\hline
\model{Netw} & \multirow{2}{*}{$R_\mathrm{min}$} & $4589$ & \multirow{2}{*}{$1173$} & \multirow{2}{*}{2.56} & ${=}\, \mathbf{3.2}$ & $[3.03, 3.2]$ & ${\ge}\, 2.97$ & $[3.17, 3.2]$ & ${\ge}\, 3.17$ & ${\ge}\, 3.16$ & ${\ge}\, \mathbf{3.2}$ & ${\ge}\, \mathbf{3.2}$\\
2-8-20 &  & $6973$ &  &  & \ensuremath{38.4}\,|\,$4{\cdot} 10^{6}$ & \ensuremath{42.1}\,|\,$4{\cdot} 10^{4}$ & \ensuremath{${<}\,$ 1}\,|\,$2{\cdot} 10^{4}$ & \ensuremath{521}\,|\,$5{\cdot} 10^{5}$ & \ensuremath{${<}\,$ 1}\,|\,$5{\cdot} 10^{4}$ & \ensuremath{${<}\,$ 1}\,|\,$5{\cdot} 10^{4}$ & 8\,|\,$2{\cdot} 10^{6}$ & 11\,|\,$1{\cdot} 10^{7}$\\
\hline
\model{Netw} & \multirow{2}{*}{$R_\mathrm{min}$} & $843$ & \multirow{2}{*}{$111$} & \multirow{2}{*}{1.18} & ${=}\, \mathbf{1.9}$ & $[1.64, 1.92]$ & ${\ge}\, 1.53$ & $[1.85, 1.9]$ & ${\ge}\, 1.84$ & ${\ge}\, 1.78$ & ${\ge}\, \mathbf{1.9}$ & ${\ge}\, \mathbf{1.9}$\\
3-5-2 &  & $1515$ &  &  & \ensuremath{3.12}\,|\,$4{\cdot} 10^{5}$ & \ensuremath{5.65}\,|\,$3{\cdot} 10^{4}$ & \ensuremath{${<}\,$ 1}\,|\,$6970$ & \ensuremath{835}\,|\,$5{\cdot} 10^{6}$ & \ensuremath{${<}\,$ 1}\,|\,$3{\cdot} 10^{4}$ & \ensuremath{${<}\,$ 1}\,|\,$3{\cdot} 10^{4}$ & 7\,|\,$2{\cdot} 10^{6}$ & 12\,|\,$2{\cdot} 10^{7}$\\
\hline
\model{Netw} & \multirow{2}{*}{$R_\mathrm{min}$} & $2{\cdot} 10^{4}$ & \multirow{2}{*}{$2205$} & \multirow{2}{*}{3.88} & \multirow{2}{*}{MO} & $[5.54, 6.77]$ & ${\ge}\, 5.11$ & \multirow{2}{*}{MO} & ${\ge}\, 6.37$ & ${\ge}\, 6.35$ & ${\ge}\, 6.17$ & ${\ge}\, \mathbf{6.73}$$^\dagger$\\
3-8-20 &  & $3{\cdot} 10^{4}$ &  &  &  & \ensuremath{1777}\,|\,$7{\cdot} 10^{5}$ & \ensuremath{4.42}\,|\,$2{\cdot} 10^{5}$ &  & \ensuremath{31.8}\,|\,$1{\cdot} 10^{6}$ & \ensuremath{32.6}\,|\,$1{\cdot} 10^{6}$ & 3\,|\,$1{\cdot} 10^{6}$ & 5\,|\,$1{\cdot} 10^{7}$\\
\hline\hline
\model{Nrp} & \multirow{2}{*}{$P_\mathrm{max}$} & $125$ & \multirow{2}{*}{$41$} & \multirow{2}{*}{1} & ${=}\, \mathbf{0.12}$ & $[0.13, 0.38]$ & ${\le}\, 0.25$ & $[0.13, 0.22]$ & ${\le}\, \mathbf{0.12}$ & ${\le}\, \mathbf{0.12}$ & ${=}\, \mathbf{0.12}$* & ${=}\, \mathbf{0.12}$*\\
8 &  & $161$ &  &  & \ensuremath{${<}\,$ 1}\,|\,$52$ & \ensuremath{1.57}\,|\,$2385$ & \ensuremath{${<}\,$ 1}\,|\,$53$ & \ensuremath{22.9}\,|\,$2{\cdot} 10^{5}$ & \ensuremath{${<}\,$ 1}\,|\,$32$ & \ensuremath{${<}\,$ 1}\,|\,$32$ & 6\,|\,$45$ & 6\,|\,$45$\\
\hline
\end{tabular}

		}
	}
\end{table}

\subsection{Comparison of different heuristic parameters}
Finally, we evaluated our refinement heuristic under different parameters in \Cref{tab:infpars,tab:finpars}.
We report on the best results that the refinement loop produces within 1800 seconds (as in the last column of the previous tables).
We compare the static and the dynamic approach for triangulation as well as 6 heuristics $h_i$.

\noindent
$h_0$ refers to the heuristic parameters as described in \Cref{sec:algorithm} and \Cref{app:whichobs}, i.e.:
\begin{itemize}
\item The triangulation resolutions are initialised with $\resolution[\mathit{init}]=3$ and iteratively increased by factor $f_{\resolution} = 2$.
\item The threshold for the score of refined observations is initially set to $\rho_{\observations} = 0.1$ and $f_{\observations} \cdot (1-\rho_{\observations})$ is added for each refinement step with $f_\observations = 0.1$.
\item The number of allowed exploration steps is initially unlimited and then set to $\rho_\mathit{step}=f_\mathit{step} \cdot |\abstbeliefstates|$ with $f_\mathit{step} = 4$.
\item The maximal gap for cut-offs is initialised with $\rho_\mathit{gap} =0.1$ and iteratively decreased by factor $f_\mathit{gap} = 0.25$.
\item For exploration, only the reachable fragment of the approximation under a $\rho_{\scheds{}} = 0.001$-optimal policy is considered.
\end{itemize}

We obtained the other heuristic parameters $h_1, \dots, h_5$ from $h_0$ as follows\footnote{All unmentioned parameters are as $h_0$.}:
\begin{itemize}
	\item For $h_1$ we set $f_{\resolution} = 1.4142135624 \approx \sqrt{2}$.
	\item For $h_2$ we set $f_\observations = 0.05$.
	\item For $h_3$ we set $f_\mathit{step} = 2$.
	\item For $h_4$ we set $f_\mathit{gap} = 0.5$.
	\item For $h_5$ we set $\rho_{\scheds{}} = 0.5$.
\end{itemize}
We observe that the different refinement heuristics often yield similar results, suggesting that the influence of the refinement parameters is limited.
A more extensive analysis of different strategies for refinement is left for future work.

\begin{table}[p]
	\caption{Comparison of different heuristic parameters for POMDPs with infinite belief MDP.}
	\label{tab:infpars}
	\scriptsize{
		\adjustbox{{max width=\textwidth}}{%
			\begin{tabular}{|cc||rr||r|r|r|r|r|r||r|r|r|r|r|r|}
\hline
\multicolumn{2}{|c||}{Benchmark} & \multicolumn{2}{c||}{Data} & \multicolumn{12}{c|}{refine / \tool{Storm} / t=1800}\\
\multicolumn{1}{|c}{Model} & \multicolumn{1}{c||}{$\phi$} & \multicolumn{1}{c}{$\states$/$\actions$} & \multicolumn{1}{c||}{$\observations$} & \multicolumn{6}{c||}{dynamic triangulation} & \multicolumn{6}{c|}{static triangulation}\\
\multicolumn{1}{|c}{} & \multicolumn{1}{c||}{} & \multicolumn{1}{c}{} & \multicolumn{1}{c||}{} & \multicolumn{1}{c|}{$h_0$} & \multicolumn{1}{c||}{$h_1$} & \multicolumn{1}{c|}{$h_2$} & \multicolumn{1}{c|}{$h_3$} & \multicolumn{1}{c|}{$h_4$} & \multicolumn{1}{c||}{$h_5$} & \multicolumn{1}{c|}{$h_0$} & \multicolumn{1}{c|}{$h_1$} & \multicolumn{1}{c|}{$h_2$} & \multicolumn{1}{c|}{$h_3$} & \multicolumn{1}{c|}{$h_4$} & \multicolumn{1}{c|}{$h_5$}\\
\hline\hline
\model{Drone} & \multirow{2}{*}{$P_\mathrm{max}$} & $1226$ & \multirow{2}{*}{$384$} & ${\le}\, 0.96$$^\dagger$ & ${\le}\, 0.96$$^\dagger$ & ${\le}\, 0.96$$^\dagger$ & ${\le}\, 0.96$$^\dagger$ & ${\le}\, 0.96$$^\dagger$ & ${\le}\, 0.96$$^\dagger$ & ${\le}\, 0.97$$^\dagger$ & ${\le}\, \mathbf{0.95}$$^\dagger$ & ${\le}\, 0.97$$^\dagger$ & ${\le}\, 0.97$$^\dagger$ & ${\le}\, 0.97$$^\dagger$ & ${\le}\, 0.97$$^\dagger$\\
4-1 &  & $2954$ &  & 3\,|\,$1{\cdot} 10^{6}$ & 5\,|\,$3{\cdot} 10^{6}$ & 3\,|\,$1{\cdot} 10^{6}$ & 4\,|\,$3{\cdot} 10^{6}$ & 3\,|\,$1{\cdot} 10^{6}$ & 3\,|\,$1{\cdot} 10^{6}$ & 3\,|\,$3{\cdot} 10^{6}$ & 5\,|\,$4{\cdot} 10^{6}$ & 3\,|\,$3{\cdot} 10^{6}$ & 4\,|\,$8{\cdot} 10^{6}$ & 3\,|\,$3{\cdot} 10^{6}$ & 3\,|\,$3{\cdot} 10^{6}$\\
\hline
\model{Drone} & \multirow{2}{*}{$P_\mathrm{max}$} & $1226$ & \multirow{2}{*}{$761$} & ${\le}\, 0.98$$^\dagger$ & ${\le}\, \mathbf{0.97}$$^\dagger$ & ${\le}\, 0.98$$^\dagger$ & ${\le}\, 0.98$$^\dagger$ & ${\le}\, 0.98$$^\dagger$ & ${\le}\, 0.98$$^\dagger$ & ${\le}\, \mathbf{0.97}$$^\dagger$ & ${\le}\, \mathbf{0.97}$$^\dagger$ & ${\le}\, \mathbf{0.97}$$^\dagger$ & ${\le}\, 0.98$$^\dagger$ & ${\le}\, 0.98$$^\dagger$ & ${\le}\, \mathbf{0.97}$$^\dagger$\\
4-2 &  & $2954$ &  & 4\,|\,$2{\cdot} 10^{6}$ & 6\,|\,$2{\cdot} 10^{6}$ & 4\,|\,$2{\cdot} 10^{6}$ & 5\,|\,$2{\cdot} 10^{6}$ & 4\,|\,$1{\cdot} 10^{6}$ & 4\,|\,$2{\cdot} 10^{6}$ & 4\,|\,$3{\cdot} 10^{6}$ & 6\,|\,$3{\cdot} 10^{6}$ & 4\,|\,$3{\cdot} 10^{6}$ & 5\,|\,$3{\cdot} 10^{6}$ & 4\,|\,$3{\cdot} 10^{6}$ & 4\,|\,$3{\cdot} 10^{6}$\\
\hline
\model{Drone} & \multirow{2}{*}{$P_\mathrm{max}$} & $2557$ & \multirow{2}{*}{$580$} & ${\le}\, \mathbf{0.98}$$^\dagger$ & ${\le}\, 0.99$$^\dagger$ & ${\le}\, \mathbf{0.98}$$^\dagger$ & ${\le}\, \mathbf{0.98}$$^\dagger$ & ${\le}\, \mathbf{0.98}$$^\dagger$ & ${\le}\, \mathbf{0.98}$$^\dagger$ & ${\le}\, 0.99$$^\dagger$ & ${\le}\, 0.99$$^\dagger$ & ${\le}\, 0.99$$^\dagger$ & ${\le}\, 0.99$$^\dagger$ & ${\le}\, 0.99$$^\dagger$ & ${\le}\, 0.99$$^\dagger$\\
5-1 &  & $6232$ &  & 3\,|\,$7{\cdot} 10^{6}$ & 4\,|\,$3{\cdot} 10^{6}$ & 3\,|\,$7{\cdot} 10^{6}$ & 3\,|\,$3{\cdot} 10^{6}$ & 3\,|\,$6{\cdot} 10^{6}$ & 3\,|\,$7{\cdot} 10^{6}$ & 2\,|\,$3{\cdot} 10^{6}$ & 4\,|\,$3{\cdot} 10^{6}$ & 2\,|\,$3{\cdot} 10^{6}$ & 3\,|\,$6{\cdot} 10^{6}$ & 2\,|\,$2{\cdot} 10^{6}$ & 2\,|\,$3{\cdot} 10^{6}$\\
\hline
\model{Drone} & \multirow{2}{*}{$P_\mathrm{max}$} & $2557$ & \multirow{2}{*}{$1848$} & ${\le}\, \mathbf{0.99}$$^\dagger$ & ${\le}\, \mathbf{0.99}$$^\dagger$ & ${\le}\, \mathbf{0.99}$$^\dagger$ & ${\le}\, \mathbf{0.99}$$^\dagger$ & ${\le}\, \mathbf{0.99}$$^\dagger$ & ${\le}\, \mathbf{0.99}$$^\dagger$ & ${\le}\, \mathbf{0.99}$$^\dagger$ & ${\le}\, \mathbf{0.99}$$^\dagger$ & ${\le}\, \mathbf{0.99}$$^\dagger$ & ${\le}\, \mathbf{0.99}$$^\dagger$ & ${\le}\, \mathbf{0.99}$$^\dagger$ & ${\le}\, \mathbf{0.99}$$^\dagger$\\
5-3 &  & $6232$ &  & 4\,|\,$3{\cdot} 10^{6}$ & 6\,|\,$8{\cdot} 10^{6}$ & 4\,|\,$3{\cdot} 10^{6}$ & 5\,|\,$5{\cdot} 10^{6}$ & 4\,|\,$3{\cdot} 10^{6}$ & 4\,|\,$3{\cdot} 10^{6}$ & 4\,|\,$8{\cdot} 10^{6}$ & 5\,|\,$2{\cdot} 10^{6}$ & 4\,|\,$8{\cdot} 10^{6}$ & 4\,|\,$2{\cdot} 10^{6}$ & 4\,|\,$7{\cdot} 10^{6}$ & 4\,|\,$8{\cdot} 10^{6}$\\
\hline\hline
\model{Grid-av} & \multirow{2}{*}{$P_\mathrm{max}$} & $17$ & \multirow{2}{*}{$4$} & ${\le}\, 0.94$$^\dagger$ & ${\le}\, 0.94$$^\dagger$ & ${\le}\, 0.94$$^\dagger$ & ${\le}\, \mathbf{0.93}$$^\dagger$ & ${\le}\, 0.94$$^\dagger$ & ${\le}\, 0.94$$^\dagger$ & ${\le}\, 0.94$$^\dagger$ & ${\le}\, 0.94$$^\dagger$ & ${\le}\, 0.94$$^\dagger$ & ${\le}\, 0.94$$^\dagger$ & ${\le}\, 0.94$$^\dagger$ & ${\le}\, 0.94$$^\dagger$\\
4-0.1 &  & $59$ &  & 3\,|\,$3{\cdot} 10^{5}$ & 6\,|\,$2{\cdot} 10^{6}$ & 3\,|\,$3{\cdot} 10^{5}$ & 4\,|\,$3{\cdot} 10^{6}$ & 3\,|\,$3{\cdot} 10^{5}$ & 3\,|\,$3{\cdot} 10^{5}$ & 3\,|\,$3{\cdot} 10^{5}$ & 6\,|\,$2{\cdot} 10^{6}$ & 3\,|\,$3{\cdot} 10^{5}$ & 4\,|\,$3{\cdot} 10^{6}$ & 3\,|\,$3{\cdot} 10^{5}$ & 3\,|\,$3{\cdot} 10^{5}$\\
\hline
\model{Grid-av} & \multirow{2}{*}{$P_\mathrm{max}$} & $17$ & \multirow{2}{*}{$4$} & ${\le}\, 0.95$$^\dagger$ & ${\le}\, 0.95$$^\dagger$ & ${\le}\, 0.95$$^\dagger$ & ${\le}\, \mathbf{0.93}$$^\dagger$ & ${\le}\, 0.95$$^\dagger$ & ${\le}\, 0.95$$^\dagger$ & ${\le}\, 0.95$$^\dagger$ & ${\le}\, 0.95$$^\dagger$ & ${\le}\, 0.95$$^\dagger$ & ${\le}\, 0.94$$^\dagger$ & ${\le}\, 0.95$$^\dagger$ & ${\le}\, 0.95$$^\dagger$\\
4-0.3 &  & $59$ &  & 3\,|\,$3{\cdot} 10^{5}$ & 6\,|\,$2{\cdot} 10^{6}$ & 3\,|\,$3{\cdot} 10^{5}$ & 4\,|\,$3{\cdot} 10^{6}$ & 3\,|\,$3{\cdot} 10^{5}$ & 3\,|\,$3{\cdot} 10^{5}$ & 3\,|\,$3{\cdot} 10^{5}$ & 6\,|\,$2{\cdot} 10^{6}$ & 3\,|\,$3{\cdot} 10^{5}$ & 4\,|\,$3{\cdot} 10^{6}$ & 3\,|\,$3{\cdot} 10^{5}$ & 3\,|\,$3{\cdot} 10^{5}$\\
\hline\hline
\model{Grid} & \multirow{2}{*}{$R_\mathrm{min}$} & $17$ & \multirow{2}{*}{$3$} & ${\ge}\, \mathbf{4.65}$$^\dagger$ & ${\ge}\, 4.59$$^\dagger$ & ${\ge}\, \mathbf{4.65}$$^\dagger$ & ${\ge}\, 4.59$$^\dagger$ & ${\ge}\, \mathbf{4.65}$$^\dagger$ & ${\ge}\, \mathbf{4.65}$$^\dagger$ & ${\ge}\, 4.61$$^\dagger$ & ${\ge}\, 4.59$$^\dagger$ & ${\ge}\, 4.61$$^\dagger$ & ${\ge}\, 4.57$$^\dagger$ & ${\ge}\, 4.61$$^\dagger$ & ${\ge}\, 4.6$$^\dagger$\\
4-0.1 &  & $62$ &  & 4\,|\,$6{\cdot} 10^{6}$ & 6\,|\,$2{\cdot} 10^{6}$ & 4\,|\,$6{\cdot} 10^{6}$ & 4\,|\,$2{\cdot} 10^{6}$ & 4\,|\,$6{\cdot} 10^{6}$ & 4\,|\,$6{\cdot} 10^{6}$ & 4\,|\,$4{\cdot} 10^{6}$ & 6\,|\,$1{\cdot} 10^{6}$ & 4\,|\,$4{\cdot} 10^{6}$ & 4\,|\,$1{\cdot} 10^{6}$ & 4\,|\,$4{\cdot} 10^{6}$ & 4\,|\,$4{\cdot} 10^{6}$\\
\hline
\model{Grid} & \multirow{2}{*}{$R_\mathrm{min}$} & $17$ & \multirow{2}{*}{$3$} & ${\ge}\, 5.9$$^\dagger$ & ${\ge}\, \mathbf{6.18}$$^\dagger$ & ${\ge}\, 5.9$$^\dagger$ & ${\ge}\, 5.85$$^\dagger$ & ${\ge}\, 5.9$$^\dagger$ & ${\ge}\, 5.9$$^\dagger$ & ${\ge}\, 5.92$$^\dagger$ & ${\ge}\, \mathbf{6.18}$$^\dagger$ & ${\ge}\, 5.92$$^\dagger$ & ${\ge}\, 5.85$$^\dagger$ & ${\ge}\, 5.92$$^\dagger$ & ${\ge}\, 5.92$$^\dagger$\\
4-0.3 &  & $62$ &  & 3\,|\,$2{\cdot} 10^{5}$ & 6\,|\,$2{\cdot} 10^{6}$ & 3\,|\,$2{\cdot} 10^{5}$ & 4\,|\,$2{\cdot} 10^{6}$ & 3\,|\,$2{\cdot} 10^{5}$ & 3\,|\,$2{\cdot} 10^{5}$ & 4\,|\,$4{\cdot} 10^{6}$ & 6\,|\,$1{\cdot} 10^{6}$ & 4\,|\,$4{\cdot} 10^{6}$ & 4\,|\,$2{\cdot} 10^{6}$ & 4\,|\,$4{\cdot} 10^{6}$ & 4\,|\,$4{\cdot} 10^{6}$\\
\hline\hline
\model{Maze2} & \multirow{2}{*}{$R_\mathrm{min}$} & $15$ & \multirow{2}{*}{$8$} & ${\ge}\, \mathbf{6.32}$$^\dagger$ & ${\ge}\, \mathbf{6.32}$$^\dagger$ & ${\ge}\, \mathbf{6.32}$$^\dagger$ & ${\ge}\, \mathbf{6.32}$ & ${\ge}\, \mathbf{6.32}$$^\dagger$ & ${\ge}\, \mathbf{6.32}$$^\dagger$ & ${\ge}\, \mathbf{6.32}$$^\dagger$ & ${\ge}\, \mathbf{6.32}$$^\dagger$ & ${\ge}\, \mathbf{6.32}$$^\dagger$ & ${\ge}\, \mathbf{6.32}$$^\dagger$ & ${\ge}\, \mathbf{6.32}$$^\dagger$ & ${\ge}\, \mathbf{6.32}$$^\dagger$\\
0.1 &  & $54$ &  & 7\,|\,$4{\cdot} 10^{6}$ & 14\,|\,$5{\cdot} 10^{6}$ & 7\,|\,$4{\cdot} 10^{6}$ & 7\,|\,$6{\cdot} 10^{5}$ & 7\,|\,$4{\cdot} 10^{6}$ & 7\,|\,$3{\cdot} 10^{6}$ & 8\,|\,$6{\cdot} 10^{6}$ & 17\,|\,$8{\cdot} 10^{6}$ & 8\,|\,$6{\cdot} 10^{6}$ & 9\,|\,$9{\cdot} 10^{6}$ & 8\,|\,$6{\cdot} 10^{6}$ & 8\,|\,$4{\cdot} 10^{6}$\\
\hline
\model{Maze2} & \multirow{2}{*}{$R_\mathrm{min}$} & $15$ & \multirow{2}{*}{$8$} & ${\ge}\, \mathbf{8.13}$ & ${\ge}\, \mathbf{8.13}$$^\dagger$ & ${\ge}\, \mathbf{8.13}$ & ${\ge}\, \mathbf{8.13}$ & ${\ge}\, \mathbf{8.13}$ & ${\ge}\, \mathbf{8.13}$ & ${\ge}\, \mathbf{8.13}$ & ${\ge}\, \mathbf{8.13}$$^\dagger$ & ${\ge}\, \mathbf{8.13}$ & ${\ge}\, \mathbf{8.13}$ & ${\ge}\, \mathbf{8.13}$ & ${\ge}\, \mathbf{8.13}$$^\dagger$\\
0.3 &  & $54$ &  & 6\,|\,$8{\cdot} 10^{5}$ & 12\,|\,$5{\cdot} 10^{6}$ & 6\,|\,$8{\cdot} 10^{5}$ & 7\,|\,$6{\cdot} 10^{5}$ & 6\,|\,$8{\cdot} 10^{5}$ & 5\,|\,$8{\cdot} 10^{4}$ & 5\,|\,$5{\cdot} 10^{4}$ & 14\,|\,$5{\cdot} 10^{6}$ & 5\,|\,$5{\cdot} 10^{4}$ & 7\,|\,$6{\cdot} 10^{5}$ & 5\,|\,$5{\cdot} 10^{4}$ & 8\,|\,$9{\cdot} 10^{6}$\\
\hline\hline
\model{Refuel} & \multirow{2}{*}{$P_\mathrm{max}$} & $208$ & \multirow{2}{*}{$50$} & ${=}\, \mathbf{0.67}$* & ${=}\, \mathbf{0.67}$* & ${=}\, \mathbf{0.67}$* & ${=}\, \mathbf{0.67}$* & ${=}\, \mathbf{0.67}$* & ${\le}\, \mathbf{0.67}$$^\dagger$ & ${=}\, \mathbf{0.67}$* & ${=}\, \mathbf{0.67}$* & ${=}\, \mathbf{0.67}$* & ${=}\, \mathbf{0.67}$* & ${=}\, \mathbf{0.67}$* & ${\le}\, \mathbf{0.67}$$^\dagger$\\
06 &  & $565$ &  & 11\,|\,$1{\cdot} 10^{4}$ & 18\,|\,$1{\cdot} 10^{4}$ & 14\,|\,$2{\cdot} 10^{4}$ & 11\,|\,$1{\cdot} 10^{4}$ & 11\,|\,$1{\cdot} 10^{4}$ & 7\,|\,$1{\cdot} 10^{7}$ & 59\,|\,$2{\cdot} 10^{4}$ & 113\,|\,$4911$ & 59\,|\,$2{\cdot} 10^{4}$ & 59\,|\,$2{\cdot} 10^{4}$ & 59\,|\,$2{\cdot} 10^{4}$ & 9\,|\,$2{\cdot} 10^{7}$\\
\hline
\model{Refuel} & \multirow{2}{*}{$P_\mathrm{max}$} & $470$ & \multirow{2}{*}{$66$} & ${\le}\, \mathbf{0.48}$$^\dagger$ & ${\le}\, 0.6$$^\dagger$ & ${\le}\, 0.61$$^\dagger$ & ${\le}\, 0.6$$^\dagger$ & ${\le}\, \mathbf{0.48}$$^\dagger$ & ${\le}\, \mathbf{0.48}$$^\dagger$ & ${\le}\, 0.58$$^\dagger$ & ${\le}\, 0.51$$^\dagger$ & ${\le}\, 0.58$$^\dagger$ & ${\le}\, 0.58$$^\dagger$ & ${\le}\, 0.58$$^\dagger$ & ${\le}\, \mathbf{0.48}$$^\dagger$\\
08 &  & $1431$ &  & 3\,|\,$1{\cdot} 10^{6}$ & 6\,|\,$3{\cdot} 10^{6}$ & 4\,|\,$2{\cdot} 10^{6}$ & 4\,|\,$3{\cdot} 10^{6}$ & 3\,|\,$1{\cdot} 10^{6}$ & 3\,|\,$3{\cdot} 10^{6}$ & 3\,|\,$1{\cdot} 10^{6}$ & 6\,|\,$5{\cdot} 10^{6}$ & 3\,|\,$1{\cdot} 10^{6}$ & 5\,|\,$5{\cdot} 10^{6}$ & 3\,|\,$1{\cdot} 10^{6}$ & 3\,|\,$4{\cdot} 10^{6}$\\
\hline
\model{Refuel} & \multirow{2}{*}{$P_\mathrm{max}$} & $892$ & \multirow{2}{*}{$84$} & ${\le}\, \mathbf{0.87}$$^\dagger$ & ${\le}\, \mathbf{0.87}$$^\dagger$ & ${\le}\, \mathbf{0.87}$$^\dagger$ & ${\le}\, \mathbf{0.87}$$^\dagger$ & ${\le}\, \mathbf{0.87}$$^\dagger$ & ${\le}\, \mathbf{0.87}$$^\dagger$ & ${\le}\, \mathbf{0.87}$$^\dagger$ & ${\le}\, \mathbf{0.87}$$^\dagger$ & ${\le}\, \mathbf{0.87}$$^\dagger$ & ${\le}\, \mathbf{0.87}$$^\dagger$ & ${\le}\, \mathbf{0.87}$$^\dagger$ & ${\le}\, \mathbf{0.87}$$^\dagger$\\
10 &  & $2879$ &  & 2\,|\,$1{\cdot} 10^{6}$ & 6\,|\,$2{\cdot} 10^{7}$ & 2\,|\,$1{\cdot} 10^{6}$ & 2\,|\,$1{\cdot} 10^{6}$ & 2\,|\,$1{\cdot} 10^{6}$ & 2\,|\,$1{\cdot} 10^{6}$ & 2\,|\,$1{\cdot} 10^{6}$ & 5\,|\,$3{\cdot} 10^{6}$ & 2\,|\,$1{\cdot} 10^{6}$ & 2\,|\,$1{\cdot} 10^{6}$ & 2\,|\,$1{\cdot} 10^{6}$ & 2\,|\,$1{\cdot} 10^{6}$\\
\hline\hline
\model{Rocks} & \multirow{2}{*}{$R_\mathrm{min}$} & $3241$ & \multirow{2}{*}{$817$} & ${=}\, \mathbf{14}$* & ${=}\, \mathbf{14}$* & ${=}\, \mathbf{14}$* & ${=}\, \mathbf{14}$* & ${=}\, \mathbf{14}$* & ${\ge}\, \mathbf{14}$ & ${=}\, \mathbf{14}$* & ${\ge}\, \mathbf{14}$$^\dagger$ & ${=}\, \mathbf{14}$* & ${=}\, \mathbf{14}$* & ${=}\, \mathbf{14}$* & ${\ge}\, \mathbf{14}$$^\dagger$\\
08 &  & $1{\cdot} 10^{4}$ &  & 8\,|\,$2{\cdot} 10^{4}$ & 22\,|\,$1{\cdot} 10^{5}$ & 14\,|\,$2{\cdot} 10^{4}$ & 8\,|\,$2{\cdot} 10^{4}$ & 8\,|\,$2{\cdot} 10^{4}$ & 4\,|\,$1{\cdot} 10^{6}$ & 9\,|\,$2{\cdot} 10^{4}$ & 18\,|\,$6{\cdot} 10^{6}$ & 14\,|\,$1{\cdot} 10^{4}$ & 9\,|\,$2{\cdot} 10^{4}$ & 9\,|\,$2{\cdot} 10^{4}$ & 5\,|\,$4{\cdot} 10^{6}$\\
\hline
\model{Rocks} & \multirow{2}{*}{$R_\mathrm{min}$} & $6553$ & \multirow{2}{*}{$1645$} & ${=}\, \mathbf{20}$* & ${\ge}\, \mathbf{20}$ & ${=}\, \mathbf{20}$* & ${=}\, \mathbf{20}$* & ${=}\, \mathbf{20}$* & ${\ge}\, \mathbf{20}$$^\dagger$ & ${=}\, \mathbf{20}$* & ${\ge}\, \mathbf{20}$$^\dagger$ & ${=}\, \mathbf{20}$* & ${=}\, \mathbf{20}$* & ${=}\, \mathbf{20}$* & ${\ge}\, \mathbf{20}$$^\dagger$\\
12 &  & $3{\cdot} 10^{4}$ &  & 10\,|\,$1{\cdot} 10^{5}$ & 18\,|\,$4{\cdot} 10^{6}$ & 14\,|\,$4{\cdot} 10^{4}$ & 10\,|\,$1{\cdot} 10^{5}$ & 10\,|\,$1{\cdot} 10^{5}$ & 4\,|\,$3{\cdot} 10^{6}$ & 9\,|\,$5{\cdot} 10^{4}$ & 17\,|\,$6{\cdot} 10^{6}$ & 14\,|\,$4{\cdot} 10^{4}$ & 9\,|\,$5{\cdot} 10^{4}$ & 9\,|\,$5{\cdot} 10^{4}$ & 5\,|\,$1{\cdot} 10^{7}$\\
\hline
\model{Rocks} & \multirow{2}{*}{$R_\mathrm{min}$} & $1{\cdot} 10^{4}$ & \multirow{2}{*}{$2761$} & ${\ge}\, 25.9$$^\dagger$ & ${\ge}\, 25.6$$^\dagger$ & ${\ge}\, 25.9$ & ${\ge}\, 25.8$ & ${\ge}\, 25.7$$^\dagger$ & ${\ge}\, \mathbf{26}$$^\dagger$ & ${\ge}\, 25.9$ & ${\ge}\, \mathbf{26}$$^\dagger$ & ${=}\, \mathbf{26}$* & ${\ge}\, \mathbf{26}$$^\dagger$ & ${\ge}\, \mathbf{26}$ & ${\ge}\, \mathbf{26}$$^\dagger$\\
16 &  & $5{\cdot} 10^{4}$ &  & 10\,|\,$3{\cdot} 10^{6}$ & 19\,|\,$3{\cdot} 10^{6}$ & 11\,|\,$9{\cdot} 10^{6}$ & 9\,|\,$2{\cdot} 10^{6}$ & 8\,|\,$1{\cdot} 10^{6}$ & 4\,|\,$5{\cdot} 10^{6}$ & 9\,|\,$2{\cdot} 10^{6}$ & 17\,|\,$5{\cdot} 10^{6}$ & 18\,|\,$2{\cdot} 10^{6}$ & 10\,|\,$3{\cdot} 10^{6}$ & 9\,|\,$3{\cdot} 10^{6}$ & 4\,|\,$3{\cdot} 10^{6}$\\
\hline
\end{tabular}

		}
	}
\end{table}
\begin{table}[p]
	\caption{Comparison of different heuristic parameters for POMDPs with finite belief MDP.}
	\label{tab:finpars}
	\scriptsize{
		\adjustbox{{max width=\textwidth}}{%
			\begin{tabular}{|cc||rr||r|r|r|r|r|r||r|r|r|r|r|r|}
\hline
\multicolumn{2}{|c||}{Benchmark} & \multicolumn{2}{c||}{Data} & \multicolumn{12}{c|}{refine / \tool{Storm} / t=1800}\\
\multicolumn{1}{|c}{Model} & \multicolumn{1}{c||}{$\phi$} & \multicolumn{1}{c}{$\states$/$\actions$} & \multicolumn{1}{c||}{$\observations$} & \multicolumn{6}{c||}{dynamic triangulation} & \multicolumn{6}{c|}{static triangulation}\\
\multicolumn{1}{|c}{} & \multicolumn{1}{c||}{} & \multicolumn{1}{c}{} & \multicolumn{1}{c||}{} & \multicolumn{1}{c|}{$h_0$} & \multicolumn{1}{c||}{$h_1$} & \multicolumn{1}{c|}{$h_2$} & \multicolumn{1}{c|}{$h_3$} & \multicolumn{1}{c|}{$h_4$} & \multicolumn{1}{c||}{$h_5$} & \multicolumn{1}{c|}{$h_0$} & \multicolumn{1}{c|}{$h_1$} & \multicolumn{1}{c|}{$h_2$} & \multicolumn{1}{c|}{$h_3$} & \multicolumn{1}{c|}{$h_4$} & \multicolumn{1}{c|}{$h_5$}\\
\hline\hline
\model{Crypt} & \multirow{2}{*}{$P_\mathrm{max}$} & $1972$ & \multirow{2}{*}{$510$} & ${=}\, \mathbf{0.33}$* & ${=}\, \mathbf{0.33}$* & ${=}\, \mathbf{0.33}$* & ${=}\, \mathbf{0.33}$* & ${=}\, \mathbf{0.33}$* & ${=}\, \mathbf{0.33}$* & ${=}\, \mathbf{0.33}$* & ${=}\, \mathbf{0.33}$* & ${=}\, \mathbf{0.33}$* & ${=}\, \mathbf{0.33}$* & ${=}\, \mathbf{0.33}$* & ${=}\, \mathbf{0.33}$*\\
4 &  & $4612$ &  & 7\,|\,$556$ & 26\,|\,$680$ & 12\,|\,$556$ & 7\,|\,$556$ & 7\,|\,$556$ & 7\,|\,$508$ & 6\,|\,$564$ & 165\,|\,$928$ & 9\,|\,$548$ & 6\,|\,$564$ & 6\,|\,$564$ & 6\,|\,$500$\\
\hline
\model{Crypt} & \multirow{2}{*}{$P_\mathrm{min}$} & $1972$ & \multirow{2}{*}{$510$} & ${=}\, \mathbf{0.33}$* & ${=}\, \mathbf{0.33}$* & ${=}\, \mathbf{0.33}$* & ${=}\, \mathbf{0.33}$* & ${=}\, \mathbf{0.33}$* & ${=}\, \mathbf{0.33}$* & ${=}\, \mathbf{0.33}$* & ${=}\, \mathbf{0.33}$* & ${=}\, \mathbf{0.33}$* & ${=}\, \mathbf{0.33}$* & ${=}\, \mathbf{0.33}$* & ${=}\, \mathbf{0.33}$*\\
4 &  & $4612$ &  & 7\,|\,$564$ & 9\,|\,$980$ & 12\,|\,$564$ & 7\,|\,$564$ & 7\,|\,$564$ & 7\,|\,$564$ & 32\,|\,$2000$ & 153\,|\,$920$ & 62\,|\,$2148$ & 32\,|\,$1886$ & 52\,|\,$2000$ & 32\,|\,$2000$\\
\hline
\model{Crypt} & \multirow{2}{*}{$P_\mathrm{max}$} & $7{\cdot} 10^{4}$ & \multirow{2}{*}{$6678$} & ${=}\, \mathbf{0.2}$* & ${=}\, \mathbf{0.2}$* & ${=}\, \mathbf{0.2}$* & ${=}\, \mathbf{0.2}$* & ${=}\, \mathbf{0.2}$* & ${=}\, \mathbf{0.2}$* & ${\le}\, 0.94$ & ${\le}\, 0.88$ & ${\le}\, 0.96$ & ${\le}\, 1$$^\dagger$ & ${\le}\, 0.94$ & ${\le}\, 0.94$\\
6 &  & $2{\cdot} 10^{5}$ &  & 17\,|\,$2{\cdot} 10^{4}$ & 25\,|\,$1{\cdot} 10^{4}$ & 24\,|\,$2{\cdot} 10^{4}$ & 17\,|\,$2{\cdot} 10^{4}$ & 17\,|\,$2{\cdot} 10^{4}$ & 7\,|\,$7188$ & 4\,|\,$7{\cdot} 10^{6}$ & 7\,|\,$5{\cdot} 10^{6}$ & 4\,|\,$5{\cdot} 10^{6}$ & 4\,|\,$5{\cdot} 10^{6}$ & 4\,|\,$7{\cdot} 10^{6}$ & 4\,|\,$1{\cdot} 10^{7}$\\
\hline
\model{Crypt} & \multirow{2}{*}{$P_\mathrm{min}$} & $7{\cdot} 10^{4}$ & \multirow{2}{*}{$6678$} & ${=}\, \mathbf{0.2}$* & ${\ge}\, 2{\cdot} 10^{-3}$ & ${=}\, \mathbf{0.2}$* & ${=}\, \mathbf{0.2}$* & ${\ge}\, \mathbf{0.2}$ & ${=}\, \mathbf{0.2}$* & ${\ge}\, 0$ & ${\ge}\, 0$ & ${\ge}\, 0$ & ${\ge}\, 0$ & ${\ge}\, 0$ & ${\ge}\, 0$\\
6 &  & $2{\cdot} 10^{5}$ &  & 27\,|\,$6360$ & 9\,|\,$4{\cdot} 10^{6}$ & 27\,|\,$6360$ & 27\,|\,$6360$ & 30\,|\,$6360$ & 27\,|\,$6360$ & 4\,|\,$1{\cdot} 10^{7}$ & 7\,|\,$8{\cdot} 10^{6}$ & 4\,|\,$1{\cdot} 10^{7}$ & 4\,|\,$8{\cdot} 10^{6}$ & 4\,|\,$1{\cdot} 10^{7}$ & 4\,|\,$1{\cdot} 10^{7}$\\
\hline\hline
\model{Grid-av} & \multirow{2}{*}{$P_\mathrm{max}$} & $17$ & \multirow{2}{*}{$4$} & ${=}\, \mathbf{0.93}$* & ${=}\, \mathbf{0.93}$* & ${=}\, \mathbf{0.93}$* & ${=}\, \mathbf{0.93}$* & ${=}\, \mathbf{0.93}$* & ${=}\, \mathbf{0.93}$* & ${\le}\, \mathbf{0.93}$$^\dagger$ & ${\le}\, \mathbf{0.93}$$^\dagger$ & ${\le}\, \mathbf{0.93}$$^\dagger$ & ${\le}\, \mathbf{0.93}$$^\dagger$ & ${\le}\, \mathbf{0.93}$$^\dagger$ & ${\le}\, \mathbf{0.93}$$^\dagger$\\
4-0 &  & $59$ &  & 6\,|\,$2976$ & 9\,|\,$3016$ & 6\,|\,$2976$ & 6\,|\,$2411$ & 6\,|\,$2976$ & 6\,|\,$3016$ & 26\,|\,$1{\cdot} 10^{7}$ & 50\,|\,$3{\cdot} 10^{6}$ & 26\,|\,$1{\cdot} 10^{7}$ & 27\,|\,$9{\cdot} 10^{6}$ & 26\,|\,$1{\cdot} 10^{7}$ & 26\,|\,$1{\cdot} 10^{7}$\\
\hline\hline
\model{Grid} & \multirow{2}{*}{$R_\mathrm{min}$} & $17$ & \multirow{2}{*}{$3$} & ${=}\, \mathbf{4.13}$* & ${=}\, \mathbf{4.13}$* & ${=}\, \mathbf{4.13}$* & ${=}\, \mathbf{4.13}$* & ${=}\, \mathbf{4.13}$* & ${=}\, \mathbf{4.13}$* & ${\ge}\, \mathbf{4.13}$ & ${\ge}\, \mathbf{4.13}$ & ${\ge}\, \mathbf{4.13}$ & ${\ge}\, \mathbf{4.13}$ & ${\ge}\, \mathbf{4.13}$ & ${\ge}\, \mathbf{4.13}$\\
4-0 &  & $62$ &  & 5\,|\,$891$ & 9\,|\,$891$ & 5\,|\,$891$ & 5\,|\,$886$ & 5\,|\,$891$ & 6\,|\,$891$ & 26\,|\,$9{\cdot} 10^{5}$ & 49\,|\,$9{\cdot} 10^{5}$ & 26\,|\,$9{\cdot} 10^{5}$ & 28\,|\,$1{\cdot} 10^{6}$ & 26\,|\,$9{\cdot} 10^{5}$ & 26\,|\,$9{\cdot} 10^{5}$\\
\hline\hline
\model{Maze2} & \multirow{2}{*}{$R_\mathrm{min}$} & $15$ & \multirow{2}{*}{$8$} & ${=}\, \mathbf{5.69}$* & ${=}\, \mathbf{5.69}$* & ${=}\, \mathbf{5.69}$* & ${=}\, \mathbf{5.69}$* & ${=}\, \mathbf{5.69}$* & ${=}\, \mathbf{5.69}$* & ${=}\, \mathbf{5.69}$* & ${\ge}\, \mathbf{5.69}$$^\dagger$ & ${=}\, \mathbf{5.69}$* & ${=}\, \mathbf{5.69}$* & ${=}\, \mathbf{5.69}$* & ${=}\, \mathbf{5.69}$*\\
0 &  & $54$ &  & 8\,|\,$23$ & 13\,|\,$23$ & 14\,|\,$23$ & 8\,|\,$23$ & 8\,|\,$23$ & 8\,|\,$23$ & 4\,|\,$23$ & 51\,|\,$1{\cdot} 10^{7}$ & 4\,|\,$23$ & 4\,|\,$23$ & 5\,|\,$23$ & 3\,|\,$23$\\
\hline\hline
\model{Netw-p} & \multirow{2}{*}{$R_\mathrm{max}$} & $2{\cdot} 10^{4}$ & \multirow{2}{*}{$4909$} & ${\le}\, \mathbf{557}$ & ${\le}\, \mathbf{557}$ & ${\le}\, \mathbf{557}$ & ${\le}\, \mathbf{557}$ & ${\le}\, \mathbf{557}$ & ${\le}\, \mathbf{557}$ & ${\le}\, \mathbf{557}$ & ${\le}\, \mathbf{557}$ & ${\le}\, \mathbf{557}$ & ${\le}\, \mathbf{557}$ & ${\le}\, \mathbf{557}$ & ${\le}\, \mathbf{557}$\\
2-8-20 &  & $3{\cdot} 10^{4}$ &  & 14\,|\,$2{\cdot} 10^{7}$ & 21\,|\,$2{\cdot} 10^{7}$ & 14\,|\,$2{\cdot} 10^{7}$ & 17\,|\,$2{\cdot} 10^{7}$ & 16\,|\,$2{\cdot} 10^{7}$ & 15\,|\,$2{\cdot} 10^{7}$ & 18\,|\,$2{\cdot} 10^{7}$ & 25\,|\,$1{\cdot} 10^{7}$ & 18\,|\,$2{\cdot} 10^{7}$ & 21\,|\,$2{\cdot} 10^{7}$ & 20\,|\,$2{\cdot} 10^{7}$ & 21\,|\,$9{\cdot} 10^{6}$\\
\hline
\model{Netw-p} & \multirow{2}{*}{$R_\mathrm{max}$} & $8019$ & \multirow{2}{*}{$1035$} & ${\le}\, \mathbf{64.3}$$^\dagger$ & ${\le}\, \mathbf{64.3}$$^\dagger$ & ${\le}\, \mathbf{64.3}$$^\dagger$ & ${\le}\, 65.7$$^\dagger$ & ${\le}\, \mathbf{64.3}$$^\dagger$ & ${\le}\, \mathbf{64.3}$$^\dagger$ & ${\le}\, \mathbf{64.3}$$^\dagger$ & ${\le}\, \mathbf{64.3}$ & ${\le}\, \mathbf{64.3}$$^\dagger$ & ${\le}\, \mathbf{64.3}$$^\dagger$ & ${\le}\, \mathbf{64.3}$$^\dagger$ & ${\le}\, \mathbf{64.3}$\\
3-5-2 &  & $2{\cdot} 10^{4}$ &  & 6\,|\,$1{\cdot} 10^{7}$ & 12\,|\,$1{\cdot} 10^{7}$ & 6\,|\,$1{\cdot} 10^{7}$ & 6\,|\,$7{\cdot} 10^{6}$ & 10\,|\,$1{\cdot} 10^{7}$ & 6\,|\,$5{\cdot} 10^{6}$ & 6\,|\,$1{\cdot} 10^{7}$ & 12\,|\,$8{\cdot} 10^{6}$ & 6\,|\,$1{\cdot} 10^{7}$ & 6\,|\,$1{\cdot} 10^{7}$ & 10\,|\,$9{\cdot} 10^{6}$ & 8\,|\,$1{\cdot} 10^{7}$\\
\hline
\model{Netw-p} & \multirow{2}{*}{$R_\mathrm{max}$} & $2{\cdot} 10^{5}$ & \multirow{2}{*}{$2{\cdot} 10^{4}$} & ${\le}\, 826$ & ${\le}\, 824$ & ${\le}\, 826$ & ${\le}\, 826$ & ${\le}\, 832$ & ${\le}\, 824$ & ${\le}\, 825$ & ${\le}\, 831$ & ${\le}\, 825$ & ${\le}\, 834$ & ${\le}\, 829$ & ${\le}\, \mathbf{823}$\\
3-8-20 &  & $3{\cdot} 10^{5}$ &  & 2\,|\,$2{\cdot} 10^{6}$ & 4\,|\,$3{\cdot} 10^{6}$ & 2\,|\,$2{\cdot} 10^{6}$ & 2\,|\,$2{\cdot} 10^{6}$ & 9\,|\,$4{\cdot} 10^{6}$ & 2\,|\,$2{\cdot} 10^{6}$ & 2\,|\,$4{\cdot} 10^{6}$ & 3\,|\,$2{\cdot} 10^{6}$ & 2\,|\,$4{\cdot} 10^{6}$ & 2\,|\,$3{\cdot} 10^{6}$ & 9\,|\,$4{\cdot} 10^{6}$ & 2\,|\,$4{\cdot} 10^{6}$\\
\hline\hline
\model{Netw} & \multirow{2}{*}{$R_\mathrm{min}$} & $4589$ & \multirow{2}{*}{$1173$} & ${\ge}\, \mathbf{3.2}$ & ${\ge}\, \mathbf{3.2}$ & ${\ge}\, \mathbf{3.2}$ & ${\ge}\, \mathbf{3.2}$ & ${\ge}\, \mathbf{3.2}$ & ${\ge}\, \mathbf{3.2}$ & ${\ge}\, \mathbf{3.2}$ & ${\ge}\, \mathbf{3.2}$$^\dagger$ & ${\ge}\, \mathbf{3.2}$ & ${\ge}\, \mathbf{3.2}$ & ${\ge}\, \mathbf{3.2}$ & ${\ge}\, \mathbf{3.2}$\\
2-8-20 &  & $6973$ &  & 11\,|\,$1{\cdot} 10^{7}$ & 24\,|\,$2{\cdot} 10^{7}$ & 12\,|\,$1{\cdot} 10^{7}$ & 11\,|\,$1{\cdot} 10^{7}$ & 11\,|\,$1{\cdot} 10^{7}$ & 11\,|\,$1{\cdot} 10^{7}$ & 25\,|\,$3{\cdot} 10^{7}$ & 32\,|\,$2{\cdot} 10^{7}$ & 26\,|\,$3{\cdot} 10^{7}$ & 25\,|\,$3{\cdot} 10^{7}$ & 25\,|\,$3{\cdot} 10^{7}$ & 32\,|\,$1{\cdot} 10^{7}$\\
\hline
\model{Netw} & \multirow{2}{*}{$R_\mathrm{min}$} & $843$ & \multirow{2}{*}{$111$} & ${\ge}\, \mathbf{1.9}$ & ${\ge}\, \mathbf{1.9}$ & ${\ge}\, \mathbf{1.9}$ & ${\ge}\, \mathbf{1.9}$ & ${\ge}\, \mathbf{1.9}$ & ${\ge}\, \mathbf{1.9}$ & ${\ge}\, \mathbf{1.9}$ & ${\ge}\, \mathbf{1.9}$ & ${\ge}\, \mathbf{1.9}$ & ${\ge}\, \mathbf{1.9}$ & ${\ge}\, \mathbf{1.9}$ & ${\ge}\, \mathbf{1.9}$\\
3-5-2 &  & $1515$ &  & 12\,|\,$2{\cdot} 10^{7}$ & 23\,|\,$1{\cdot} 10^{7}$ & 11\,|\,$1{\cdot} 10^{7}$ & 12\,|\,$1{\cdot} 10^{7}$ & 12\,|\,$1{\cdot} 10^{7}$ & 11\,|\,$8{\cdot} 10^{6}$ & 21\,|\,$2{\cdot} 10^{7}$ & 32\,|\,$6{\cdot} 10^{6}$ & 21\,|\,$2{\cdot} 10^{7}$ & 21\,|\,$2{\cdot} 10^{7}$ & 23\,|\,$1{\cdot} 10^{7}$ & 36\,|\,$3{\cdot} 10^{6}$\\
\hline
\model{Netw} & \multirow{2}{*}{$R_\mathrm{min}$} & $2{\cdot} 10^{4}$ & \multirow{2}{*}{$2205$} & ${\ge}\, 6.73$$^\dagger$ & ${\ge}\, 6.71$$^\dagger$ & ${\ge}\, 6.73$$^\dagger$ & ${\ge}\, 5.55$$^\dagger$ & ${\ge}\, 6.73$$^\dagger$ & ${\ge}\, \mathbf{6.74}$$^\dagger$ & ${\ge}\, 6.72$$^\dagger$ & ${\ge}\, 6.68$$^\dagger$ & ${\ge}\, 6.72$$^\dagger$ & ${\ge}\, 6.16$$^\dagger$ & ${\ge}\, 6.72$$^\dagger$ & ${\ge}\, \mathbf{6.74}$$^\dagger$\\
3-8-20 &  & $3{\cdot} 10^{4}$ &  & 5\,|\,$1{\cdot} 10^{7}$ & 10\,|\,$1{\cdot} 10^{7}$ & 5\,|\,$1{\cdot} 10^{7}$ & 6\,|\,$1{\cdot} 10^{7}$ & 5\,|\,$1{\cdot} 10^{7}$ & 6\,|\,$2{\cdot} 10^{7}$ & 5\,|\,$1{\cdot} 10^{7}$ & 9\,|\,$1{\cdot} 10^{7}$ & 5\,|\,$1{\cdot} 10^{7}$ & 6\,|\,$2{\cdot} 10^{7}$ & 5\,|\,$1{\cdot} 10^{7}$ & 6\,|\,$1{\cdot} 10^{7}$\\
\hline\hline
\model{Nrp} & \multirow{2}{*}{$P_\mathrm{max}$} & $125$ & \multirow{2}{*}{$41$} & ${=}\, \mathbf{0.12}$* & ${=}\, \mathbf{0.12}$* & ${=}\, \mathbf{0.12}$* & ${=}\, \mathbf{0.12}$* & ${=}\, \mathbf{0.12}$* & ${=}\, \mathbf{0.12}$* & ${=}\, \mathbf{0.12}$* & ${=}\, \mathbf{0.12}$* & ${=}\, \mathbf{0.12}$* & ${=}\, \mathbf{0.12}$* & ${=}\, \mathbf{0.12}$* & ${=}\, \mathbf{0.12}$*\\
8 &  & $161$ &  & 6\,|\,$45$ & 8\,|\,$45$ & 7\,|\,$45$ & 6\,|\,$45$ & 6\,|\,$45$ & 6\,|\,$38$ & 70\,|\,$32$ & 136\,|\,$32$ & 71\,|\,$32$ & 70\,|\,$32$ & 70\,|\,$32$ & 59\,|\,$32$\\
\hline
\end{tabular}

		}
	}
\end{table}

\end{document}